\documentclass[letter, 12pt]{article}
\usepackage[margin = 0.95in]{geometry}
\usepackage[english]{babel}
\usepackage{standalone}
\usepackage[T1]{fontenc}
\usepackage[singlespacing]{setspace}

\usepackage{amsmath, amssymb, amsthm, amsbsy, amsfonts, natbib, wrapfig,graphicx, comment, enumitem, algorithm, relsize,  algorithmic,indentfirst, colortbl, titlesec} 
\usepackage[colorinlistoftodos]{todonotes}
\usepackage[colorlinks=true, allcolors=blue, pdfencoding=auto, psdextra]{hyperref}

\theoremstyle{definition}
\newtheorem{thm}{Theorem}
\newtheorem{defn}{Definition}

\newtheorem{cor}{Corollary}
\newtheorem{pro}{Proposition}
\newtheorem{rem}{Remark}
\newtheorem{cla}{Claim}

\newcommand{\x}{\mathbf{x}}
\newcommand{\X}{\mathbf{X}}
\newcommand{\y}{\mathbf{y}}

\newcommand{\e}{\mathbf{e}}

\newcommand{\E}{\mbox{E}}
\newcommand{\V}{\mbox{V}}

\newcommand{\bs}{\boldsymbol}

\newcommand{\1}{\mathbf{1}}
\newcommand{\0}{\mathbf{0}}
\newcommand{\diag}{\mbox{diag}}

\allowdisplaybreaks 
\makeatletter
\g@addto@macro\normalsize{\setlength\abovedisplayskip{3pt}}
\g@addto@macro\normalsize{\setlength\belowdisplayskip{3pt}}
\makeatother

\titlespacing*{\section}
{0pt}{3pt plus 3pt minus 3pt}{1pt plus 1pt}
\titlespacing*{\subsection}
{0pt}{3pt plus 3pt minus 3pt}{1pt plus 1pt}
\usepackage[font=small,skip=3pt]{caption}

\title{
\vspace{-24pt}
\Large Adaptive Noisy Data Augmentation for Regularized Estimation and Inference in Generalized Linear Models}
\vspace{-24pt}
\author{\normalsize Yinan Li$^{1}$ and Fang Liu$^1$\footnote{Corresponding author email: fang.liu.131@nd.edu}\\
\small$^1$ Department of Applied and Computational Mathematics and Statistics\\
\small University of Notre Dame, Notre Dame, IN 46556, U.S.A. }
\date{}

\begin{document}
\maketitle
\vspace{-32pt}
\begin{abstract}	
We propose the AdaPtive Noise Augmentation (PANDA) procedure to regularize the estimation and inference of generalized linear models (GLMs). PANDA iteratively optimizes the objective function given noise augmented data until convergence to obtain the regularized model estimates.  The augmented noises are designed to achieve various regularization effects, including $l_0$, bridge (lasso and ridge included), elastic net, adaptive lasso, and SCAD, as well as group lasso and fused ridge. We examine the tail bound of the noise-augmented loss function and establish the almost sure convergence of the noise-augmented loss function and its minimizer to the expected penalized loss function and its minimizer, respectively. We derive the asymptotic distributions for the regularized parameters, based on which, inferences can be obtained simultaneously with variable selection. PANDA exhibits ensemble learning behaviors that help further decrease the generalization error. Computationally, PANDA is easy to code, leveraging existing software for implementing GLMs, without resorting to complicated  optimization techniques.  We demonstrate the superior or similar performance of PANDA  against the existing  approaches of the same type of regularizers in simulated and real-life data. We show that the inferences through PANDA achieve nominal or near-nominal coverage and are far more efficient compared to a popular existing post-selection procedure.   
\vspace{6pt}

\noindent \textbf{keywords}: $l_0$ penalty, augmented Fisher information, ensemble learning, noise injection and augmentation,  regularization and penalization, inference
\end{abstract}

\section{Introduction}\label{sec:intro}\vspace{-3pt}
Regularization of generalized linear models (GLMs) to mitigate overfitting and conduct variable selection  is a well-studied topic. There exist a variety of regularizers, such as bridge \citep{bridge}, ridge ($l_2$), lasso ($l_1$)  \citep{lasso},  elastic net \citep{en}, SCAD \citep{SCAD},  adaptive lasso \citep{adaptivelasso}, group lasso \citep{grouplasso}, fused lasso \citep{fusedlasso}, sparse group lasso (SGL) \citep{SGL}, among others. As for the inference for regression coefficients in penalized GLMs, many existing approaches are post-election procedures, meaning the inference is initiated after variable selection and oftentimes  non-selected variables are assumed to be of no inferential interest and no uncertainty quantification are provided for the corresponding regression coefficients \citep{CIleeb2005model,CIleeb2006can,CIberk2013valid,CIZhang2013,CIAdel2014, CIlockhart2014significance,CIefron2014estimation,CIJason2016, CIRyan2016, CIStephen2017,CIJonathan2017}. Procedures for simultaneous variable selection and inference do exist. \citet{SCAD}  provide simultaneous variance estimates for the coefficients of selected estimated variables (so non-zero coefficient estimates).  For linear regression, \citep{CIZhang2013, CIAdel2014} provide simultaneous variable selection with the lasso penalty and inference for both zero or non-zero coefficient estimates. \citet{CIvan2014asymptotically} propose a procedure for constructing confidence intervals and  hypothesis testing for a low-dimensional subset of a large parameter vector in the high-dimensional GLM setting with convex loss functions with the lasso penalty. 

Despite the extensiveness of the work on regularized  variable selection in GLMs, there is still room for improvement over the existing solutions. Two of these areas are the $l_0$ regularization and inference in regularized GLMs. Optimization with the $l_0$ penalty is NP-hard.  \citet{Lin} propose the seamless-$l_0$ (SELO) penalty to approximate the $l_0$ penalty and a coordinate descent algorithm to obtain solutions in the context of the least-squares optimization and $p<n$. SELO outperforms SCAD in model error and variable selection accuracy rate per the empirical studies.   \citet{LiGang} propose an  EM algorithm that approximates the $l_0$ regularized regression by solving a sequence of $l_2$ optimizations. The method  deals with $p>n$, but is examined only in the  least-squares setting and does not provide inferential procedures. Regarding the inference for regularized GLMs, as mentioned above, the majority of existing methods operate  in a post-selection matter and thus focus on the inference for  selected variables only. \citet{SCAD} and \citet{Lin} provide standard errors for the  parameter estimates in non-convex optimization, and again for selected variables only.   

We propose a novel general regularization framework, AdaPtive Noisy Data Augmentation  (PANDA), for GLMs that 1) achieves the $l_0$ penalty in addition to all the above mentioned existing penalty types, 2) obtains inference in regularized GLMs for both zero and non-zero coefficients, and 3) enjoys simple practical implementation that would greatly appeal to practitioners. 
In brief, PANDA augments the original $n$ observations with properly designed $n_e$ noise terms to achieve the desired regularization effects on model parameters.  PANDA is  iterative and the variance terms of the augmented noise are adaptive to the most updated parameter estimates  until the algorithm converges. One requirement on $n_e$ is the augmented data  size $n+n_e>p$ so to allow for the ordinary least squares (OLS) or maximum likelihood estimation (MLE) procedures to be applied without resorting to complicated optimization algorithms. As such, \emph{PANDA is computationally straightforward and efficient}. 
\emph{PANDA is also flexible and general}. By properly designing the variance of the augmented  noise, PANDA can achieve various regularization effects, including $l_{\gamma}$, for $0\le\gamma\le2$ (including $l_0$, lasso, ridge as special cases), elastic net, SCAD, group lasso, and fused ridge. PANDA achieves  close-to-exact  $l_0$ regularization
by promoting orthogonality between the coefficients and the augmented noise vector. 
When $n_e<p$,  PANDA shrinks exactly $n_e$ parameters towards $0$ upon convergence. \emph{PANDA is more capable and more efficient inferentially compared to existing inferential approaches for regularized GLMs}. It conducts variable selection and provides  inference for  coefficients simultaneously, whether the coefficients are estimated to be zero or not.  Our empirical results suggest the inference based on PANDA is valid and more efficient compared to some existing post-selection procedures.  Finally, \emph{PANDA is theoretically justified}. We establish the Gaussian tail of the noise-augmented loss function  and the almost sure convergence to its expectation under some regularity conditions, providing theoretical justification for PANDA as a regularization technique and that the noise-augmented loss function is trainable for practical implementation. 

The optimizer calculated by PANDA from a GLM is similar to the local quadratic approximation (LQA) technique \citep{lasso,SCAD}, but with several important differences. First, LQA cannot yield the $l_0$ penalty while PANDA can achieve close-to-exact $l_0$ regularization; second, LQA relies on analytical work to approximate  penalized loss function with a quadratic form, followed by the optimization of the quadratic function, whereas PANDA only needs to augment the original data with noisy samples and then leverage existing software to compute OLS/MLE from GLMs. 

The rest of the paper is organized as follows. Sec \ref{sec:method} presents the PANDA algorithm and the  regularization effects it brings to GLMs. Sec \ref{sec:theory} establishes the consistency on the noise-augmented loss function and the regularized parameter estimates,  presents the Fisher information of the model parameters in augmented data, examines PANDA's ensemble learning behavior, and provides the asymptotic distributions for the parameter estimates via PANDA. Sec \ref{sec:examples} demonstrates the $l_0$ penalty realized by PANDA, compares PANDA  to a popular post-selection inferential approach in statistical inferences  for  GLMs, and implements PANDA in simulated and real-life studies to show its effectiveness in regularizing GLM estimation. Sec \ref{sec:discussion} provides some concluding remarks and offers future research directions on PANDA.


\section{Methodology}\label{sec:method}\vspace{-3pt}
\subsection{Noise Augmentation Scheme and Regularization \normalsize{Effect in PANDA}} \label{sec:regularization}\vspace{-3pt}
Let $Y$ be the outcome variable and  $\X=(X_1,\ldots,X_p)^T$ be the  independent variables. GLM is based on the assumption that the conditional distribution of $Y$ given  $\X$ comes from an exponential family
\begin{equation}\label{eqn:expfam}
p(Y|\X)=\exp\left(Y\eta-B(\eta)+h(Y)\right),
\end{equation}
\begin{wrapfigure}{r}{0.35\textwidth}
\centering\includegraphics[width=0.31\textwidth]{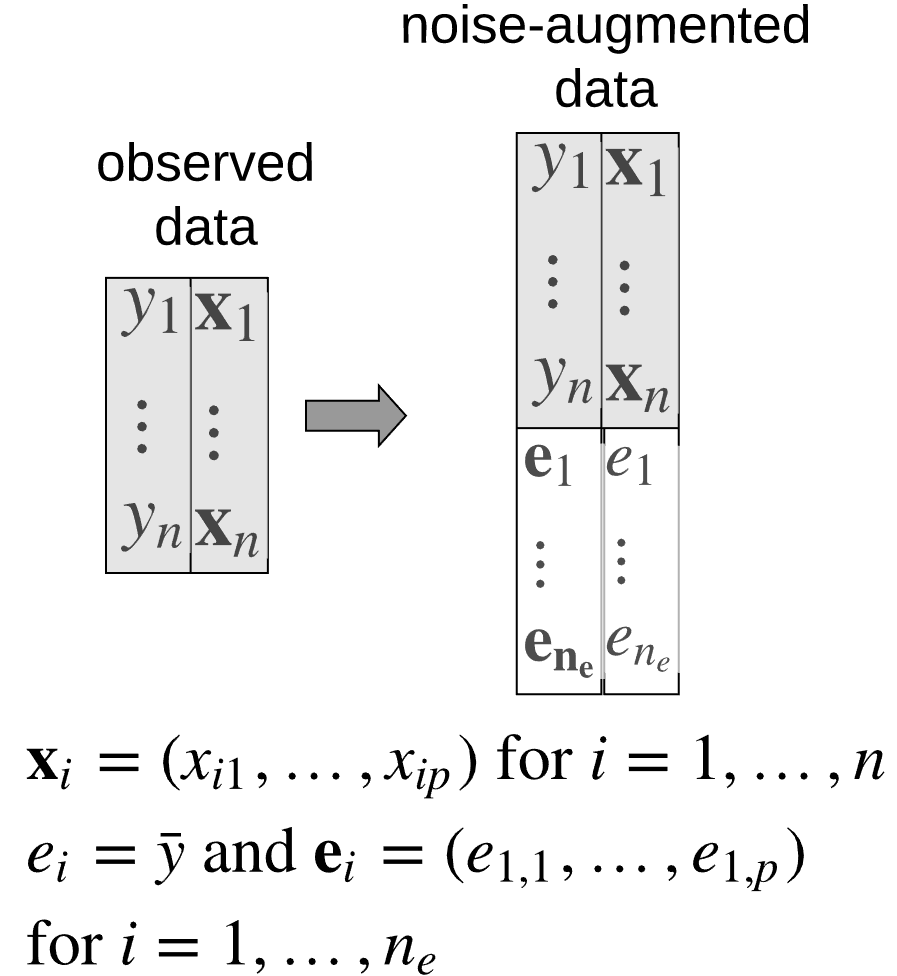}
\caption{A schematic of the data augmentation for GLM in PANDA (\small for logistic regression, $e_{y,i}\sim$ Bern($\hat{p}$), where $\hat{p}$ is the sample proportion of an event)}\label{fig:pandaGLM}\vspace{3pt}
\end{wrapfigure}
where $\eta=\bs\theta_{0}+\bs\theta\X$ if the canonical link is used  (e.g., the identity link for Gaussian $Y$; the logit link for Bernoulli $Y$). When $p$ is large, regularization or penalty is often imposed  on $\bs\theta$ when estimating $\bs\theta$.

PANDA regularizes the estimation of  $\bs{\bs\theta}$ by first augmenting the  observed data with a noisy data matrix. Fig \ref{fig:pandaGLM} depicts a schematic of data augmentation in PANDA, where the augmented noise $e_i$ to $\y$ is $\bar{y}$, the sample average of $\y$. For logistic regression, $e_i\sim$ Bern($\hat{p}$), where $\hat{p}$ is the sample proportion of an event. The augmented data $\e_x$ to $\x$ are drawn from the \emph{Noise Generating Distributions} (NGD), the variance term of which is function of $\bs\theta$ and tuning parameters $\bs\lambda$.
\begin{equation}\label{eqn:NGD}
\e_x\sim N\left(0, \V(\bs\theta; \bs\lambda)\right) 
\end{equation}

\begin{pro}[\textbf{regularization effects of PANDA for GLM}]\label{prop:glmregularization}
Denote the loss function given the observed data $(\x,\y)$ by $l(\!\bs\theta|\x,\y)\!=\!-\!\sum_{i=1}^n\! \left(h(y_i)\!+\!\left(\theta_{0}\!+\!\!\sum_j\!\theta_jx_{ij}\right)y_i\!-\!j(\theta_{0}\!+\!\sum_j\!\theta_jx_{ij})\right)$ (the negative log-likelihood function), and that given the noise augmented data $(\tilde{\x},\tilde{\y}))$ by
\begin{equation}\label{eqn:GLMloss}
\textstyle l_p(\bs\theta|\tilde{\x}, \tilde{\y})\!=\! -\left\{\sum_{i=1}^{n+n_e} \left(h(\tilde{y}_i)\!+\!\left(\theta_{0}+\sum_j\theta_j\tilde{x}_{ij}\right)\tilde{y}_i\right)\!-\!B_j\left(\theta_0+\sum_j\theta_j\tilde{x}_{ij}\right)\right\},
\end{equation}
where $\bs\theta\!=\!(\theta_0,\{\theta_j\}_{j=1,\ldots,p})$. The expectation of the Taylor series of $l_p$ around $\sum_j\theta_jx_{ij}=0$ over the distribution of $\e_x$ is
\begin{align}
&\E_{\e_x}(l_p(\bs\theta|\tilde{\x},\tilde{\y}))= l(\bs\theta|\x,\y)+P(\bs\theta), \mbox{ where}\notag \\
&P(\bs\theta)\!=\!\textstyle
n_e\!\left(C_1\!\sum_j\theta^2_j\mbox{V}(e_j)\right) \!+\!O\!\left(\!n_e\!\sum_j\!\left(\theta_j^4\V^2(e_j)\right)\right)+C,\label{eqn:glmregularization}
\end{align} 
where $C_1\!=\!2^{-1}B''(\theta_{0})$ and $C\!=\!\sum_{i=1}^{n_e} \left(h(e_{y,i})\!+\!e_{y,i}\theta_0\right)\!+\!B(\theta_0)$ are constants independent of $\bs\theta_{-0}$.\vspace{-6pt}
\end{pro}
The proof is given in Sec \ref{app:glmregularization} of the supplementary materials. The regularization effect  $P(\bs\theta)$ in Eqn (\ref{eqn:glmregularization}) depends on the variance term of the NGD in Eqn (\ref{eqn:NGD}) from which the augmented noise to $\x$ is sampled.  Eqns (\ref{eqn:bridge}) to (\ref{eqn:scad}) list some examples of the NGD  from which $e_j$, the noise term that augments $\x_j$ for $j=1,\ldots,p$, is drawn  and  their expected regularization effects. 
\begin{align}
\!\!\!\mbox{bridge: } & e_j\sim N\left(0, \lambda|\theta_j|^{-\gamma}\right)\mbox{ for $\gamma\in[0,2]$} \label{eqn:bridge} \\ 
&\qquad\mbox{including $l_0$ when $\gamma=2$, lasso when $\gamma=1$, and ridge when $\gamma=0$;}\notag\\
\!\!\!\mbox{elastic net: } & e_j\sim N\left(0, \lambda|\theta_j|^{-1}+\sigma^2\right); \label{eqn:en}\\
\!\!\!\mbox{adaptive lasso: } & e_j\sim N\left(0, \lambda |{\theta}_j|^{-1}|\hat{\theta}_j|^{-\gamma}\right),\mbox{where  $\hat{\theta}_j$ is a consistent estimate for $\theta_j$};\label{eqn:adaptivelasso}\\
\!\!\!\mbox{SCAD: } e_j & 
\begin{cases}
=0 \mbox{ if }|\theta_j|>an_e\lambda \\
\sim \!N\!\left(0,\left(\!\frac{\lambda}{|{\theta}_j|}\!-\!\frac{(a+1)}{2a^2n_e}\!\right)\!1_{(0,n_e\lambda]}(|{\theta}_j|) \!+\! \frac{1}{(a-1)}\!\left(\!\frac{a\lambda }{|{\theta}_j|}\!-\!\frac{\lambda^2n_e}{2{\theta}_j^2}\!-\!\frac{2a^2-1}{2a^2n_e}\right)\! 1_{(n_e\lambda,an_e\lambda]} (|{\theta}_j|)\!\right)\mbox{o.w.}
\end{cases}, \notag \\
& \mbox{ where $1_{(l,u)}(|{\theta}_j|)\!=\!1$ if $l\!<\!|{\theta}_j|\!<\!u$, 0 o.w.;}\label{eqn:scad}
\end{align}
For regularizing a group of $q$ parameters $\bs\theta_q=(\theta_1,\ldots,\theta_q)$ simultaneously (e.g., genes on the same pathway, binary dummy variables created from the same categorical attribute), the NGDs in Eqns (\ref{eqn:group}) and (\ref{eqn:fusedlasso}) can be used. Specifically, the group-lasso penalty sets all $q$ parameters in $\bs\theta_q$  either at zero or nonzero simultaneously; and the  fused-ridge and fused-lasso penalties  promote numerical similarity among $\bs\theta_q$. 
\begin{align}
\mbox{group lasso: } & e_{j{(l)}}\sim N\left(\!0, \frac{\lambda\sqrt{q_l}}{||\bs\theta_l||_2} \right)\!,
\mbox{where }\bs\theta^{(l)}\!=\!\{\theta_{j{(l)}}\} \mbox{ for } j\!=\!1,\ldots,q_l; l\!=\!1,\ldots,g \mbox{ groups}\!\! \label{eqn:group}\\
\!\!\!\mbox{fused ridge: }&\e=(e_1,\ldots,e_q)\sim N_{(q)}\left(0,\lambda(\mathbf{T}\mathbf{T}')\right),\mbox{where entries in $\mathbf{T}$ are}\label{eqn:fusedridge} \\
&\mbox{\hspace{54pt}$T_{k,k}=1,T_{k+1-k\cdot1(k=q),k}=-1$; and 0 o.w.};\notag\\
\!\!\!\mbox{fused lasso: }&\e=(e_1,\ldots,e_q)\sim N_{(q)}\left(0,\lambda(\mathbf{T}\mathbf{T}')\right),\mbox{where } T_{kk'}\!=\!\lambda|\theta_j\!-\!\bs\theta_{k'}|^{-1} \mbox{for }k\!\ne\! k'.\label{eqn:fusedlasso} 
\end{align}
The tuning parameters $\sigma^2\ge0, \lambda>0$, $0\le\gamma<2$, $a>2$ in Eqn (\ref{eqn:bridge}) and (\ref{eqn:fusedlasso}) can be user-specified or chosen by a model selection criterion such as  cross-validation (CV), AIC, or BIC. The dispersion of the noise term varies by $X$ in general. $X_j$ associated with small  $|\theta_j|$ is augmented with more spread-out noises, and $X_j$ with large  $|\theta_j|$ is augmented with noises concentrated around 0. 
The exceptions are the ridge ($\gamma=0$ in Eqn (\ref{eqn:bridge})) and fused ridge regularizations (Eqn (\ref{eqn:fusedridge})), where the variance term remains constant for $\theta_j$ for all $j$.
 
For linear regression,  the noise-augmented loss function is  $l_p(\bs\theta|\tilde{\x},\tilde{\y})\!=\!
\sum_{i=1}^{n+n_e}\!\big(\tilde{y}_{i}\!-\!\sum_j\tilde{x}_{ij}\theta_j\!\big)^2\!$, and the penalty $P(\bs\theta)$ realized by PANDA with different types of NGD can be obtained in closed form (Table \ref{tab:GGM}) and are exact as suggested by the names.  $P(\bs\theta)$ in Table \ref{tab:GGM} can be easily derived based on the results in Sec \ref{app:glmregularization} of the supplementary materials.  Compared to the original SCAD in  \citet{SCAD} for linear regression, the SCAD penalty realized by PANDA for $|\theta_j|<\lambda n_e$ is not $l_1$ as in the original SCAD but closer to a $l_{0.5}$ penalty; the middle segment penalty  $|\theta_j|\in(\lambda n_e, an_e\lambda]$ is not exactly the same as the original SCAD either, but it has the same functionality by shrinking $|\theta_j|\in(\lambda n_e, |\theta_j|<a\lambda n_e]$ toward 0 and connecting the two end segments to form an overall smooth penalty for $\theta_j$; for $|\theta_j|>a\lambda n_e$, there is no penalty as in the original SCAD.
\begin{table}[!htb]
\begin{center}
\caption{Close-formed penalty term in regularized linear regression (Gaussian $Y$) via PANDA}\label{tab:GGM}
\begin{tabular}{lll}
\hline 
\multicolumn{2}{c}{NGD} & $P(\bs\theta)$ when $Y$ is Gaussian   \\
\hline
$l_\gamma$ & Eqn (\ref{eqn:bridge}) &  $(\lambda n_e)\sum_{j=1}^{p}\sum_{j\neq k}|\theta_j|^{2-\gamma}$\\
EN &  Eqn (\ref{eqn:en}) & $(\lambda n_e)\sum_{j=1}^{p}\sum_{j\neq k}|\theta_j|+(\sigma^2 n_e)\sum_{j=1}^{p}\sum_{j\neq k}\theta_j^2$ \\
adaptive & Eqn (\ref{eqn:adaptivelasso}) & $(\lambda n_e)\sum_{j=1}^{p}\sum_{j\neq k}|\theta_j||\hat{\theta}_j|^{-\gamma}$,\\
SCAD &  Eqn (\ref{eqn:scad})  & $\!\sum_{j=1}^{p}\left\{\left(n_e\lambda|\theta_j|-(a+1)\theta^2/(2a^2n_e))\right)\!\cdot\!1_{|\theta_j|\le\lambda n_e}+\right.$\\
&& $\quad\quad\;\left.\left(a\lambda n_e| \theta_j| \!-\!(\lambda n_e)^2/2\!-\!\theta_j^2(1\!-\!1/(2a^2))\right)(a\!-\!1)^{-1}\!\cdot\! 1_{|\theta_j|\in(\lambda n_e,a\lambda n_e]}\right\}$\\
group lasso &  Eqn (\ref{eqn:group})  &   $(\lambda n_e)\sum_{l=1}^{g}\sqrt{q_l}||\bs\theta_l||_2$\\
\hline
\end{tabular}
\end{center}
\vspace{-18pt}
\end{table}

When $Y$ is non-Gaussian, the achieved regularization effects  $P(\bs\theta)$ in Eqns (\ref{eqn:bridge}) to (\ref{eqn:fusedlasso}) are second-order approximate. For example, Table \ref{tab:glm} lists  the analytical form of  $P(\bs\theta)$ for the lasso-type noise ($\gamma=1$ in Eqn (\ref{eqn:bridge})). For all the regression types,  in addition to the $l_1$ penalty, there is an additional big-O term on $\bs{\theta}$, which is arbitrarily small under some regularity conditions (more details are provided in Sec \ref{sec:nem}). 
\begin{table}[!htp]
\begin{center}
\caption{Expected penalty term in PANDA with lasso-type noise $e_j\sim N\left(0,\lambda|\theta_j|^{-\gamma}\right)$} \label{tab:glm}
\begin{tabular}{@{}l@{\hskip 6pt}l@{}}
\hline
$\qquad Y$   & $P(\bs\theta)$\\
\hline
Bernoulli& $\frac{\lambda n_e}{2}\frac{\exp(\theta_0)}{(1+\exp(\theta_0))^2}\sum_j|\theta_j|+O(\lambda^2 n _e ||\bs\theta||_2^2)+C$\\
Exponential&  $\frac{\lambda n_e}{2}\exp(\bs\theta_{0})\sum_j|\theta_j|+O(\lambda^2 n _e ||\bs\theta||_2^2)+C$\\
Poisson &  $\frac{\lambda n_e}{2}\exp(\theta_0)\sum_j|\theta_j|+O(\lambda^2 n _e ||\bs\theta||_2^2)+C$\\
Negative Binomial& $\frac{\lambda n_e}{2}\frac{r\exp(\bs\theta_{0})}{(r+\exp(\bs\theta_{0}))}\sum_j|\theta_j|+O(\lambda^2 n _e ||\bs\theta||_2^2)+C$ ($r$ is the \# of failures)\\
\hline
\end{tabular}
\end{center}
\vspace{-18pt}
\end{table}

For relatively small $n_e$, especially when $n_e<p$, PANDA promotes sparsity on $\bs\theta$ by imposing $n_e$ linear constraints on $\bs\theta$. Applying the second-order approximation at $\e^T_i\bs\theta=0$, we have  
\begin{align}\label{eqn:GLMloss2}
 l_p(\bs\theta|\tilde{\x})\!&=\textstyle\! -\left\{\sum_{i=1}^{n+n_e} \left(h(\tilde{y}_i)\!+\!\left(\theta_{0}+\sum_j\theta_j\tilde{x}_{ij}\right)\tilde{y}_i\right)
\!-\!B_j\left(\theta_0+\sum_j\theta_j\tilde{x}_{ij}\right)\right\}\notag\\
&\textstyle \approx l(\bs\theta|\x)+ C_1\sum_{i=1}^{n_e} \left(\sum_j\theta_j e_{ij} \right)^2+C,
\end{align}
where $C_1$ and $C$ the same as in Eqn (\ref{eqn:glmregularization}).
The regularization effect obtained in Eqn (\ref{eqn:GLMloss2}) with fixed $n_e$ is different from the  regularization presented in Proposition \ref{prop:glmregularization} in the sense that it takes effect by promoting the orthogonality between $\bs\theta$ and $\e_{x,i},i=1\ldots,n_e$ rather than penalizing the individual parameters. The formal results are given in Proposition \ref{prop:glmorthogonalregularization}. The proof is provided in  Sec \ref{app:glmorthogonalregularization} of the supplementary materials.  

\begin{figure}[!htb]
\begin{minipage}{1\textwidth}
\includegraphics[width=0.325\linewidth,trim={0.5cm 2cm 0.5cm 0.5cm},clip]{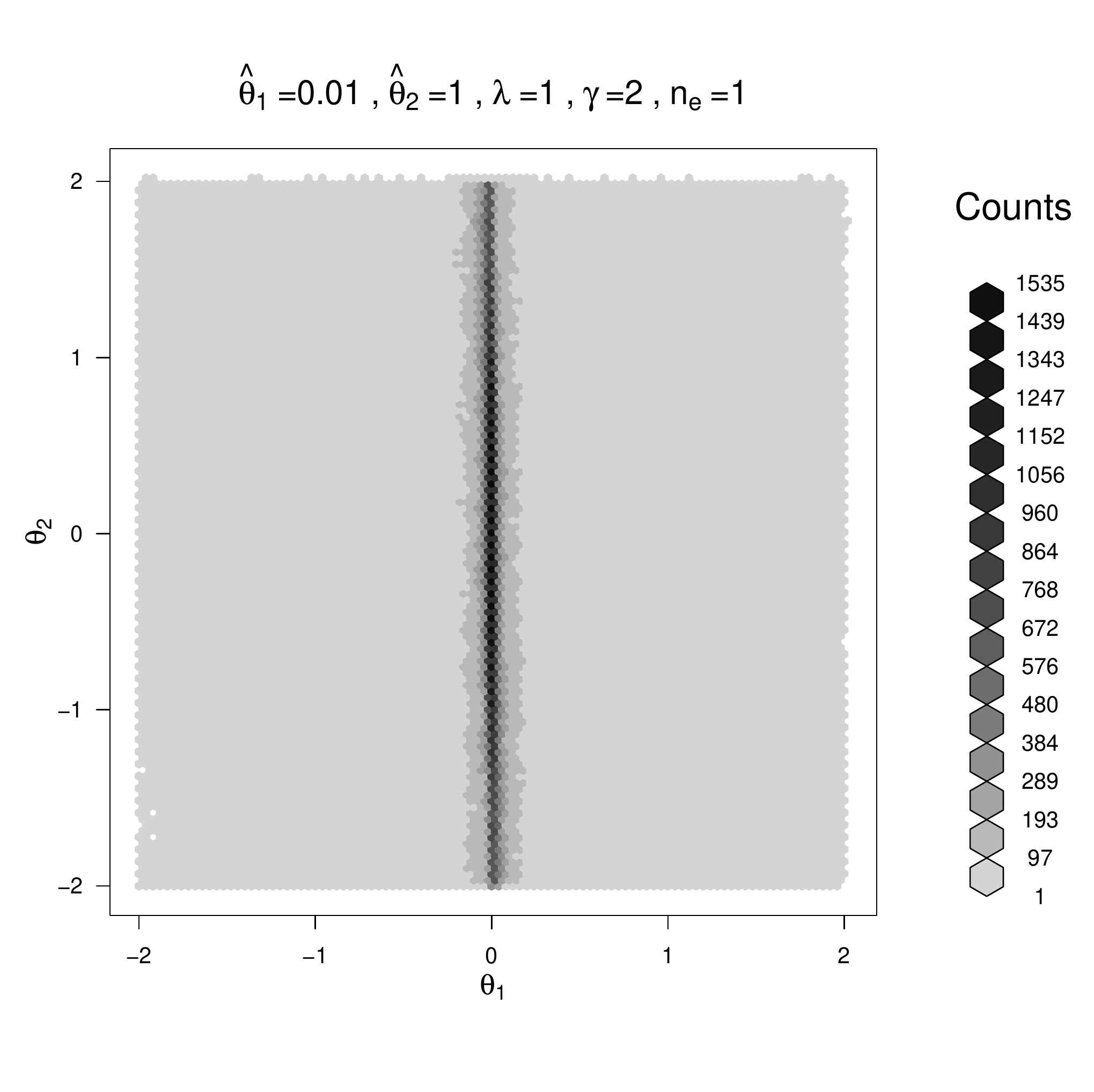}
\includegraphics[width=0.325\linewidth,trim={0.5cm 2cm 0.5cm 0.5cm},clip]{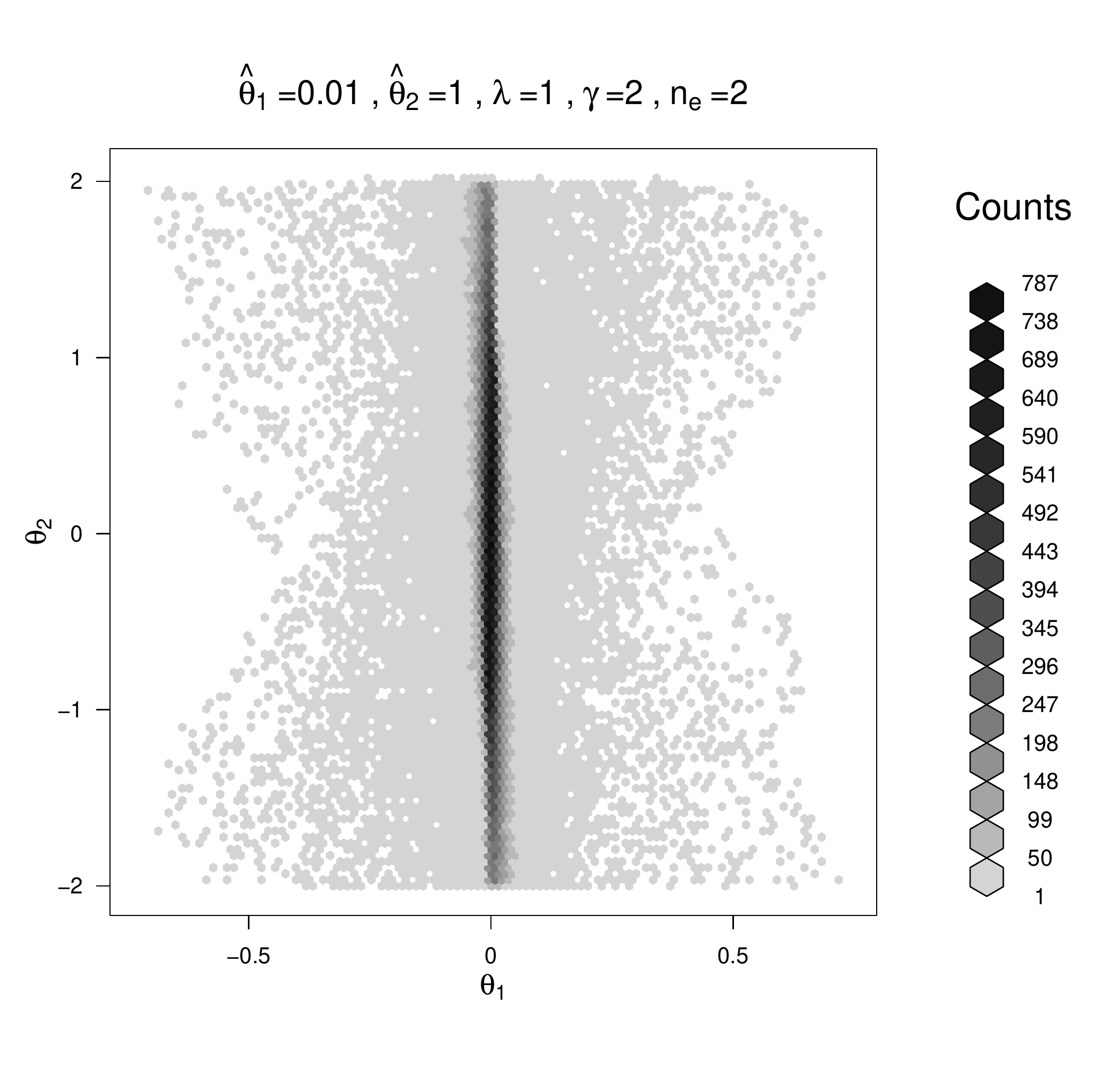}
\includegraphics[width=0.325\linewidth,trim={0.5cm 2cm 0.5cm 0.5cm},clip]{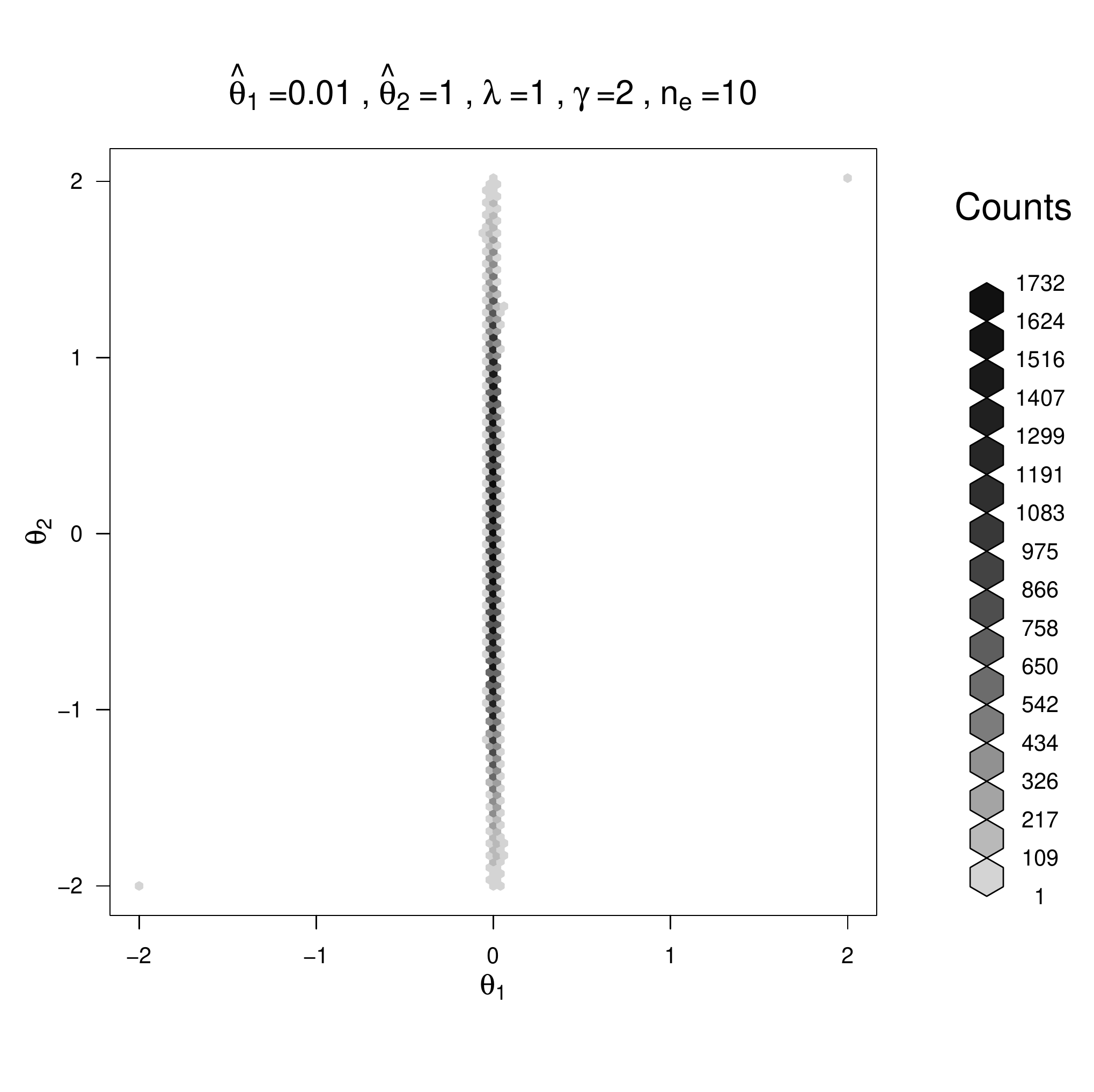}\\
\includegraphics[width=0.325\linewidth,trim={0.5cm 2cm 0.5cm 0.5cm},clip]{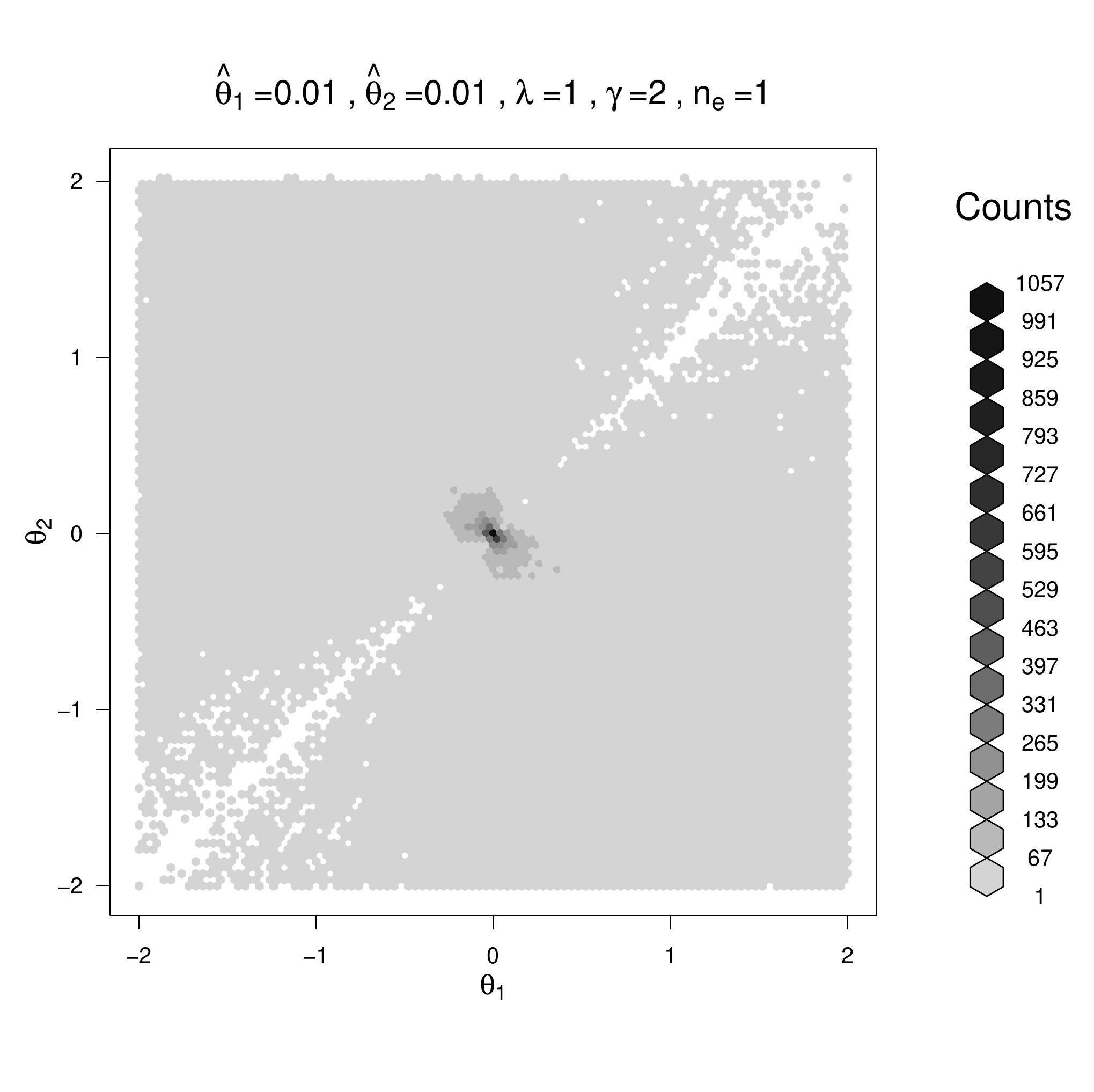}
\includegraphics[width=0.325\linewidth,trim={0.5cm 2cm 0.5cm 0.5cm},clip]{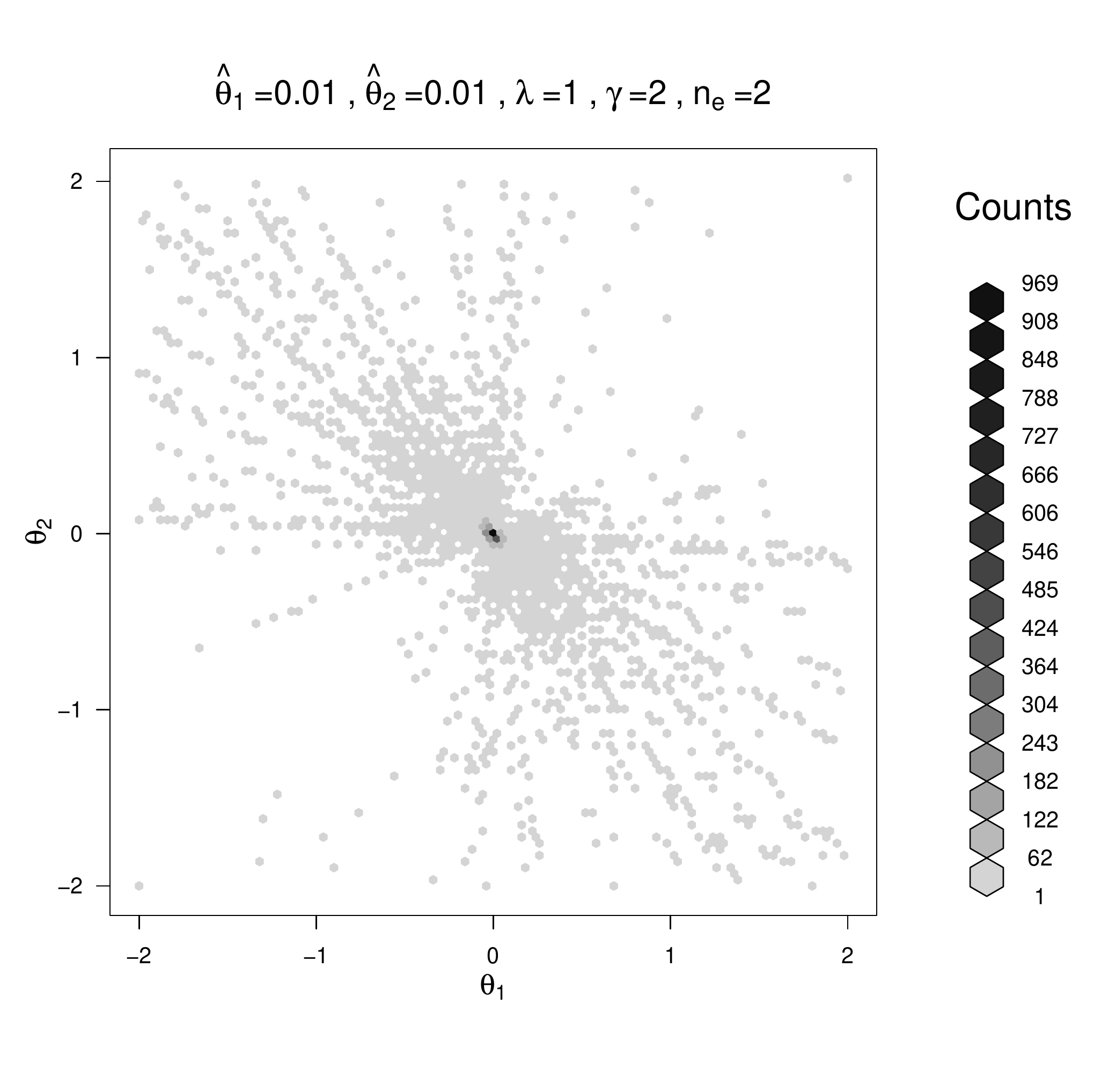}
\includegraphics[width=0.325\linewidth,trim={0.5cm 2cm 0.5cm 0.5cm},clip]{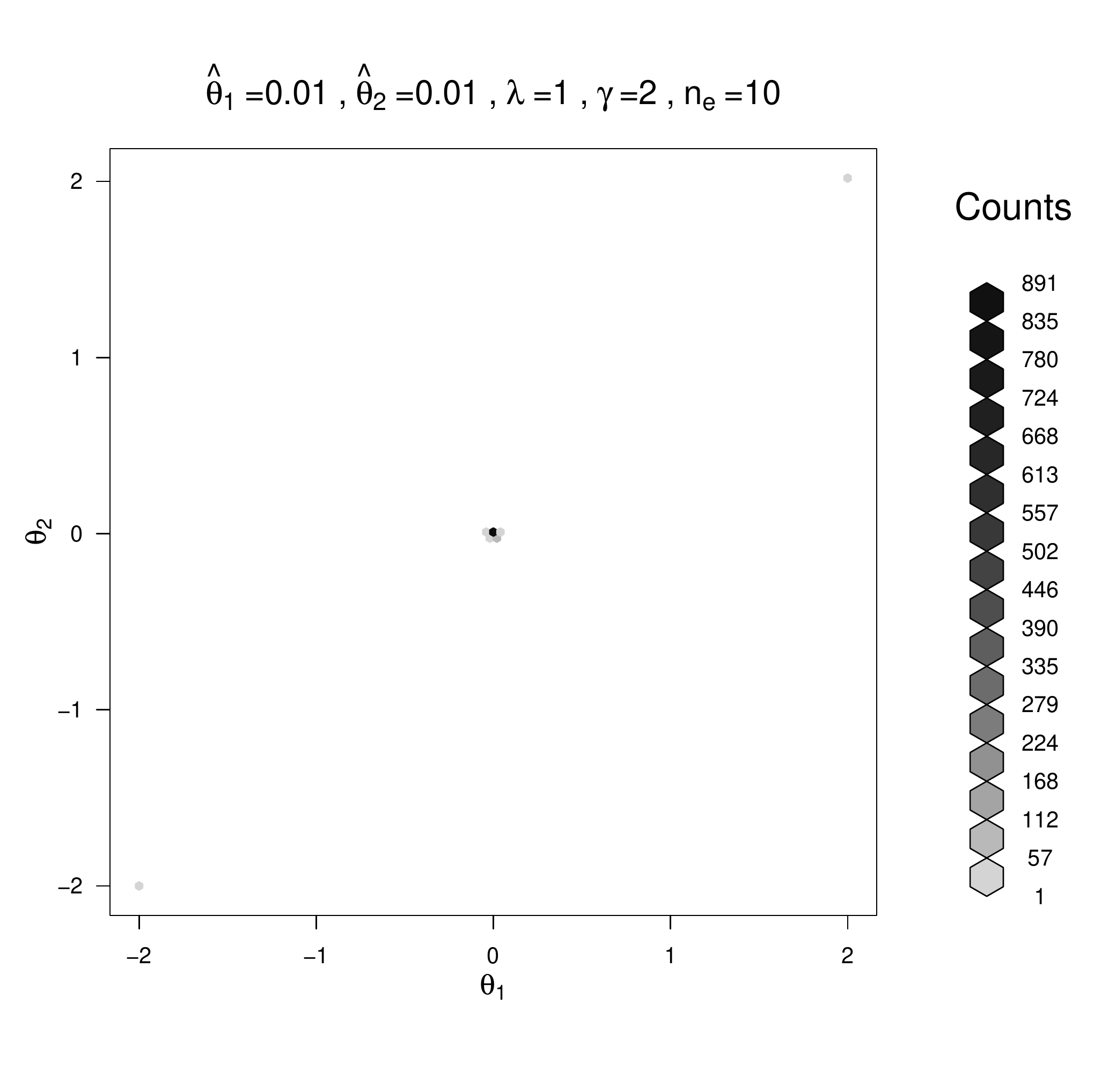}
\end{minipage}
\caption{Heat maps of the $l_0$ constraint through the orthogonal regularization in PANDA with fixed $n_e$ for 2-dimensional $\bs\theta=(\theta_1,\theta_2)$}
\label{fig:orthregconstraint}
\vspace{-12pt}
\end{figure}

\begin{pro}[\textbf{orthogonal regularization effect of PANDA for GLM with fixed $n_e$}] \label{prop:glmorthogonalregularization} 
With fixed $n_e$ and the approximate loss function in Eqn (\ref{eqn:GLMloss2}), 
PANDA estimates $\bs\theta$ by solving 
\begin{align}
\hat{\bs\theta}=\textstyle \arg\min_{\bs\theta}l_p(\bs\theta|\tilde{\x}) \approx \arg\min_{\bs\theta} (l(\bs\theta|\x)+C_1\sum_{i=1}^{n_e}
\left(\e^T_i\bs\theta\right)^2 \label{eqn:glmapproxregularization}
\end{align} 
in each iteration, which is
conceptually equivalent to the constrained optimization problem
\begin{align}
& \textstyle \mbox{min }l(\bs\theta|\x) \mbox{ subject to } \notag\\ 
&\textstyle  \sum_{i=1}^{n_e}(\e_i^T\bs\theta)^2\leq \sum_{i=1}^{n_e} \left(\e_i^T\hat{\bs\theta} \right)^2\mbox{ or equivalently, } \label{eqn:glmorthogonalregularization0}\\
& \textstyle \exists\; 0<d_i<(\sum_{i=1}^{n_e}(\e_i^T{\hat{\bs\theta}})^2 )^{1/2}\mbox{ such that } |\e_i^T\bs\theta|\leq d_i, \mbox{ for } i=1,\ldots,n_e.\label{eqn:glmorthogonalregularization1}
\end{align} \vspace{-21pt}
\end{pro}
$\hat{\bs\theta}$ in Eqns  (\ref{eqn:glmorthogonalregularization0}) and (\ref{eqn:glmorthogonalregularization1}) is the solution from Eqn (\ref{eqn:glmapproxregularization}). 
Proposition \ref{prop:glmorthogonalregularization} suggests that the (unconstrained) optimization problem PANDA solves in each iteration is equivalent to a constrained optimization problem with $n_e$ linear constraints on $\bs\theta$.  When $n_e< p$, the $n_e$ constraints in Eqn (\ref{eqn:glmorthogonalregularization1}) only affect a subset of the $p$ parameters. 
For the $l_0$ penalty ($\gamma=2$ in Eqn \ref{eqn:bridge})) and when $\lambda$ is large, the constraints take effect on exactly $n_e$ parameters. In other words,
the following two optimization problems are equivalent.
\begin{align}
\mbox{Problem 1: } & \ \bar{\bs\theta}=\E_\e(\hat{\bs\theta})=\E_\e\big\{\!\textstyle\arg\min_{\bs\theta} ( l(\bs\theta|\x)+C_1\sum_{i=1}^{n_e} \big(\sum_j\theta_j e_{ij} \big)^2\big\}\\
\mbox{Problem 2: } & \ \hat{\bs\theta}=\textstyle\arg\min_{\bs\theta} l(\bs\theta|\x), \mbox{subject to } \sum_{j=1}^{p}\1(\theta_j\neq 0)= p-n_e \label{eqn:l0orthogonalregularization}
\end{align}
Figure \ref{fig:orthregconstraint} plots the heat maps of the constrained region on  $\bs\theta=(\theta_1,\theta_2)$ as suggested by Eqn (\ref{eqn:glmorthogonalregularization0}) for $n_e=1,2$ and $10$, respectively when $\hat{\bs\theta}=(0.01, 1)$ (the upper panel) and $\hat{\bs\theta}=(0.01, 0.01)$ (the bottom panel), with the $l_0$ penalty.  Specifically, each heat map is made of $50,000$ ``dots'' uniformly distributed in the $[2,2]\times[2,2]$ solution region (for plotting purposes, we focus on the region of $[2,2]^2$; in theory, the region can be as large as $(-\infty,\infty)^2$). The relative density of a particular constraint on $\bs\theta$ out the $5,000$ repeats  is proportional to the grayness of the dot. 
In the upper panel, with $n_e=1$, the chance of constraining $\theta_1$ at 0 is much higher than at any non-zero values. As $n_e$ increases from 1 to 2 to 10, the constrained region  for $\bs\theta$ shrinks (to 0 for $\theta_1$ and to within $[-1,1]$ for $\theta_2$). In the bottom panel, setting $n_e=1$ still lead a substantial chance of getting non-zero $\bs\theta$. As $n_e$ increases to 2 and 10, the chance of $\bs\theta= 0$ drastically increases and is almost certain at $n_e=10$. 

As $n_e$ further increases, the regularization effect moves away from promoting the orthogonality between $\e_x$ and $\theta$ to focusing more on the individual parameter regularization, and eventually converges to those in Proposition \ref{prop:glmregularization}. In practice, $n_e$ can be pre-specified if users have prior knowledge on the sparsity of $\bs\theta$, otherwise be regarded as a tuning parameter, chosen by the CV procedure or an information criterion (AIC or BIC) for model selection. 

\subsection{Algorithmic Steps of PANDA}\label{sec:algorithm}\vspace{-3pt}
The practical implementation of PANDA starts with some initial values for $\bs{\theta}$. The estimates of $\bs{\theta}$ and the variance terms of the pre-specified NGD are updated iteratively until convergence. The detailed steps are listed in Algorithm \ref{alg:glm}, along with some remarks on specifying the algorithmic parameters and convergence criterion (Remarks \ref{rem:converge} to \ref{rem:nonconvex}).
\begin{algorithm}[!htb]
\caption{PANDA for GLM}\label{alg:glm}
\begin{algorithmic}[1]
\STATE \textbf{Input}:  initial parameter estimates $\bar{\bs{\bs\theta}}^{(0)}\!=\!\big(\bar{\bs\theta}^{(0)}_1,\ldots,\bar{\bs\theta}^{(0)}_p\big)$; NGD; maximum iteration $T$;  noisy data size $n_e$ and moving average (MA) window width $m$ and number of banked parameter estimates  $r$ after convergence;  threshold $\tau_0$ .
\STATE \textbf{Output}: regularized parameter estimates $\hat{\bs\theta}=(\hat{\bs\theta}_1,\ldots,\hat{\bs\theta}_p)$
\STATE Centerize the observed independent variables $\x$. 
\STATE $t\leftarrow 0$; convergence $\leftarrow 0$
\STATE\textbf{While} $t<T$ and convergence $= 0$
\begin{ALC@g} 
\STATE $t\leftarrow t+1$
\STATE Generate $\e_x^{(t)}$ from NGD N$\!\big(0, \V\!\big(\bar{\bs\theta}^{(t-1)}\big)\big)$ and set $\e_y\!\equiv\!\bar{y}$ ($\sim$Bern($\hat{p}$) for Bernoulli $Y$).
\STATE Combine $(\y,\x)$ with $(\e^{(t)}_y,\e^{(t)}_x)$ to obtain the augmented data $(\tilde{\y}^{(t)},\tilde{\x}^{(t)})$ 
\STATE Run GLM on $(\tilde{\y}^{(t)},\tilde{\x}^{(t)})$ and obtain MLE $\hat{\bs{\bs\theta}}^{(t)}$. For linear regression with Gaussian $Y$, ordinary least-squares estimates are obtained.
\STATE If $t>m$, calculate MA $\bar{\bs{\theta}}^{(t)}=m^{-1}\sum_{l=t-m+1}^t \hat{\bs{\theta}}^{(l)}$; otherwise $\bar{\bs{\theta}}^{(t)}=\hat{\bs{\bs\theta}}^{(t)}$. Calculate the loss function $l^{(t)}$  with $\bar{\bs\theta}^{(t)}$  plugged in Eqn (\ref{eqn:GLMloss}).
\STATE Calculate the averaged loss function $\bar{l}^{(t)}=m^{-1}\!\sum_{l=t-m+1}^t l^{(l)}$.
\STATE Let convergence $\leftarrow 1 $ if $\bar{l}^{(t)}$ satisfies one of the convergence criteria (Remark \ref{rem:converge}).
\end{ALC@g} 
\STATE\textbf{End While}
\STATE Run Lines 6 and 9 above for additional $m+r$ iterations, and record $\bar{\bs\theta}^{(l)}$ for $l=t\!+m+\!1,\ldots,t\!+m+\!r$. Let $\bar{\bs{\bs\theta}}\!=\! (\bar{\bs\theta}_1,\ldots,\bar{\bs\theta}_p)$, where $\bar{\bs\theta}_j\!=\!\big(\bar{\bs\theta}_j^{(t+m+1)},\ldots,\bar{\bs\theta}_j^{(t+m+r)}\big)$ for $k\!=\!1,\ldots,p$.
\STATE  Set $\hat{\theta}_j\!=\!0$ if $\max\{|\bar{\bs\theta}_j|\}<\tau_0$; and  $\hat{\theta}_j\!=\!r^{-1}\!\sum_{l={t+m+1}}^{t+m+r}\bar{\bs\theta}_j^{(l)}$ o.w.
\end{algorithmic}
\end{algorithm}
\begin{rem}[\textbf{convergence criterion}]\label{rem:converge}
We provide several choices to evaluate the convergence of the PANDA algorithm. First, we may eyeball the trace plots of  $\bar{l}^{(t)}$, which is often sufficient. Second, we can apply a cutoff value, say $\tau$ on the absolute percentage change on $\bar{l}^{(t)}$ from two consecutive iterations: if $|\bar{l}^{(t+1)}-\bar{l}^{(t)}|/\bar{l}^{(t)}<\tau$, then we may declare convergence.  $\tau$ is supposed to be close-to-0 upon convergence, but being arbitrarily close to 0 would be difficult to achieve given the fluctuation around  $\bar{l}^{(t)}$  with finite $m$ or $n_e$ due to the randomness of the augmented noises from iteration to iteration. 
Finally, we develop a formal statistical test for convergence based on $\bar{l}^{(t)}$; but the test should be used with caution as it tends to claim non-convergence. The details of the test are provided in Sec \ref{sec:convergence} of the supplementary materials. 
\end{rem}
\begin{rem}[\textbf{maximum iteration  $T$}]\label{rem:T}
$T$ should be set at a number large enough so to allow the algorithm to reach  convergence within a reasonable time period. When $n_e$ is large, we expect the algorithm to converge with a relatively small $T$ (in the examples in Sec \ref{sec:examples}, convergence is achieved for $T\le20$ with a large $n_e$). If $n_e$ is small, especially when PANDA is used to realize the $l_0$ regularization, $T$ should be set a large number for convergence.
\end{rem}
\begin{rem}[\textbf{$m$ and $r$}]\label{rem:mr}
In practical implement, $n_e$, no matter how large, is still finite. In addition, one might not want to set $n_e$ at a very large value as it will slow down the per-iteration computation. With a finite $n_e$, there is random fluctuation around the loss function and parameter estimates since each iteration is based on a different set of finite samples, even when the PANDA algorithm converges. To mitigate the random fluctuation, we can take the moving averages of the estimated parameters over multiple ($m$) iterations. The same rationale applies to the banking of $r$ estimates after convergence. In addition, taking the averages of the estimates obtained from the multiple augmented data sets also leads to a small generalization error due to the ensemble-learning type of effect PANDA brings (see Sec \ref{sec:ensemble} for more details). In our empirical studies, $r= O(10)$ seems to be sufficient.
\end{rem}
\begin{rem}[\textbf{Bounding at $\tau_0$}]\label{rem:tau0}
The bounding at $\tau_0$ is necessary.  Despite the fact that estimates of zero-valued $\theta$ can get arbitrarily  close to 0 (see Sec \ref{sec:consist} for the almost sure convergence of the minimizers in PANDA),  being exactly 0  cannot be achieved computationally in practice due to the numerical nature of PANDA. In addition, after the convergence of the PANDA algorithm, there is still mild fluctuation around the parameter estimates due to the randomness of the augmented noise, especially when  $n_e$ or $m$ is not large. We suggest bounding the absolute maximum of the estimates over a sequence of iterations as given in Algorithm \ref{alg:glm}, which seems to be a robust criterion in the empirical studies in Sec \ref{sec:examples}. 
\end{rem}
\begin{rem}[\textbf{non-convex regularizers}]\label{rem:nonconvex}
PANDA optimizes a convex objective function in each iteration of the GLM on the augmented data even when the targeted regularizer itself is non-convex, such as the SCAD or $l_0$. As such, PANDA does not run into the same type of computational difficulties that gradient-based techniques often experience  for non-convex optimization (e.g., getting stuck in a local optimum). As a matter of fact, due to the stochastic nature of PANDA from iteration to iteration, it can escape from a local optimum especially if it is unstable, and lands at a more stable local optimum or even the global optimum. That being said,  the initial values used in PANDA would also affect the final solutions when the targeted regularizer is non-convex. 
\end{rem}

\vspace{-6pt}\subsection{\texorpdfstring{$n_e$ vs. $m$}{}}\label{sec:nem}\vspace{-3pt}
Upon convergence, the expected regularization in Proposition \ref{prop:glmregularization}  can be realized either by letting  $m\rightarrow \infty$ suggested by $\lim_{m\rightarrow\infty} m^{-1}\sum_{t=1}^m\!
\sum_{i=1}^{n_e}\!\big(e^{(t)}_i\!-\!\sum_je^{(t)}_{ik}\theta_j\big)^2$ or  by letting  $n_e\rightarrow\infty$ suggested by $n_e\lim_{n_e\rightarrow\infty} n_e^{-1}\!
\sum_{i=1}^{n_e}\big(e^{(t)}_i\!-\!\sum_je^{(t)}_{ik}\theta_j\big)^2$ under the constraint $n_e\mbox{V}(e_j)=O(1)\;\forall\; \theta_j$. The constraint $n_e\mbox{V}(e_j)=O(1)$ guarantees that injected noise $\e$ does not over-regularize or overwhelm the information about $\bs\theta$ contained in the observed data $\x$ even when $n_e$ is large. For example, $\mbox{V}(e_j)=\lambda|\theta_j|^{-1}$ in the case of the lasso-type noise, and $n_e\lambda$ can be treated  as one tuning parameter. 
The  targeted regularization implied by the lower-order term $n_e\!\big(C_{1}\!\sum_j\theta^2_j\mbox{V}(e_j)\big)$ in Eqn (\ref{eqn:glmregularization}) can be approximated arbitrarily well as $n_e\rightarrow\infty$ with $n_eV(e_j)=O(1)$. When $m\rightarrow\infty$ and $n_e$ is fixed, there exists, more or less, other type of regularization on $\bs\theta$ on top of the targeted regularization given that the higher-order term $O\big(\sum_j\!\left(\theta_j^4n_e\V^2(e_j)\!\right)\!\big)$ in Eqn (\ref{eqn:glmregularization}) does not disappear. If we also require $n_eV(e_j)=O(1)$ in the large $m$ and small $n_e$ case, then $O\big(\sum_j\!\big(\theta_j^4n_e\V^2(e_j)\!\big)\!\big)= O\big(\sum_j\!\big(\theta_j^4\V(e_j)\!\big)\!\big)$, then the high-order term would also be ignorable if $\theta^4_jV(e_j)$ is small. 

\begin{figure}[!htb]
\centering
\includegraphics[width=0.7\linewidth]{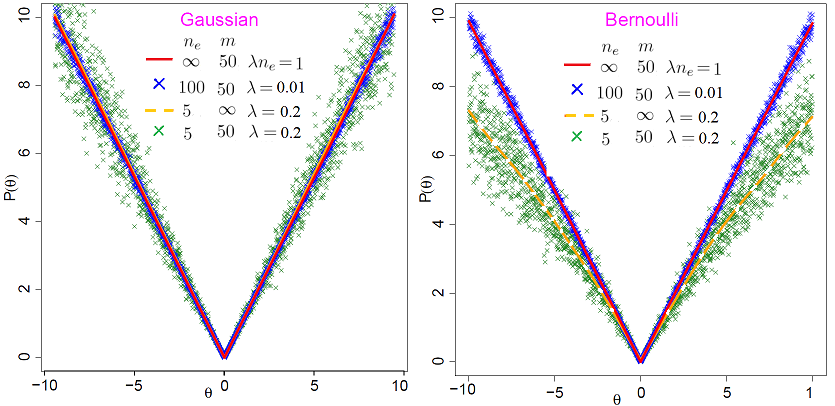}
\includegraphics[width=0.7\linewidth]{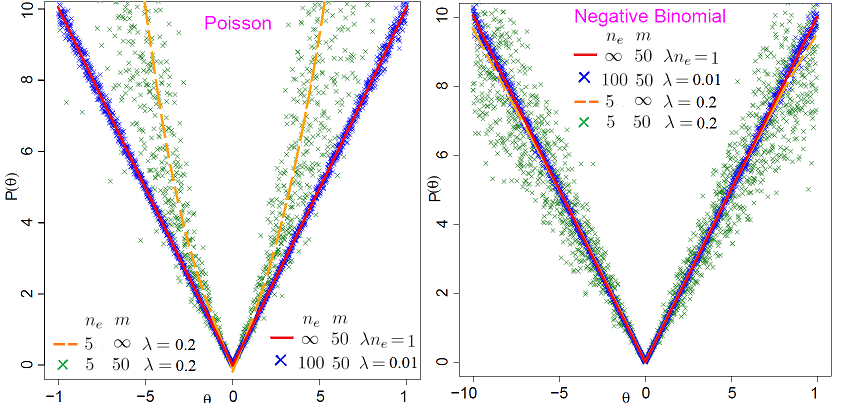}
\caption{Realized regularization by PANDA in different GLMs for the targeted penalty $P(\bs\theta)=|\bs\theta|$.} 
\label{fig:regeffect}
\vspace{-9pt}
\end{figure}

Figure \ref{fig:regeffect}  illustrates the differences between the realized regularization effect $P(\bs\theta)$, when the targeted regularization is lasso ($P(\bs\theta)=|\bs\theta|$),  by letting $n_e\rightarrow\infty$ ($m$ is small)  vs $m\rightarrow\infty$ ($n_e$ is small) and its relationships with $\bs\theta$ for several types of GLM (the regularization effects when $Y$ follows an exponential distribution are similar to when $Y$ is Poisson and the results from the former are not provided). For $n_e\rightarrow\infty$ ($\lambda n_e=1$ fixed at 1 and $m=50$), the realized penalty is identical to the targeted lasso in all four regression types, and is very close lasso at $n_e=100$ except for some very mild random fluctuation. The realized regularization on $\bs\theta$ at $m\rightarrow\infty$ and small $n_e$ varies by regression type. When $|\bs\theta|$ is small, the target regularization is realized as the higher-order term that involves $|\bs\theta|$ in Eqn (\ref{eqn:glmregularization}) is ignorable. As $|\theta|$ increases, the the higher-order  term becomes less ignorable and regularization deviates from lasso, except for linear regression where the higher-order term is analytically 0. Specifically, the realized regularization is sub-linear for logistic and  NB regression, and super-linear for Poisson regression.

\begin{figure}[!htb]
\begin{center}
\begin{minipage}{0.4\textwidth}
\includegraphics[width=1\linewidth]{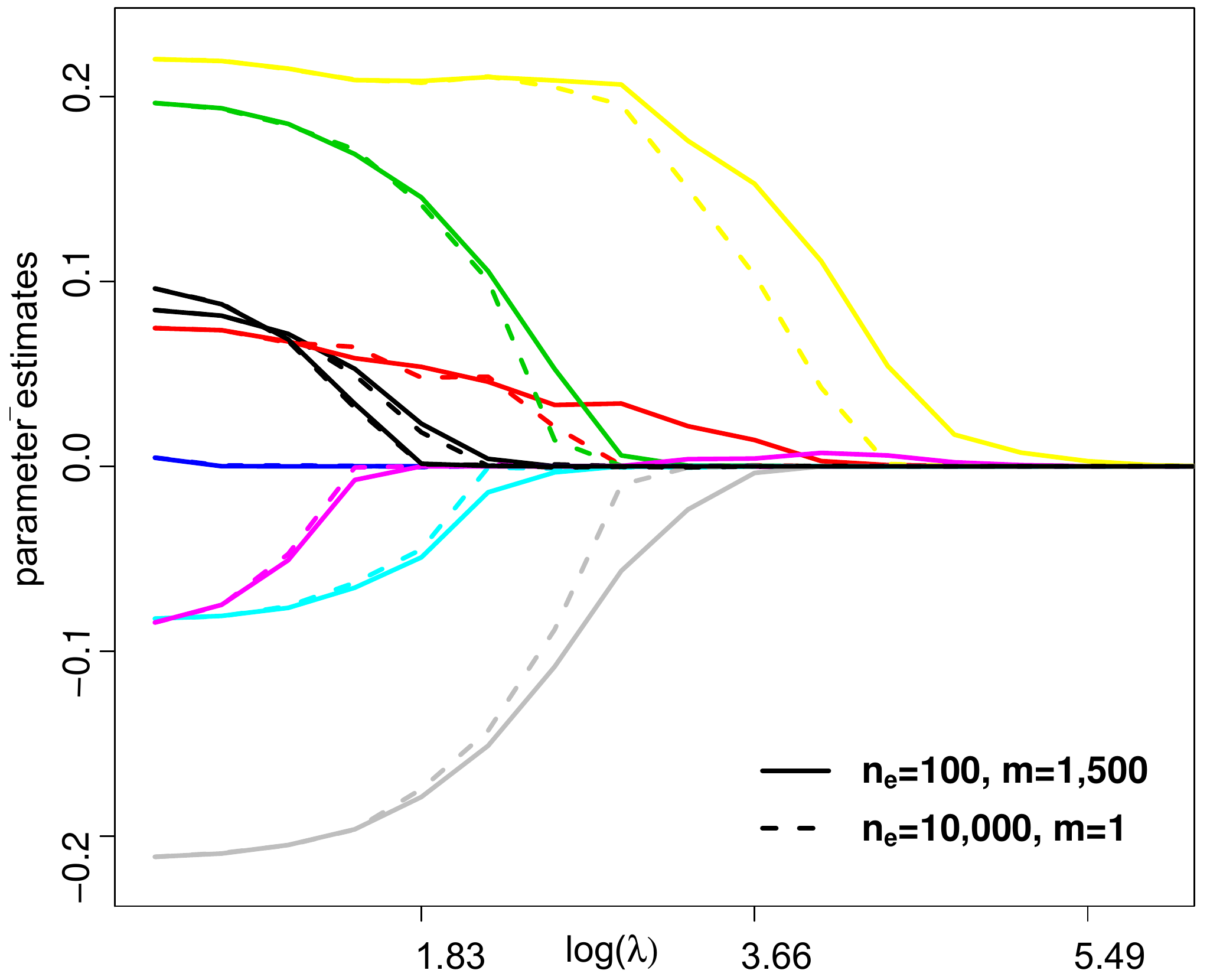}
\end{minipage}
\begin{minipage}{0.4\textwidth}
\includegraphics[width=1\textwidth]{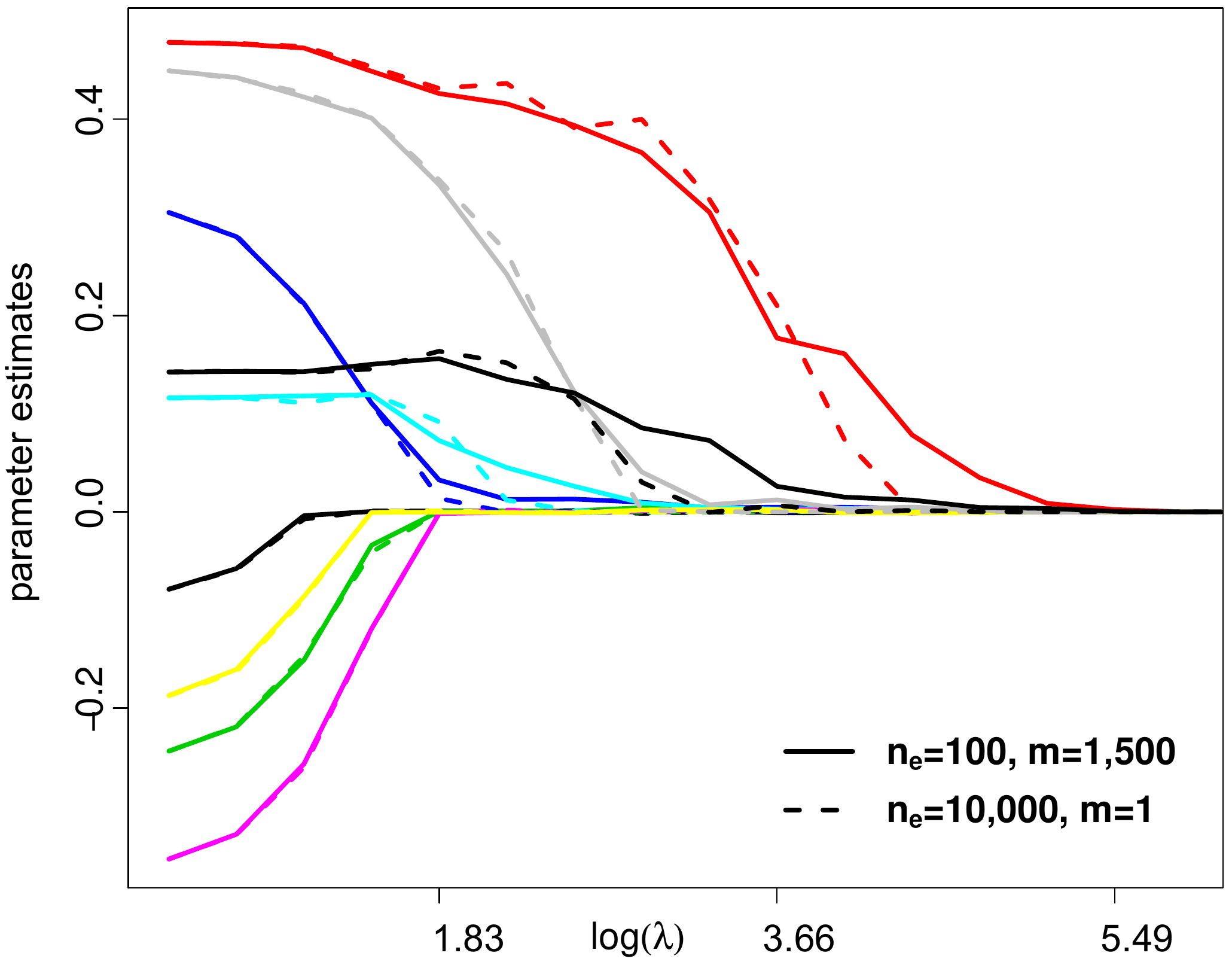}
\end{minipage}
\end{center}\vspace{-12pt}
\caption{Estimation trajectories of zero-valued regression coefficients across $\lambda$  in linear regression (left) and Poisson regression (right) with lasso-type noise in PANDA}\label{fig:trace}
\vspace{-9pt}
\end{figure}

In summary, to achieve the expected regularization effect in Proposition \ref{prop:glmregularization}, one can set either $m$ or $n_e$ at a large number. Computationally, a large $n_e$ often requires less iterations even when $m$ is as small as 1. 
On the other hand, a very large $n_e$ slows down the computation per iteration. Taken together, the actual time taken to reach convergence might not differ that much between the two cases.  
In some sense, the choices on $n_e$ and $m$ more or less depends on each other. If a large $n_e$ still results in noticeable fluctuation around  $\hat{\bs\theta}$, then a large $m$ can be used to speed up the convergence. For a small $n_e$, a relatively large $m$ should be used to yield stable penalty.  Fig \ref{fig:trace} shows the parameter estimation trajectories of zero-valued regression coefficients across   with $\lambda$ in linear regression  and Poisson regression on simulated data when the lasso-type noise is used in PANDA
There are 30 predictors ($p=30$) and $n=100$ in each case. In the linear regression, the predictors were simulated from N$(0,1)$; in the Poisson regression, the predictors were simulated from Unif$(-0.3,0.5)$. Out of the 30 coefficients, 9 were set at 0, and the other 21 non-zero coefficients ranged from 0.5 to 1. The estimation trajectories for the 9 zero-valued parameters look very similar between large $n_e$ vs. large $m$ in both regression settings. 

If the targeted penalty is $l_0$ (Proposition \ref{prop:glmorthogonalregularization}) and $n>p$, $n_e$ can be tuned within $[1,p]$. There are other considerations regarding the choices of $m$ and $n_e$ when using PANDA to obtain inference on $\bs\theta$. More details are provided in Sec \ref{sec:asymp.dist}.

\section{Theoretical Properties and Statistical Inferences}\label{sec:theory}\vspace{-3pt}
In this section, we establish the almost sure (a.\ s.) convergence of the data augmented loss function to its expectation and the a.\ s.\ convergence of the minimizer of the former to the minimizer of the expected loss function  as $n_e\rightarrow\infty$ or $m\rightarrow\infty$  (Sec \ref{sec:consist}). We also examine the Fisher information of the parameters in noise-augmented data (Sec \ref{sec:fisher}) and statistical inferences of the parameters via PANDA (Sec \ref{sec:asymp.dist}), and claim that PANDA exhibits ensemble learning behavior (Sec \ref{sec:ensemble}).

\vspace{-3pt}\subsection{Almost sure convergence of noise augmented loss function and its minimizer} \label{sec:consist}\vspace{-3pt}
Let $\bar{l}_p(\bs\theta|\tilde{\x},\tilde{\y})$ denote the average loss function over $m$ iterations of the PANDA algorithm upon convergence. Theorem \ref{thm:as} presents the asymptotic properties of $\bar{l}_p(\bs\theta|\tilde{\x},\tilde{\y})$ under two scenarios: 1) $n_e\rightarrow\infty$ while $n_e\V(e_j)=O(1)$ for a given $\theta_j$ and $m\;(\ge 1)$ is fixed at a constant; 2) $m\rightarrow\infty$ when $n_e (>p)$ takes a finite constant. 
\begin{thm}{\textbf{(asymptotic properties of the noise-augmented loss function and its minimizer for PANDA)}}\label{thm:as} 
Assume $\bs\theta$ belongs to a compact set. Let $l_p(\bs\theta|\x)=\E_{\e}(l_p(\bs\theta|\tilde{\x},\tilde{\y}))$. \\
1) If $n_e\rightarrow\infty$ while $n_e\V(e_j)=O(1)$ for any given $\theta_j$ and $m\ge1$ is held at a constant, then
\begin{align}
n_e^{1/2}C_1^{-1}\left(\bar{l}_p(\bs\theta|\tilde{\x},\tilde{\y})-l_p(\bs\theta|\x)\right)&\overset{d}{\longrightarrow} N(0,1)\label{eqn:d1}\\
\bar{l}_p(\bs\theta|\tilde{\x},\tilde{\y})&\overset{a.s.}{\longrightarrow}l_p(\bs\theta|\x)\!\overset{n_e\rightarrow \infty}{\longrightarrow}\!\textstyle l(\bs\theta|\x)\!+P(\bs\theta)+C\label{eqn:as1l}\\
\arg\inf\limits_{\bs\theta}\bar{l}_p(\bs\theta|\tilde{\x},\tilde{\y})&\overset{{a.s.}}{\longrightarrow}\arg\inf\limits_{\bs\theta}l_p(\bs\theta|\x),\label{eqn:as1argmin}
\end{align}
2) If $m\rightarrow\infty$ while  $n_e (>p)$ is fixed, then 
\begin{align}
m^{1/2}C_2^{-1}\left(\bar{l}_p(\bs\theta|\tilde{\x},\tilde{\y})-l_p(\bs\theta|\x)\right)&\overset{d}{\longrightarrow}N(0,1)\label{eqn:d2}\\
\bar{l}_p(\bs\theta|\tilde{\x},\tilde{\y})&\overset{a.s.}{\longrightarrow}l_p(\bs\theta|\x)
\overset{m\rightarrow \infty}{\longrightarrow}
l(\bs\theta|\x)+ P(\bs\theta)+C\label{eqn:as2l}\\
\arg\inf\limits_{\bs\theta}\bar{l}_p(\bs\theta|\tilde{\x},\tilde{\y})&\overset{{a.s.}}{\longrightarrow}\arg\inf\limits_{\bs\theta}l_p(\bs\theta|\x).\label{eqn:as2argmin}
\end{align}
$P(\bs\theta)$ in Eqns (\ref{eqn:as1l}) and 
(\ref{eqn:as2l}) is the same as defined in Proposition \ref{prop:glmregularization}. $C_1$ and $C_2$ are functions of $\bs\theta$ and take different forms for different types of $Y$.
\end{thm}\vspace{-6pt}
The proof of Theorem \ref{thm:as} is provided in Sec \ref{app:pelf2nmelf} of the supplementary materials. 
There are two important takeaways. First, Theorem \ref{thm:as} states that $\bar{l}_p(\bs\theta|\tilde{\x},\tilde{\y})$ follows a Gaussian distribution at the rate of $\sqrt{n_e}$ and $\sqrt{m}$ under the two scenarios, respectively, implying that the augmented loss function in PANDA is trainable for practical implementation. The fluctuation of $\bar{l}_p(\bs\theta|\tilde{\x},\tilde{\y})$ around its expected value due to noise augmentation  is controlled and the tail of the distribution of $d=\bar{l}_p(\bs\theta|\tilde{\x},\tilde{\y})-l_p(\bs\theta|\x)$  decays to 0 exponentially fast in $n_e$ and $m$ as $\Pr(d\!>\!t)\!\le\!\exp(-n_et^2/2C^2)$ and $\Pr(d\!>\!t)\!\le\!\exp(-mt^2/2C^2)$ for any $t\!>\!0$. 
Second, $\bar{l}_p(\bs\theta|\tilde{\x},\tilde{\y})$ converges a.s.\ to its expectation (the penalized loss function given the observed data $(\x,\y)$ with the targeted penalty term), 
guaranteeing that PANDA does what it is designed to do. 

When there exists multicollinearity among $\X$, the loss function minimized in PANDA has an optimum region rather than a single  optimum point. To examine the asymptotic properties in this case, we define  the \emph{optimum parameter set} (Definition \ref{def:optimum}) and  show that the parameters learned by PANDA fall in the optimum parameter set asymptotically (Proposition \ref{prop:consistmcl}).  
\begin{defn}{(\textbf{optimum parameter set})}\label{def:optimum} Let the expected loss function $l_p(\bs{\bs\theta}|\x)$ be a continuous function in $\bs{\bs\theta}$. The optimum set is defined as 
$\bs\Theta^0\!=\!\left\{\bs{\theta}^0\!\in\!{\bs\Theta} \mid l_p( \bs{\theta}^0|\x)\!\leq\!l_p(\bs{\theta}|\x), \forall\;\bs{\theta}\!\in\!{\bs\Theta}\right\}$, where $\bs\Theta$ is the set containing all possible parameter values. The distance from $\bs{\theta}\in\bs\Theta$ to $\bs\Theta^0$ is defined as $d\left(\bs{\bs\theta}, {\bs\Theta}^0\right)= \min\limits_{\bs{\theta}^0\in\bs\Theta^0}||\bs{\bs\theta}-\bs{\bs\theta}^0||_2$.\vspace{-6pt}
\end{defn}
\begin{pro}{\textbf{(consistency of parameter estimate in presence of multicollinearity)}}\label{prop:consistmcl}
Let $\hat{\bs{\bs\theta}}_p^0\!=\!\arg\min\limits_{\bs\theta}\bar{l}_p(\bs\theta|\tilde{\x},\tilde{\y})$.  Given
\begin{equation}\label{eqn:sup}
\!\!\!\sup\limits_{\bs\theta}\left|\bar{l}_p(\bs\theta|\tilde{\x},\tilde{\y})\!-\!\bar{l}_p(\bs\theta|\x)\right|\rightarrow0 \mbox{ as  $n_e\!\rightarrow\!\infty \bigcap n_e\V(e_j)\!=\!O(1)\; \forall j =1,...,p$ or $m\rightarrow\infty$;} \vspace{-6pt}
\end{equation}
and assume ${\bs\theta}$ is compact, then
$\Pr\left(\limsup\limits_{m\rightarrow\infty\mbox{ or }n_e\rightarrow\infty} d\big(\hat{\bs{\bs\theta}}_p^0,\bs\Theta^0\big) \leq\delta \right) =1\;\forall\; \delta>0.$\vspace{-6pt}
\end{pro}
The proof is given in  Sec \ref{app:consistmcl} of the supplementary materials. Multicollinearity does not affect the convergence of the loss functions in PANDA; therefore, Eqn (\ref{eqn:sup}) holds per the proof of Theorem \ref{thm:as}.

\subsection{Fisher Information in Noise Augmented Data}\label{sec:fisher}\vspace{-3pt}
The augmented noise in PANDA brings endogenous information to observed data $\x$ to regularize the estimation of $\bs\theta$. The expected regularization can be achieved  by letting $(n_e\rightarrow\infty) \cap (n_e\V(e_j)=O(1))$.  At the first sight,  it seems that a large amount of augmented noisy data could potentially overshadow the information about the parameters in the observed data, leading to over-regularization. We claim that this is not the case because of the constraint  $n_e\V(e_j)=O(1) \;\forall\;j$. In other words, $n_e$ combined with the tuning parameters from the NGD variance term is treated as a single tuning parameter. For example, with the lasso-type noise, $n_e\lambda$ is treated one tuning parameter: if $n_e$ is large, then $\lambda$ takes a small value so to keep $n_e\lambda=O(1)$. Proposition \ref{prop:fisher} provides the theoretical justification that, as long as $n_e\V(e_j)=O(1)$  for any given $\theta_j$, the amount of regularization brought by the augmented data to $\theta_j$ remains as constant even for  $n_e\rightarrow\infty$. Proposition \ref{prop:fisher} is established in the context of the bridge-type noise; the same conclusion can be obtained for other noise types in a similar fashion. The proof is provided in Sec \ref{app:fisher} of the supplementary materials.  
\begin{pro}[\textbf{Fisher information in noise augmented data}]\label{prop:fisher}
The regularization on the coefficients $\bs\theta$ in GLM introduced through the augmented bridge-type noise is proportional to $n_e\lambda|\bs\theta|^{-\gamma}$. Specifically, $I_{\tilde{\x},\tilde{\y}}(\bs\theta)$, the Fisher information on $\bs\theta$ contained in the augmented data $(\tilde{\x},\tilde{\y})$ is the summation of $I_{\x,\y}(\bs\theta)$, the Fisher information on $\bs\theta$ contained in the observed data, and $I_{\e}(\bs\theta)$, the amount of regularization on  $\bs\theta$.
\begin{equation}
\textstyle {I_{\tilde{\x},\tilde{\y}}}(\bs\theta)= {I_{\x,\y}}(\bs\theta)+(\lambda n_e){B}''(\theta_0+0) \mbox{Diag}\{|\theta_1|^{-\gamma},\ldots,|\theta_p|^{-\gamma}\} +O\big(\lambda n_e^{1/2}\big)J,\label{eqn:fisher}
\vspace{-6pt}
\end{equation}\vspace{-9pt}
\end{pro}
where $J$ is a $p\times p$ matrix with all elements at 1. The higher-order term $O\big(\lambda n_e^{1/2}\big)$ becomes $O(\lambda^{1/2})$ if $\lambda n_e=O(1)$ and is ignorable if $\lambda$ is small.  Eqn (\ref{eqn:fisher}) suggests that the information about $\bs\theta$ does not increase with $n_e$ as along as $\lambda n_e|\bs\theta|^{-\gamma}$ is kept at a constant. In addition, the closer $|\bs\theta|$ is to 0, the stronger the regularization the augmented information brings to  $\bs\theta$.

\subsection{Asymptotic Distribution of Regularized  Parameters via PANDA}\label{sec:asymp.dist}\vspace{-6pt}
Proposition \ref{prop:asymp.dist.UGM} presents the asymptotic distribution of the estimated $\hat{\bs\theta}$ via PANDA, based on which we can obtain statistical inferences for $\bs{\theta}$. The  proof is given in  Sec \ref{app:CI.UGM} of the supplementary materials. 
\begin{pro}[\textbf{asymptotic distribution of parameter estimates via PANDA}]\label{prop:asymp.dist.UGM}\vspace{-3pt}
Let $\hat{\bs\theta}^{(t)}$ denote the estimate of $\bs{\bs\theta}$ in iteration $t$ of the PANDA algorithm. The final estimate for $\bs{\bs\theta}$ is denoted by $\bar{\bs{\bs\theta}}=r^{-1}\sum_{t=1}^r\hat{\bs\theta}^{(t)}$ from $r\ge1$ iterations after convergence. Assume $n_e\V(e)=o(\sqrt{n})\;\forall\bs\theta$.
\begin{align}
\sqrt{n}(\hat{\bs{\bs\theta}}^{(t)}-\bs{\bs\theta})
&\overset{d}{\rightarrow}  N(\0,\Sigma^{(t)})
\mbox{ as } n\rightarrow\infty, \label{eqn:conditional.asym.disn}\\
\sqrt{n}(\bar{\bs{\bs\theta}}-\bs{\bs\theta})
&\overset{d}{\rightarrow} N\left( \0,\bar{\Sigma}+ \Lambda \right)\mbox{ as } n\rightarrow\infty;r\rightarrow\infty, \label{eqn:asym.disn}
\end{align} 
where $\Sigma^{(t)}= I_{\tilde{\x}^{(t)},\tilde{\y}}(\bs\theta)^{-1}I_{\x,\y}(\bs\theta)I_{\tilde{\x}^{(t)},\tilde{\x}}(\bs\theta)^{-1}$ in iteration $t$,  $\bar{\Sigma}=r^{-1}\sum_{t=1}^r \Sigma^{(t)}$, and
$\Lambda=\V(\hat{\bs\theta}^{(t)})$ is the between iteration variability of $\hat{\bs\theta}^{(t)}$.\vspace{-3pt}
\end{pro}
The regularity condition  $n_e\V(e)=o(\sqrt{n})$ takes different forms for different NGDs (e.g., for the bridge-type noise, it is $\lambda n_e=o(\sqrt{n})$).  The asymptotic variance of $\hat{\bs{\bs\theta}}^{(t)}$ involves the inverse of $I_{\tilde{\x}^{(t)},\tilde{\y}}(\bs\theta)$, which always exists given the augmented data.  Eqn (\ref{eqn:asym.disn}) suggests the overall variance on $\bar{\bs\theta}$ is the summation of two variance components, $\bar{\Sigma}$, the per-iteration variance of $\hat{\bs{\bs\theta}}^{(t)}$,  and  $\Lambda$, the between-iteration variance of $\hat{\bs{\bs\theta}}^{(t)}$.  $\bar{\Sigma}$  contains the unknown $\bs\theta$ and can be estimated by plugging in $\hat{\bs{\bs\theta}}^{(t)}$, with the caveat that the uncertainty around $\hat{\bs{\bs\theta}}^{(t)}$ is not accounted for. $\Lambda$ can be estimated by the sample variance of  $\hat{\bs{\bs\theta}}^{(t)}$ over $r$ iterations; that is, $(r-1)^{-1}\sum_{t=1}^r\big(\hat{\bs{\bs\theta}}^{(t)}-\bar{\bs\theta}\big)\big(\hat{\bs{\bs\theta}}^{(t)}-\bar{\bs\theta}\big)^T$.

In the  case of linear regression, the asymptotic distribution of $\hat{\bs{\bs\theta}}^{(t)}$ in Eqn (\ref{eqn:conditional.asym.disn})  becomes
\begin{equation}\label{eqn:conditional.asym.disn.gaussian}
\sqrt{n}(\hat{\bs{\bs\theta}}^{(t)}-\bs{\bs\theta})\overset{d}{\rightarrow} N\left(\0,\sigma^2 (\bs M^{(t)})^{-1}(\x^T\x)(\bs M^{(t)})^{-1} \right),
\end{equation}
where $\mathbf{M}^{(t)}\!=\!(\x^T\x+n_e\mbox{diag}(\V(\e))$. The asymptotic variance in Eqn (\ref{eqn:conditional.asym.disn.gaussian}) contains  the unknown $\sigma^2$, which can be estimated by  $\hat{\sigma}^2=\mbox{SSE}/(n-\nu)=\big(\y-\x\hat{\bs{\bs\theta}}^{(t)}\big)^T\big(\y-\x\hat{\bs{\bs\theta}}^{(t)}\big)/(n-\nu)$ with the degree of freedom $\nu\!=\!\mbox{tr}(\x(\bs M^{(t)})^{-1}\x^T)$.  $\hat{\sigma}^2$ converges to  $\sigma^2\chi_{n-\nu}^2$ in distribution. 

When applying PANDA to obtain inference in GLMs, we should set $n_e$ at a small number and $m$ at a large number to achieve valid inference and targeted regularization effect  simultaneously. We recommend $n_e=o(n)$ as long as $n_e+n>p$ (e.g., one order of magnitude smaller than $n$), especially when $n$ is small.  This is different from when the main goal is variable selection (except for $l_0$), regularized estimation, or prediction without uncertainty quantification, where  a large $n_e$ can be used to achieve the targeted regularization effect with fewer iterations per Proposition \ref{prop:glmregularization}. The reason is that a large $n_e$ (relative to $n$) tends to  underestimate $\bar{\Sigma}+\Lambda$, the asymptotic variance of $\bar{\bs{\bs\theta}}$, resulting in a lower-than-nominal coverage rate and an inflated type I error rate. As mentioned above, $\bar{\Sigma}=r^{-1}\sum_{t=1}^r \Sigma^{(t)}$ is estimated by plugging in $\hat{\bs\theta}\!^{\;(t)}$ for $t=1,\ldots,r$ upon convergence, pretending that it is the true parameter value and thus ignoring the uncertainty around it. Though this issue exists regardless of whether a large or a small $n_e$ is used, using  a small $n_e$ helps to re-capture this lost variability with the between-iteration variability $\Lambda$. Specifically, $\hat{\bs{\bs\theta}}\!^{\;(t)}$ is a regularized estimate 
from minimizing a loss  function summed over the data component $\x$ and a penalty term, or equivalently, a summation of the loss functions constructed from the data component $\x$ and the augmented data component $\e$ in the context of PANDA. Instead of focusing on how $\hat{\bs{\bs\theta}}\!^{\;(t)}$ changes with sample data $\x$, which is fixed throughout iterations, we quantify how it changes with $\e$.  If a large $n_e$ is used, the ignored sampling variability around $\hat{\bs{\bs\theta}}\!^{\;(t)}$ can hardly be recovered through $\Lambda$ as it is close to 0, which is easy to understand as the realized regularization effect with a large $n_e$ is close to its expectation and it is almost like solving the same analytical constrained optimization at every iteration,  leading to very similar $\hat{\bs{\bs\theta}}\!^{\;(t)}$ across iterations upon convergence.

\subsection{Ensemble Learning Behavior of PANDA with Fixed \texorpdfstring{$n_e$}{}}\label{sec:ensemble}\vspace{-3pt}
Ensemble learning methods combine multiple learners to  achieve better predictive performance than that from an individual learner.  Let $Y$ be the observed outcome and $\bar{Y}$ be its prediction from an ensemble method. \citet{Brown2005ManagingEnsembles} suggest that the generalization error of the ensemble method made of $M$ learners, $E(\bar{Y}-Y)^2$, can be decomposed as
\begin{align}\label{eqn:ge}
\textstyle M^{-2}\!\left[\!\big(\sum_i(\E(\hat{Y}_i)-Y)\big)^2\!+\!\sum_i \E\big(\hat{Y}_i\!-\!E(\hat{Y}_i)\big)^2\!+\!\sum_i \sum_{j\neq i} E\big((\hat{Y}_i-\E(\hat{Y}_i))(\hat{Y}_j-E(\hat{Y}_j))\big)\!\right],
\end{align}
where $\hat{Y}_i$ refers to the prediction from the $i$-th learner in the ensemble for $i=1,\ldots, M$. The success of ensemble methods, in part, can be attributed to the  diversity term among the $M$ learners that is captured by the third term  (covariance) in Eqn (\ref{eqn:ge}): as the diversity increase, the covariance decrease, and the overall  generalization error decreases.  The diversity can be achieved by perturbing the training data such as taking a subset of observation, or a subset of attributes to train the learners.  

We show that PANDA, in addition to achieving the targeted regularization effects, also exhibits some ensemble learning behavior with a fixed $n_e$, which may propel it to edge out the existing constrained regularization approaches with smaller generalization error in prediction. Intuitively, upon convergence, the final estimates of $\bs\theta$ are averages over the estimates trained from different sets of noise augmented data from $r$ iterations, generating the diversity among the learners needed for the ensemble learning. 
\begin{cla}[\textbf{Ensemble learning behavior of PANDA with fixed $n_e$}]\label{prop:ensemblelearning}\vspace{-3pt}
Upon convergence, the average estimates over the sets of parameter estimates from multiple iterations of PANDA with a fixed $n_e$ can be regarded an ensemble learner.\vspace{-9pt}
\end{cla}
If the diversity brought by PANDA with a fixed (small) $n_e$ and a large $m$ surpasses the increase in MSE (the sum of the first two terms in Eqn (\ref{eqn:ge})), PANDA would lead to a smaller generalization error compared to the existing constrained optimization approaches for penalized GLM regression that don't promote diversity.


\section{Numerical Examples}\label{sec:examples}\vspace{-3pt}
\subsection{\texorpdfstring{$l_0$}{} Regularization via PANDA}\label{sec:l0}\vspace{-3pt} 
We demonstrate the PANDA-$l_0$ regularization  in linear regression using the prostate cancer dataset and in logistic regression using the kyphosis dataset \citep{lasso}. The prostate cancer dataset consists of 8 $X$'s and 97 observations. We standardized the $X$'s and centralized $Y$ prior to the application of the PANDA algorithm. The kyphosis dataset consists of 81 observations ($64\!:\!17$ for $Y\!=\!0\!:\!1$). We included both the linear and the quadratic terms of the three standardized  $X$'s   ($X_4,X_5,X_6$ are the quadratic terms of  $X_1, X_2,X_3$, respectively), in the logistic regression following \citet{lasso}.  We examine the regression coefficient estimation trajectories as $n_e$ increases from 1 to $p$ and as  $\lambda$ increases while holding $n_e$ constant. For comparison, we also run the lasso regression in each case via the R package \texttt{glmnet}. 

\begin{figure}[!htb]
\begin{minipage}{1\textwidth}
\centering A \hspace{1.8in} B \hspace{1.8in} C 
\end{minipage}
\begin{minipage}{1\textwidth}\centering
\includegraphics[trim={0cm 0cm 0.1cm 0.5cm},clip, scale=0.75]{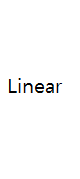}
\includegraphics[trim={0.5cm 1cm 0cm 0cm},clip, width=0.3\linewidth,height=0.28\linewidth]{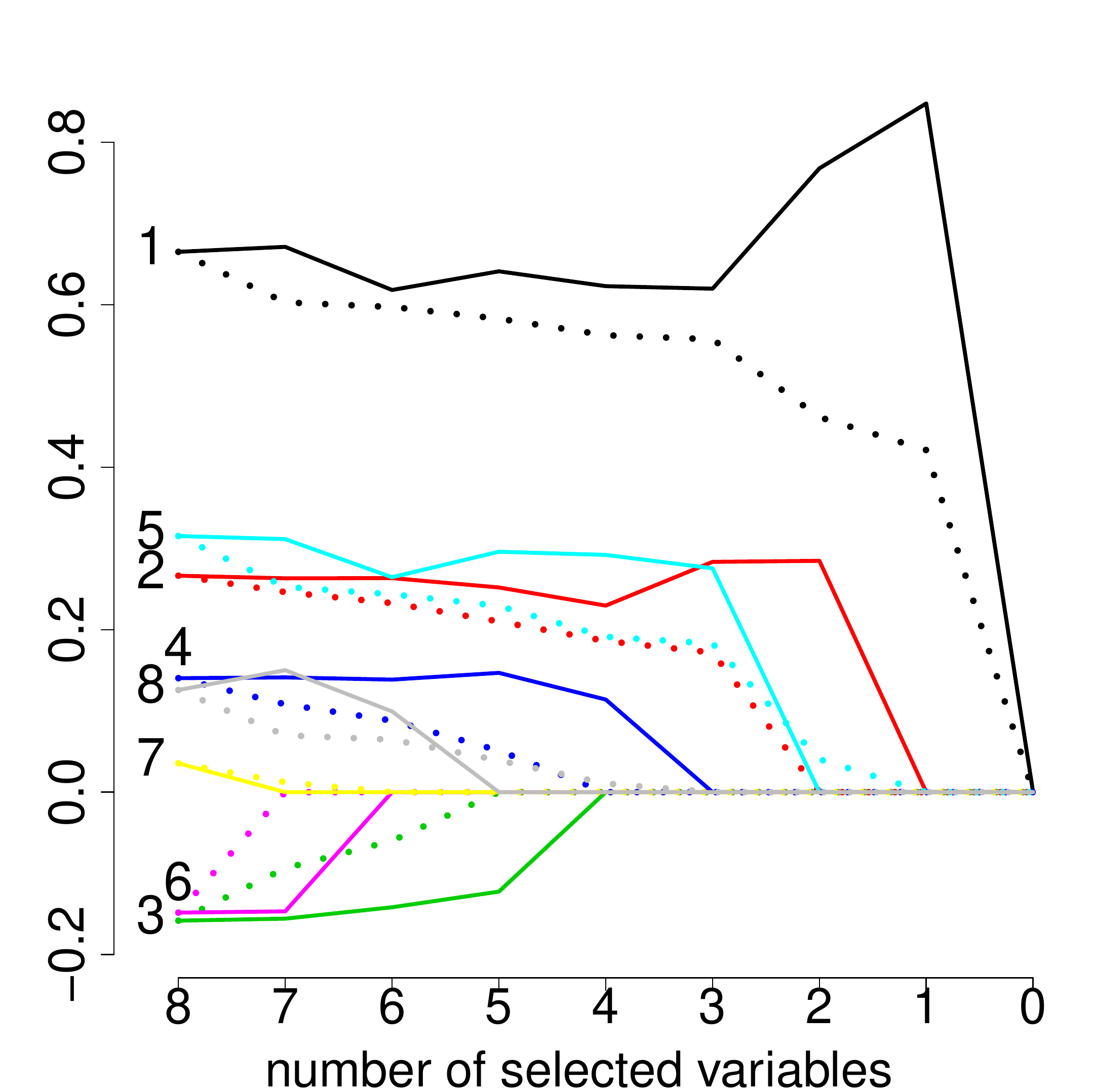}
\includegraphics[trim={0.5cm 1cm 0cm 0cm},clip,width=0.3\linewidth,height=0.28\linewidth]{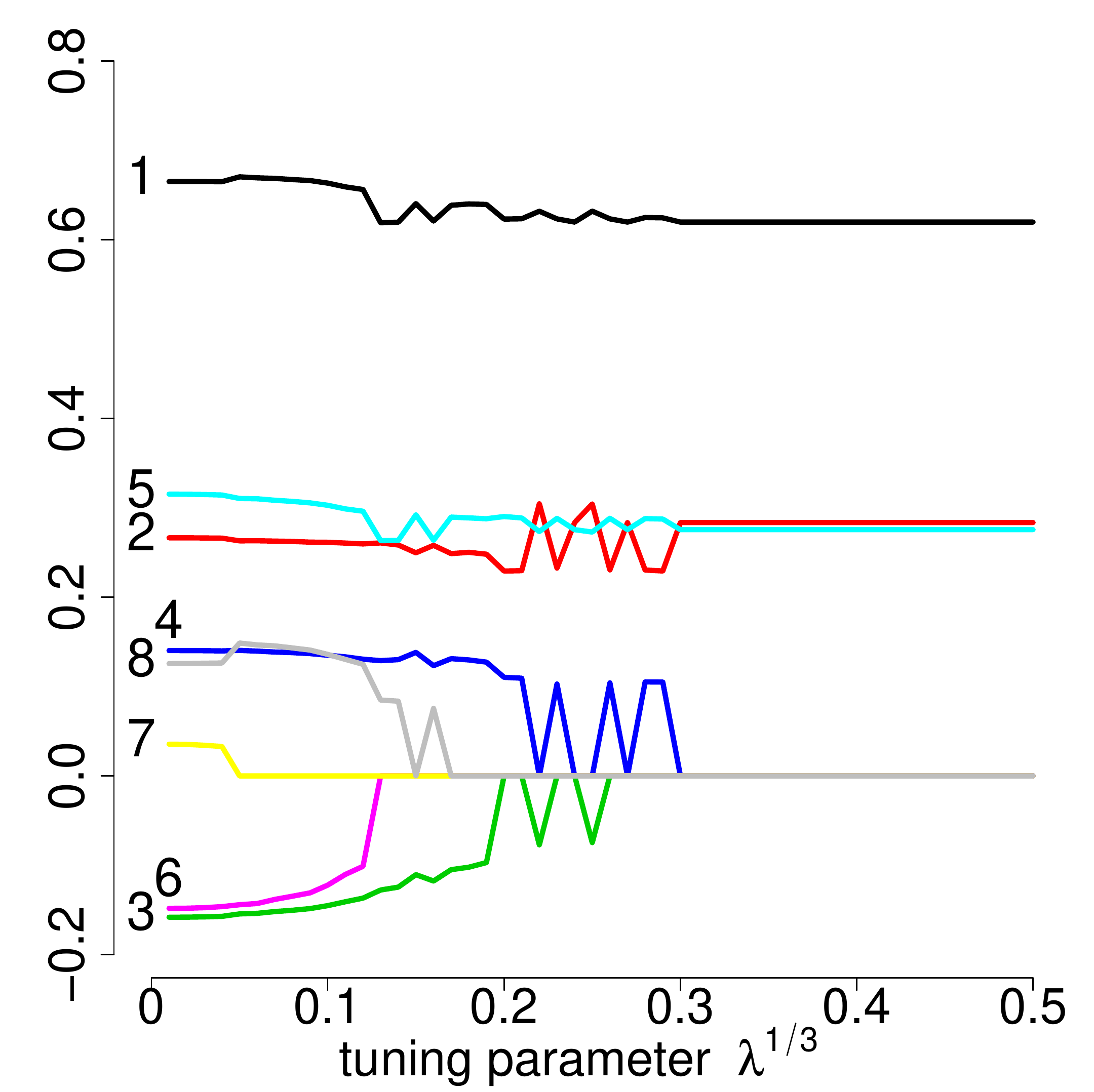}
\includegraphics[trim={0.5cm 1cm 0cm 0cm},clip,width=0.3\linewidth,height=0.28\linewidth]{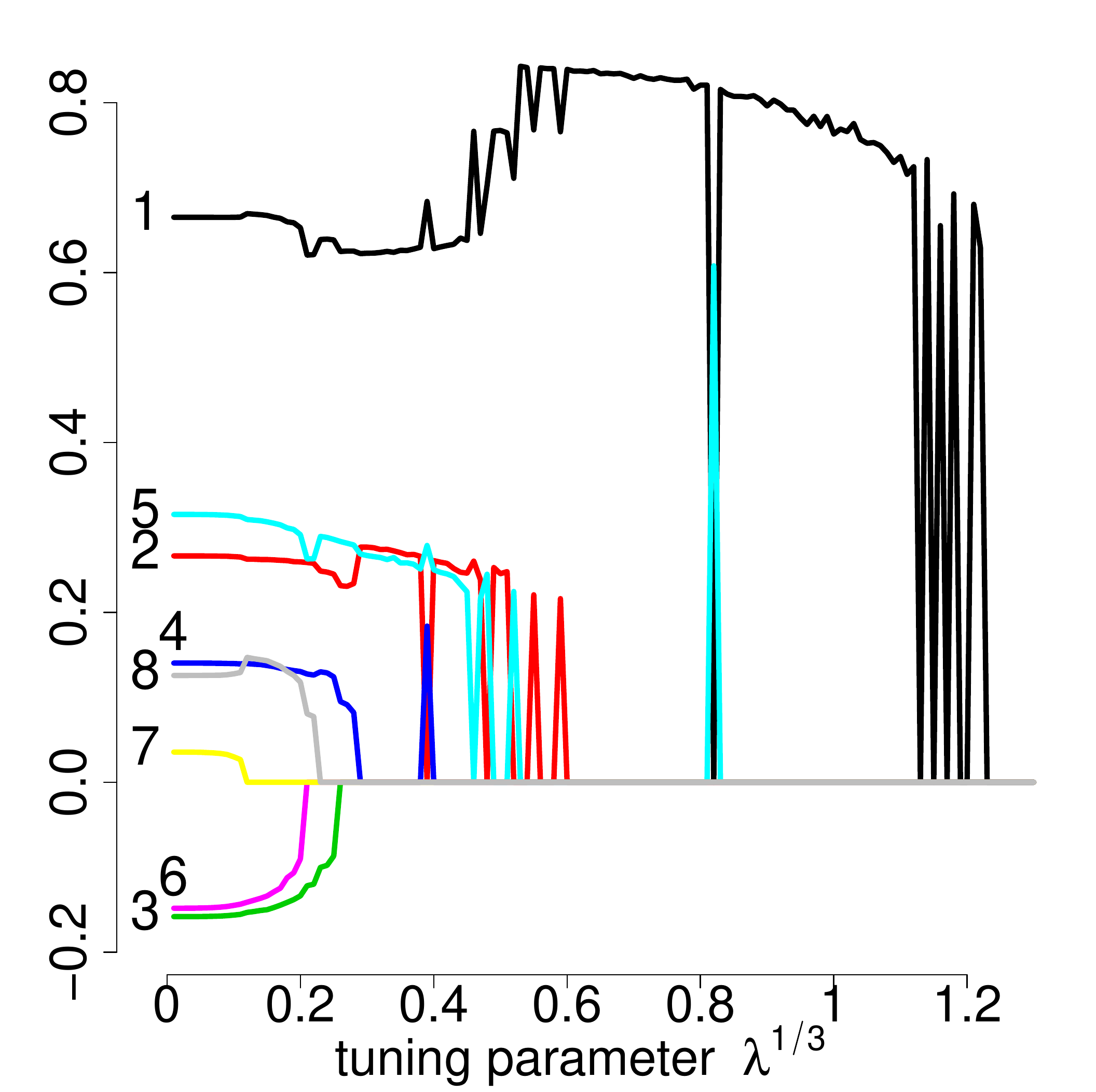}
\end{minipage}
\begin{minipage}{1\textwidth}\centering
\includegraphics[trim={0cm 0cm 0.1cm 0.5cm},clip, scale=0.75]{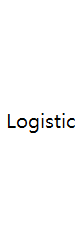}
\includegraphics[trim={0.5cm 1cm 0cm 0cm},clip,width=0.3\linewidth,height=0.28\linewidth]{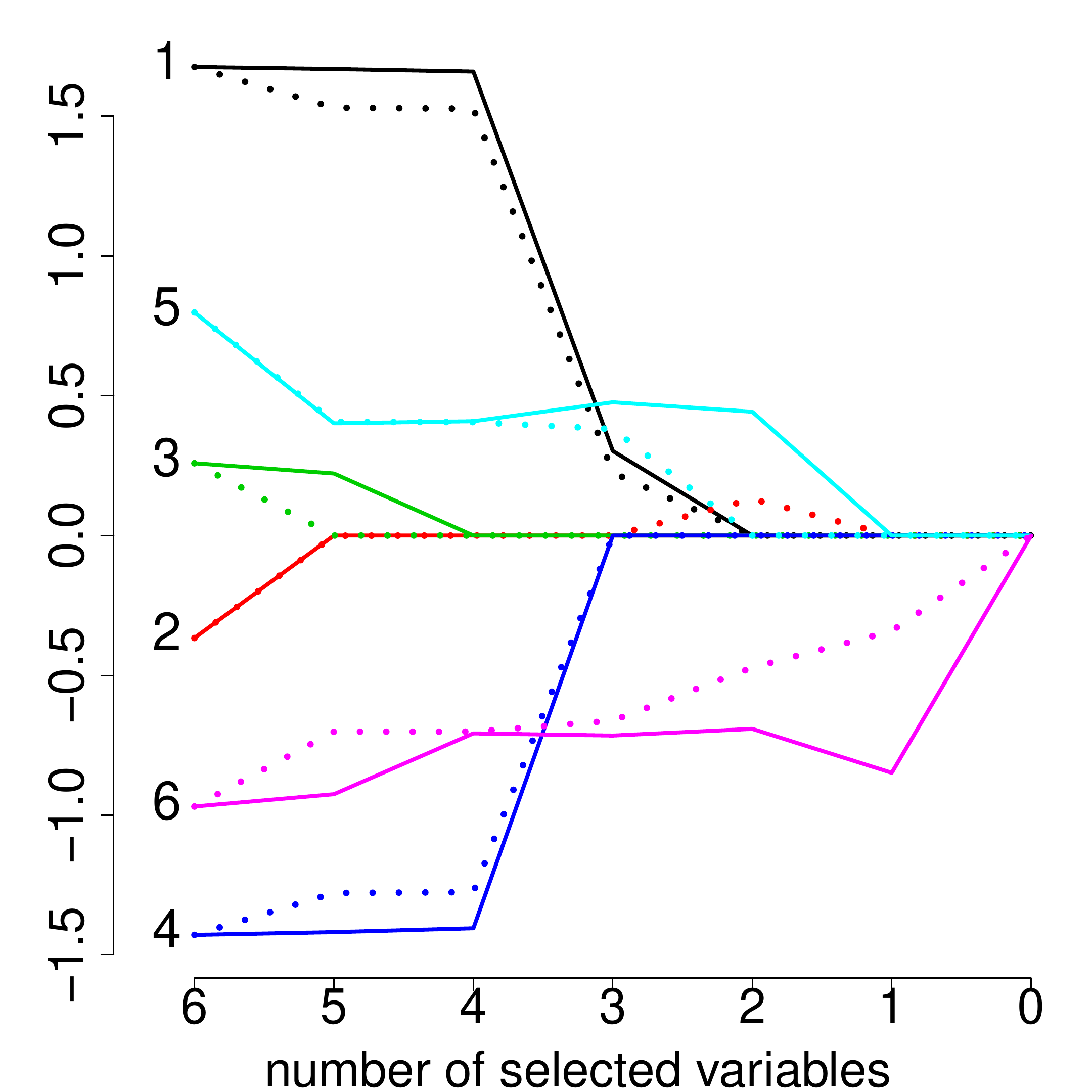}
\includegraphics[trim={0.5cm 1cm 0cm 0cm},clip,width=0.3\linewidth,height=0.28\linewidth]{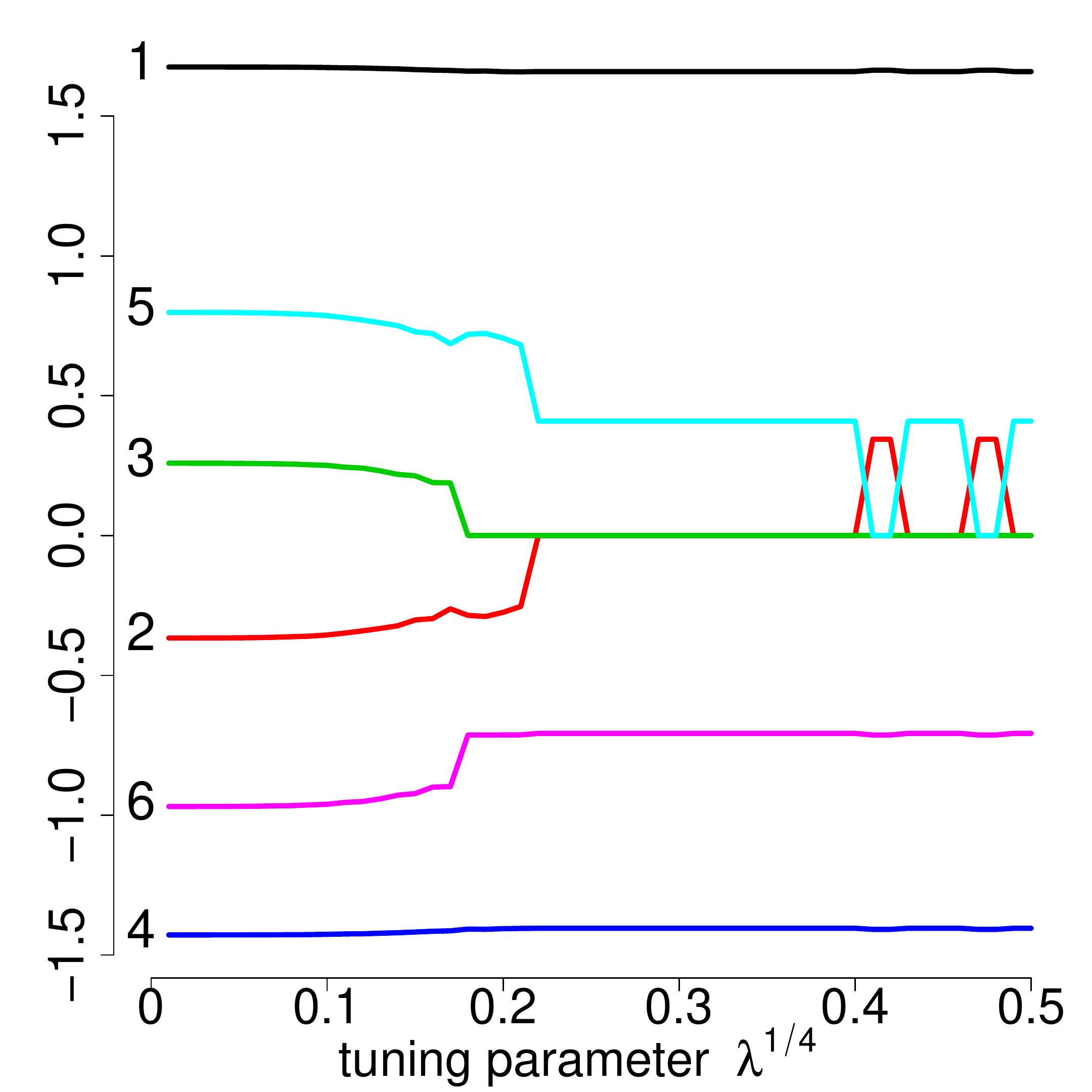}
\includegraphics[trim={0.5cm 1cm 0cm 0cm},clip,width=0.3\linewidth,height=0.28\linewidth]{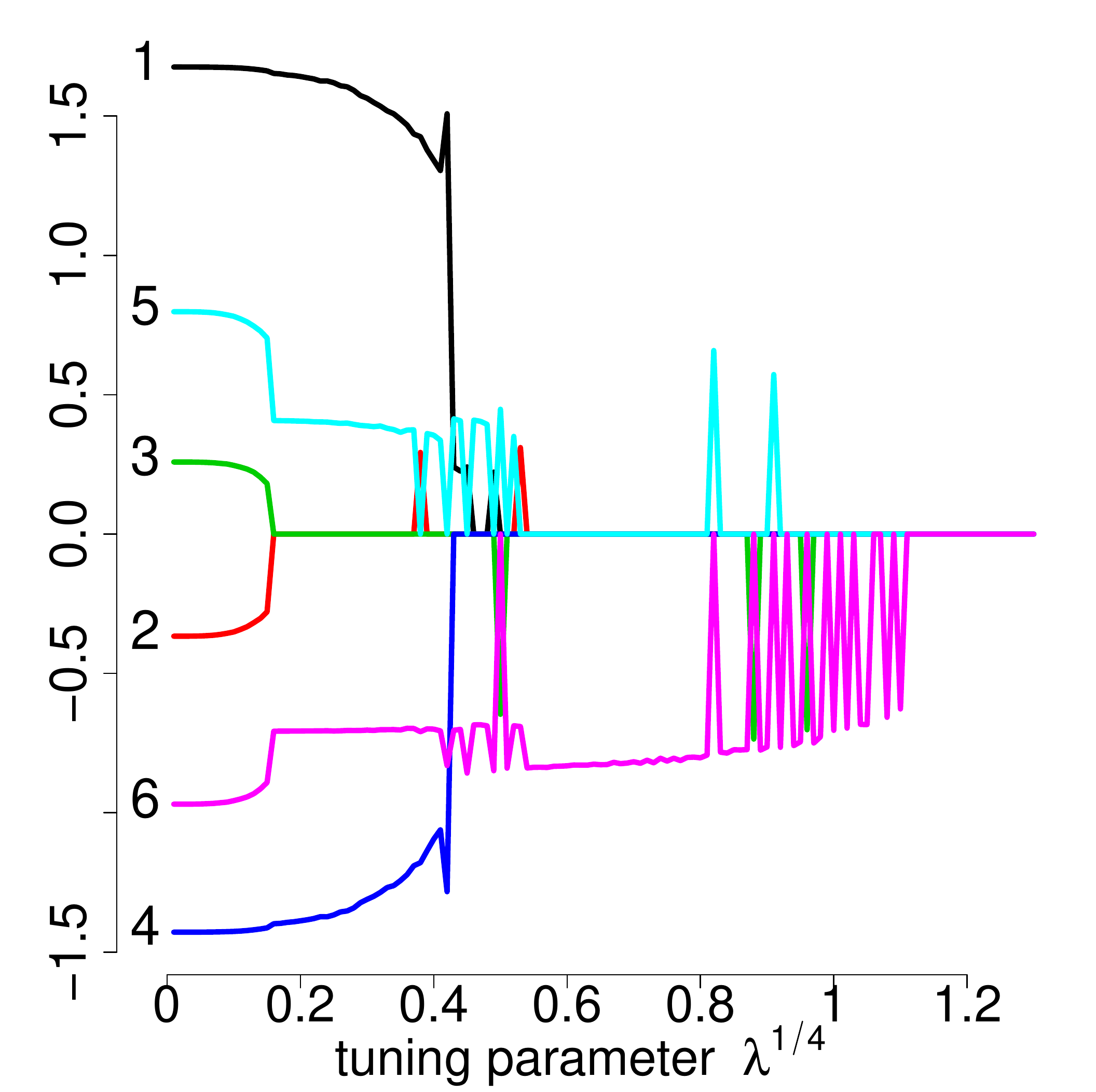}
\end{minipage}
\caption{Estimation trajectory in linear and logistic regression as $n_e$ changes (column A), $\lambda$ changes at fixed $n_e<p$ (column B), and $\lambda$ changes at $n_e=p$ (column C). The solid lines in column A are from the PANDA-$l_0$ regularization by varying $n_e$ from 1 to $p$ and the dash lines represent the lasso regression via R package \texttt{glmnet} with the smallest $\lambda$ that yields $p-n_e$ non-zero estimates.} \label{fig:l0trace}
\vspace{-9pt}
\end{figure}

The results are presented in Fig \ref{fig:l0trace}. Column A shows that the PANDA-$l_0$ regularization  shrinks only $n_e$ coefficients towards $0$, leaving the other coefficients unregularized, but lasso shrinks all coefficients simultaneously. The observations are consistent with Proposition \ref{prop:glmorthogonalregularization} and Eqn (\ref{eqn:l0orthogonalregularization}), which state the number of selected variables through PANDA-$l_0$ is $p-n_e$ for $n_e<p$. In the logistic regression case, due to the high correlations ($0.957$, $0.969$ and $0.974$) between the linear and the quadratic terms, the shrinkage occurs roughly around the same $\lambda$ for a fixed $n_e$ for each linear$+$quadratic pair in the trajectory. The  plots in column B examine the effect of $\lambda$ on the estimation trajectory fixing $n_e$ at $p-p_0$ in PANDA-$l_0$, where $p_0$ is number of variables selected by lasso ($p_0=3$ in linear regression and $p_0=4$ in logistic regression). As $\lambda$ increases, $n_e$ coefficients shrinks to $0$.  Further increasing  $\lambda$ has no regularization effect on the remaining non-zero coefficients, despite some minor fluctuation around the non-zero parameter estimates.  The plots in column C are similar to column B but $n_e$ is fixed at $p$. For small $\lambda$, the estimation trajectories are similar to using $n_e<p$ as in column B; as $\lambda$ continues to increase, the non-zero coefficients eventually get shrunk to 0, but in a different manner than lasso in the sense that its shrinkage process is not gradual but rather abruptly. For $n_e>p$, the estimation trajectories would be somewhere between column C and the lasso trajectories (e.g. Fig 4), and eventually become the lasso trajectories as $n_e$ becomes very large. 

\subsection{Inference for GLM parameters via PANDA}\label{sec:inference}\vspace{-3pt}
We investigate the inferential validity for GLM coefficients based on the asymptotic distributions in Proposition \ref{prop:asymp.dist.UGM} via simulation studies.  We examine Gaussian  ($\sigma^2=1$), Poisson,  Bernoulli, exponential (exp), and Negative Binomial (NB) (number of failure fixed at $r=5$) outcomes  with $p=30$ in each case. For the Gaussian and NB outcomes, the predictors were simulated from N$(0,1)$; for the Bernoulli, exp, and Poisson outcomes, the predictors were simulated from Unif$(-3,3)$, Unif$(-1,2)$, and Unif$(-0.3,0.5)$, respectively. We examine three sample size scenarios $n=50,70,100$, with 500 repetitions in each simulation case. The bridge-type noise is employed  with $\gamma=1, n_e=n, \lambda n_e\in (1.5,7)$ in logistic regression and $\gamma=2, n_e=9, \lambda=n/10$ in the other GLMs. The achieved regularization effect is lasso in the logistic regression and $l_0$ in the other GLMs as $n_e=9$ is set at the number of zero coefficients.  
In each repetition, we calculate the 95\% CIs for the 30 regression coefficients  ($21$ are non-zero and $9$ are zero) and examine the coverage probabilities (CP) and the CI widths.  Tables \ref{tab:ci} presents the results, benchmarked against the post-lasso inferential procedure \citep{CIJason2016,CIJonathan2017} implemented via the R package \texttt{selectiveInference}.   

\begin{table}[!htb]
\begin{center}
\caption{Empirical CP and CI width via PANDA and post-selection procedures with lasso penalty} \label{tab:ci}
\resizebox{0.95\columnwidth}{!}{
\begin{tabular}{l ccc l ccc l ccc l ccc}
\hline
& \multicolumn{7}{c}{\cellcolor{gray!15} \textbf{zero coefficients (9)} } &&  \multicolumn{7}{c}{\cellcolor{gray!15} \textbf{non-zero coefficients (21)} }\\
\hline 
& \multicolumn{3}{c}{PANDA} && 
\multicolumn{3}{c}{post selection } && 
\multicolumn{3}{c}{PANDA} && 
\multicolumn{3}{c}{post selection }\\
\cline{2-4}\cline{6-8}\cline{10-12}\cline{14-16}
sample size & 50&70&100 && 50&70&100 && 50&70&100 && 50&70&100\\
\hline
&\multicolumn{7}{c}{mean CP (\%) among the 9 coefficients} && \multicolumn{7}{c}{mean CP (\%) among the 21 coefficients}\\
\hline
Gaussian & 98.2&99.5&99.9 && NA&NA&NA && 91.4&96.5&97.7 && 92.3&93.2&94.2 \\
Bernoulli & 100&99.5&96.6 && NA&NA&NA && 97.3&88.4&92.6 && 65.9&75.2&82.9 \\ 
Exp & 95.1&95.5&96.5 && -&-&- && 87.1&99.5&94.4 && -&-&- \\
Poisson & 92.2&95.8&98.5 && -&-&- && 87.0&87.5&94.1 && -&-&- \\
NB & 95.8&99.4&100 && -&-&- && 83.1&95.4&99.9 && -&-&- \\
\hline
&\multicolumn{7}{c}{mean CI width among the 9 coefficients} && \multicolumn{7}{c}{mean CI width among the 21 coefficients}\\
\hline
Gaussian & 0.28&0.15&0.08 && NA&NA&NA && 0.91&0.74&0.57 && 29.6&2.01&1.26 \\
Bernoulli & 14.6&1.32&0.93 && NA&NA&NA && 24.8&2.15&1.46 && 22.0&10.6&4.64 \\ 
Exp & 0.39&0.23&0.14 && -&-&- && 1.07&0.95&0.77 && -&-&- \\
Poisson & 0.76&0.44&0.25 && -&-&- && 1.28&1.08&0.91 && -&-&- \\
NB & 0.54&0.28&0.15 && -&-&- && 1.19&1.11&1.05 && -&-&- \\
\hline
\end{tabular}}
\resizebox{0.95\columnwidth}{!}{\begin{tabular}{l}
\footnotesize NA: Not Available. R Package \texttt{selectiveInference} does not provide inference for coefficients whose estimates are 0. \\
\footnotesize  In addition, it only produces CIs for  linear and logistic regression with the lasso regularization. CIs obtained by \\
\footnotesize \texttt{selectiveInference} that  have infinite lower/upper bounds are excluded from the summary (4 $\sim$ 18\%). \\ 
\hline
\end{tabular}}
\end{center}\vspace{-21pt}
\end{table}

For true zero-valued coefficients,  PANDA  maintains the nominal 95\% coverage for all the examined outcome types and sample sizes.  The R \texttt{selectiveInference} package does not provide inference for coefficients whose estimates are 0 (that is, not selected by lasso in the first place). Among these 9 zero-valued coefficients, lasso only selected some of them a few times out of the 500 repetitions.  When the true coefficients are not 0, the CIs from PANDA have better coverage with much narrower CIs than the post-selection procedure (except for logistic regression at $n\!=\!50$). 
The post-selection procedure experiences severe under-coverage in the logistic regression for all $n$. 
We also examined the case of a larger $n_e$ ($n_e\!=\!2n$ in the logistic regression and $n_e\!=\!n$ for the other GLMs). There was some under-coverage (CP $\ge\sim\!90\%$ for zero coefficients; $\ge\sim\!80\%$ for non-zero coefficients), which improved as $n$ increased.

\subsection{Comparison with Existing Regularization in Linear and Logistic Regression}\label{sec:linear}\vspace{-3pt}
To examine the regularization effects by PANDA in linear and logistic regression, we use the same simulation setting as Examples 4.1 and 4.3 in \citet{SCAD}. In the linear regression, $Y\!=\!\x\bs\beta+\epsilon$, where $\bs{\beta}^T=(3,1.5,0,0, 2, 0, 0, 0)$ ($p=8$),  $x_j\sim N(0,1)$ for $j=1,\ldots,p$ with corr$(x_j,x_{j'})=0.5^{|j-j'|}$, and $\epsilon\sim N(0,\sigma^2)$. Three sets of $(n,\sigma)$ were examined: (40, 3), (40, 1), and (60, 1).  For the logistic regression, $n$ was set at 200; $Y\sim  \mbox{Ber}\big(e^{\X^T\bs\beta}/(1+e^{\X^T\bs\beta})\big)$, where the first six components of $\X$ were the same as those in linear regression and the last two components $x$ were drawn from Bernoulli$(0.5)$ independently. 

The medians of relative model error (MRME) and the number of correctly and incorrectly identified zero coefficients (out of 5)  over 100 repetitions were obtained in each regression case. The estimates from the ridge, lasso, adaptive lasso and EN regressions via the existing approaches were obtained from R package \texttt{glmnet} and those from SCAD were from R package \texttt{ncvreg}. We examine two scenarios of PANDA: large $n_e$/small $m$ and large $n_e$/small $m$. The specific values of $n_e$ and $m$, along with other PANDA algorithmic parameters  are summarized in Table \ref{tab:simset} in the supplementary materials. The results are presented in Table \ref{tab:linear} for the linear regression and in Table \ref{tab:logistic} for the logistic regression. In summary, PANDA is either consistent with or performs better  (due to its additional ensemble behaviors) than existing approaches for the same type of regularizer, per the MRME and the true 0/false 0 counts. The superiority of PANDA is specially obvious in the logistic regression. In general, PANDA-SCAD and PANDA-$l_0$ have the best performance. 

\begin{table}[!htb]
\begin{center}
\caption{PANDA vs. Existing Approaches in Penalized Linear Regression with Various Regularizers}\label{tab:linear}
\resizebox{1.0\textwidth}{!}{
\begin{tabular}{l@{\hskip6pt|} c@{\hskip6pt}c@{\hskip6pt}c@{\hskip6pt}c@{\hskip6pt}c@{\hskip6pt}c@{\hskip6pt}|c@{\hskip6pt}c@{\hskip6pt}c@{\hskip6pt}c@{\hskip6pt}c@{\hskip6pt}c@{\hskip6pt}} 
\hline
& ridge & lasso & adaptive & EN & SCAD & $l_0$& ridge & lasso & adaptive & EN & SCAD & $l_0$\\
 &  &  & lasso &  &  &&  &  & lasso &  &  & \\
\cline{2-13}
& \multicolumn{6}{c|}{MRME} &  \multicolumn{6}{c}{\# of correctly/incorrectly identified zero coefficients} \\
\cline{2-13}
\rowcolor[gray]{0.9} &\multicolumn{12}{c}{ $n=40,\sigma=3$}\\
\cline{2-13}
Existing            & 80.09 &  67.86  & 68.47 & 68.24 & 72.79 & & 0/0    & 2.78/0.04 & 2.80/0    & 2.66/0.03 & 3.59/0.09 &\\
PANDA  & 80.06 & 67.70 & 67.18 & 68.31 & 72.50 & 78.99 & 
0.01/0  & 2.37/0.01 & 2.69/0.01 & 2.50/0.01 & 4.01/0.17 & 3.83/0.13\\
\cline{2-13}
\rowcolor[gray]{0.9} &\multicolumn{12}{c}{$n=40,\sigma=1$}\\
\cline{2-13}
Existing            & 94.60  &  68.03  & 68.32 & 69.45 & 44.42 && 0/0    & 2.87/0 & 2.83/0  & 2.56/0.03 & 4.72/0 &\\
PANDA  & 95.24 &  67.38  & 63.58 & 68.40 & 44.87 & 45.11& 
0.13/0 & 2.69/0 & 3.07/0 & 2.62/0  & 4.91/0 & 4.86/0\\
\cline{2-13}
\rowcolor[gray]{0.9} &\multicolumn{12}{c}{$n=60,\sigma=1$ }\\
\cline{2-13}
Existing            & 97.40  &  66.40  & 68.34 & 67.92 & 44.91 & & 0/0    & 2.61/0 & 2.66/0  & 2.55/0.03 & 4.96/0 &\\
PANDA  & 97.62  & 66.22 & 61.48 & 67.02 & 44.82 & 44.77& 
0.19/0 & 2.55/0 & 3.06/0  & 2.43/0 & 5.00/0 & 5.00/0\\
\hline
\end{tabular}}
\end{center}\vspace{-6pt}
\end{table}

\begin{table}[!htb]
\begin{center}
\caption{PANDA vs. Existing Approaches in Penalized Logistic Regression with Various Regularizers}\label{tab:logistic}
\resizebox{1.0\textwidth}{!}{
\begin{tabular}{l@{\hskip6pt}| c@{\hskip6pt}c@{\hskip6pt}c@{\hskip6pt}c@{\hskip6pt}c@{\hskip6pt}c@{\hskip6pt}|c@{\hskip6pt}c@{\hskip6pt}c@{\hskip6pt}c@{\hskip6pt}c@{\hskip6pt}c@{\hskip6pt}} 
\hline
& ridge & lasso & adaptive & EN & SCAD & $l_0$& ridge & lasso & adaptive & EN & SCAD & $l_0$\\
 &  &  & lasso &  &  &&  &  & lasso &  &  & \\
\cline{2-13}
& \multicolumn{6}{c|}{MRME} &  \multicolumn{6}{c}{\# of correctly/incorrectly identified zero coefficients} \\
\cline{2-13}
Existing            & 85.16  &  68.67  & 67.96 & 69.71 & 48.14 && 
0.07/0 & 2.10/0 & 2.05/0  & 2.09/0 & 4.31/0 &\\
PANDA  & 76.50  &  61.15  & 58.60 & 62.14 & 34.87 & 37.66& 
0.17/0  & 2.43/0 & 2.83/0  & 2.44/0 & 4.97/0.02 & 4.89/0\\
\hline
\end{tabular}}
\end{center}\vspace{-15pt}
\end{table}

\subsection{Sports Article Objectivity Data}\label{sec:sports}\vspace{-3pt}
We implemented PANDA in a real-life dataset that contains  1000 sports articles that are labeled 
``objective'' or ``subjective''. The data set is available for download from the UCI Machine Learning Repository \citep{sports}. There are 59 variables in  the original data.  The independent variables $X$'s are the extracted features from the articles such as the  frequencies of different types of words, (e.g., the objective and subjective SENTIWORDNET scores,  foreign words, subordinating preposition or conjunction) and frequencies of different types of punctuation (e.g., questions marks, exclamation marks), and text complexity score, among others. After removing the redundant features (perfectly linear dependent variables) and the highly imbalanced features (e.g, >99\% in one category), and adjusting for the total word counts, we kept 48 $X$'s plus $Y$ (365 ``subjective'' and 635 ``objective'').   We split the 1000 cases into 800 training samples and 200 testing samples (100 subjective vs. 100 objective). 

We learned the logistic regression parameters based on the 800 training samples and make predictions for the 200 testing samples via the trained model. We run the logistic regression with lasso, ridge, EN, and adaptive lasso penalties via the R package \texttt{glmnet}, and with the SCAD penalty via the \texttt{ncvreg} package, and obtained the regularized regression with the same types of penalty listed the above using PANDA. For the existing approaches, the 10-fold CV was used for hyper-parameter tuning. For PANDA, we run 100 iterations with $n_e=1000$ and $m=10$.  The algorithm converged after $10\sim 15$ iterations, and the final parameter estimates were averaged over the last $r=20$ iterations with $\tau_0=0.01$. 

Table \ref{tab:sports} presents the results on the MSE, classification accuracy rate, and computational time.  Compared to the MLE from the non-regularized logistic regression, the  prediction MSE and the  accuracy rate on the testing samples via PANDA are similar or slightly better with the regularizers realized with the R packages \texttt{glmnet} and  \texttt{ncvreg}. Specifically, the prediction MSE decreases by $\approx10\%$; the accuracy increases by 1.5\% to 2\% for the same regularizer types.  The number of zero coefficients ranges about 10 to 20 (out of a total of 48), depending on which regularizer is used per \texttt{glmnet} and  \texttt{ncvreg}. PANDA took about 1.5 to 2 seconds to run 50 iterations. However, 25 iterations (costing 0.7 to 1 seconds) would also be sufficient for this application. Suppose $t$ values of tuning parameters are used in a $K$-fold cross-validation. The total time including the hyper-parameter tuning would be around $0.7tK$ to $tK$ seconds. Say $t=10$ and $K=10$, then it will take about 1 to 1.5 mins for PANDA. This is significantly longer than the existing method, which is expected since PANDA involves random sampling of data points and running GLM for every iteration.  
\begin{table}[!htb]
\begin{center}
\caption{PANDA vs Existing Approaches in the Sports Article Objectivity Data}  \label{tab:sports}
\resizebox{0.9\textwidth}{!}{
\begin{tabular}{l cccccc} 
\hline
penalty &  ridge & lasso & EN & adaptive lasso & SCAD &$l_0$\\
\hline
\rowcolor[gray]{0.9} &\multicolumn{6}{c}{Prediction MSE (0.1573 with MLE)}\\
 \hline
Existing & 0.1539 & 0.1561 & 0.1544 & 0.1561  & 0.1629 &\\
PANDA   & 0.1312 & 0.1260 & 0.1277 & 0.1280 & 0.1340 & 0.1448\\
\hline
\rowcolor[gray]{0.9}
&\multicolumn{6}{c}{Accuracy Rate/Sensitivity/Specificity (\%): 78.5/94/63 with MLE} \\
 \hline
Existing & 78.5/95/62 & 77.5/95/60 & 77.5/95/60 & 78/95/61 & 77/94/60 &\\
PANDA  & 83.5/91/76 & 82.5/87/78 & 84.5/90/79 & 82/86/78 & 81.5/85/78 &79.5/92/67\\
\hline 
\rowcolor[gray]{0.9}
&\multicolumn{6}{c}{\# of zero-valued coefficients (0 with MLE) } \\
 \hline
Existing & 0 & 8  & 4& 10 &  25& \\
PANDA$^{\ddagger}$ & 1& 8 & 4  &10  & 25 & 19\\
 \hline
\rowcolor[gray]{0.9}
 &\multicolumn{6}{c}{Computational Time (sec)$^{||}$} \\
 \hline
Existing & \multicolumn{5}{c}{0.7 $\sim$ 0.8} \\
PANDA &\multicolumn{5}{c}{0.3 $\sim$ 0.4 per 10 iterations}  & $\sim$2.5 \\
\hline
\end{tabular}}
\resizebox{0.9\textwidth}{!}{
\begin{tabular}{l}
\footnotesize $^{\ddagger}$ hyperparameters were tuned to match the \# of zero-valued coefficients in existing methods.\hspace{48pt}\textcolor{white}{.}
\\ 
\footnotesize $^{||}$  V1.1.463 on PC (Intel Core i7-7660U CPU @ 2.50 GHz)\\
\hline
\end{tabular}}
\end{center}\vspace{-15pt}
\end{table}

We also run PANDA using the same tuning parameters selected by the R packages for the existing approaches for the same type of regularizer. PANDA performs better than the existing approaches with smaller RMSE, slightly better accuracy rates, and doubled zero coefficients in most cases.

\vspace{-6pt}\section{Discussion} \label{sec:discussion}\vspace{-3pt}
PANDA is a regularization technique through noise augmentation. PANDA effectively regularizes parameter estimation and allows valid inferences for GLMs, and displays ensemble learning behavior in certain cases. We establish the Gaussian tail of the noise-augmented loss function  and the almost sure convergence to its expectation -- a penalized loss function with the targeted regularizer, providing the theoretical justification for PANDA as a regularization technique and that the noise-augmented loss function is trainable. For a pre-fixed $n_e<p$, we show that PANDA is equivalent to imposing $n_e$ linear constraints on parameters and can lead to the $l_0$ regularization. 
PANDA is straightforward to implement. There is no need for sophisticated optimization techniques as PANDA can leverage existing functions or procedures for running GLMs in any statistical software. In terms of the computational time, large $n_e$ usually leads to convergence with a small number of iterations, but the per-iteration computational cost can be high. If PANDA is applied to yield the $l_0$ regularization or to obtain inference in GLMs on top of variable selection, a small $n_e$ with a relatively large $m$ should be used. 

The PANDA algorithm calculates $\bar{\bs{\bs\theta}}$, the  average of $m$ minimizers of $l(\bs\theta|\tilde{\x},\tilde{\y})$ from the latest $m$ iterations, so to leverage the existing software for running GLM and to maintain its computational advantage over the existing approaches that employ sophisticated optimization techniques.  Proposition \ref{prop:glmregularization} suggests the average of $m$ noise-augmented loss function $l(\bs\theta|\tilde{\x},\tilde{\y})$ yields a single minimizer $\hat{\bs{\bs\theta}}$, the Monte Carlo version of $\E_{\e}(l_p(\bs{\bs\theta}|\tilde{\x},\tilde{\y})$ as $m\rightarrow\infty$. 
We establish in Corollary \ref{cor:average} in the supplementary materials that $\bar{\bs{\bs\theta}}$ and $\hat{\bs{\bs\theta}}$  are first-order equivalent for large $m$ and $n_e$ for PANDA in linear regression, 
We also present simulation results in the linear regression and Poisson regression settings to  illustrate the similarity between  $\bar{\bs{\bs\theta}}$ and $\hat{\bs{\bs\theta}}$. 

For linear regression, the OLS estimator obtained from the noise-augmented data in each iteration of the PANDA algorithm is a \emph{weighted ridge estimator} on the observed data. Compared to a regular ridge estimator, where the same constant is added to all the diagonal elements of $\x^T\x$, different constants are used for different diagonal elements in weighted ridge regression. The formal results and the proof are provided in Sec \ref{sec:wrr} of the supplementary materials.


PANDA and the noise augmentation technique, in general, can be extended to regularize other types of learning problems such as undirected graphical models, where some of the existing techniques are GLM-based. The realized $l_0$ penalty by noise augmentation can be used to regularize learning problems where such penalty is desired but hard to realize due to computational constraints. Regarding the ensemble learning behavior of PANDA, it is worthwhile to study further the underlying theory and run more empirical studies to quantify the benefits of the diversity term enabled by PANDA in generalization error reduction.

\small
\setlength{\bibsep}{3pt}
\bibliographystyle{apa}

\clearpage
\normalsize
\setcounter{page}{1}
\setcounter{section}{0}
\setcounter{algorithm}{0}
\setcounter{figure}{0}
\setcounter{table}{0}
\setcounter{thm}{0}

\renewcommand\thesection{S.\arabic{section}}
\renewcommand\thealgorithm{S.\arabic{algorithm}}
\renewcommand\thefigure{S.\arabic{figure}}
\renewcommand\thetable{S.\arabic{table}}
\renewcommand\thecor{S.\arabic{cor}}

\begin{center}
\Large  \textbf{Supplementary Materials to }\\
\textbf{\emph{\large Adaptive Noisy Data Augmentation for Regularized Estimation and Inference of Generalized Linear Models\vspace{6pt}}}\\
\normalsize Yinan Li,  Fang Liu\\
\small Department of Applied and Computational Mathematics \& Statistics\\
\small University of Notre Dame, Notre Dame, IN 46556, U.S.A. 
\end{center}

\maketitle

\setcounter{equation}{0}
\normalsize
\section{\large Proof of Proposition \ref{prop:glmregularization}}\label{app:glmregularization}\vspace{-3pt}
We take the Taylor expansion of $l_p(\bs\theta|\tilde{\x},\tilde{\y})$, which is the negative log-likelihood, around $e_i\!=\!0$ for $i=1,\ldots,n_e$, and evaluate its expectation over the distribution of $e_i$.  
\begin{align*}
&l_p(\bs\theta|\tilde{\x},\tilde{\y}))=
l(\bs\theta|\x)+l_p(\bs\theta|\e)=
\textstyle l(\bs\theta|\x)) +\!\sum_{i=1}^{n_e} l_i(\bs\theta|\e_i)\\
=&\textstyle l(\bs\theta|\x)- \!\sum_{i=1}^{n_e} \left(h(e_i)+e_i\left(\theta_0\!+\!\sum_j\theta_je_{ij}\right)-B\left(\theta_0\!+\!\sum_j\theta_je_{ij}\right)\right)\\
=& l(\bs\theta|\x)\!+\!\textstyle\sum_{i=1}^{n_e} \!l_i(\bs\theta|\e_i)|_{\e_i=0}\!-\!
\sum_{i=1}^{n_e}
\left\{e_i\!\sum_j\left(\theta_je_{ij}\right)\!-\!
\sum_j(\theta_je_{ij})\!
B'\left(\theta_0\!+\!\sum_j\theta_je_{ij}|_{e_{ij=0}}\right)\right.\\
&\left.\textstyle\qquad\qquad\qquad\qquad-
\sum_{d=2}^{\infty}(d!)^{-1}\!\sum_j(\theta_je_{ij})^d\!B^{(d)}\! \left(\theta_0\!+\!\sum_j\theta_je_{ij}|_{e_{ij=0}}\right)\!\right\}\\
=&l(\bs\theta|\x)\!+\!C\!+\!
\textstyle\sum_{i=1}^{n_e}\!\left[(B'(\theta_0)\!-\!e_i)\sum_j(\theta_je_{ij})\!+\!\sum_{d=2}^{\infty}(d!)^{-1}B^{(d)}(\theta_0)\!\sum_j(\theta_je_{ij})^d\right],
\end{align*}
where $C=B(\theta_0)-\sum_{i=1}^{n_e} \left(h(e_i)+e_i\theta_0\right)$, a constant independent of $\bs\theta$. The expectation of $l_p(\bs\theta|\tilde{\x},\tilde{\y})$  over the distribution of $e_{ij}\sim N(0,\mbox{V}(e_j))$ is
\begin{align}
\E_{\e}(l_p(\bs\theta|\tilde{\x},\tilde{\y}))\!&=\!
l(\bs\theta|\x)\!+\!C\!+\!\textstyle
n_e\!\left(\!\frac{1}{2}B''(\theta_0)\!\sum_j\!\theta^2_j\mbox{V}(e_j)\right)\!+\! O\!\left(n_e\!\sum_j\!\left(\theta_j^4\E(e_j^4)\right)\right)\notag\\
\!&=\!l(\bs\theta|\x)\!+\!C\!+\!\textstyle
n_e\!\left(\!\frac{1}{2}B''(\theta_0)\!\sum_j\!\theta^2_j\mbox{V}(e_j)\right)\!+\!O\!\left(\!n_e\!\sum_j\!\left(\theta_j^4\V^2(e_j)\!\right)\!\right)\notag\\
= l(\bs\theta|\x)+
n_e&\textstyle\!\left(C_1\!\sum_j\theta^2_j\mbox{V}(e_j)\right)\!+\!C \!+\!O\!\left(\!n_e\!\sum_j\!\left(\theta_j^4V^2(e_j)\!\right)\!\right),\mbox{ where $C_{1}=2^{-1}B''(\theta_0)$}. \label{eqn:appB}
\end{align}

There are two ways to realize the expectation in Eqn (\ref{eqn:appB}) empirically.  One way is to approximate $l_p(\bs\theta|\e)$ by $\lim_{m\rightarrow\infty} m^{-1}\sum_{t=1}^m \sum_{i=1}^{n_e} l_i(\bs\theta|\e^{(t)}_i)$. The other way, under the constraint $n_e\mbox{V}(e_j)\!=\!O(1)$, is to let $n_e\rightarrow\infty$, in which case the second term in Eqn (\ref{eqn:appB}) becomes
$n_eC_{1}\sum_j\theta_j^2\mbox{V}(e_{ij})\!=\!
n_eC_{1}\sum_j\!\left(\!\theta^2_j\lim_{n_e\rightarrow\infty} n_e^{-1}\!
\sum_{i=1}^{n_e}\e^2_{ij}n_e\mbox{V}(e_j)\!\right)$. Between the two approaches, letting $n_e\rightarrow\infty\cap\left[n_e\mbox{V}(e_j)\!=\!O(1)\right]$  also leads to the big-$O$ term $O\left(\sum_j\!\left(\theta_j^4n_e\V^2(e_j))\!\right)\!\right)\!\rightarrow\! 0$ in Eqn (\ref{eqn:appB}); in other words, the second order Taylor approximation of $\E_{\e}(l_p(\bs\theta|\tilde{\x},\tilde{\y}))$ is arbitrarily close to $\E_{\e}(l_p(\bs\theta|\tilde{\x},\tilde{\y}))$. 

If $\theta_0=0$, then $C_{1j}\!=\!B''(0)$ and Eqn (\ref{eqn:appB}) can be simplified to
\begin{equation}\label{eqn:appB2}
l(\bs\theta|\x) + \textstyle C_2\sum_j \theta^2_j\left(n_e\mbox{V}(e_j)\right)\!+\!C \!+\!O\left(\sum_j\!\left(\theta_j^4n_e\V^2(e_j))\!\right)\!\right).
\end{equation}
For linear regression, the expectation of $l_p(\bs\theta|\tilde{\x},\tilde{\y})\!=\!\textstyle\sum_{i=1}^{n+n_e}\!\!\left(\tilde{y}_i\!-\!\sum_j\tilde{x}_{ij}\theta_j\!\right)^2\!$ over the distribution of noise $\e$ is
\begin{align}
\E_\e (l_p(\bs\theta|\tilde{\x},\tilde{\y}))\!&=\!\textstyle
\sum_{i=1}^{n}\!\left(x_{ij}\!-\!\sum_jx_{ij}\theta_j\!\right)^2\!\!+\!
\E_\e\!\left(\!\sum_{i=1}^{n_e}\!\left(\!e_i\!-\!\sum_je_{ij}\theta_j\right)^2\right)\label{eqn:Em}\\
=\textstyle\sum_{i=1}^{n}\left(x_{ij}-\sum_jx_{ij}\theta_j\right)^2&=\textstyle
\sum_{i=1}^{n_e}\E_\e \left(\sum_je_{ij}\theta_j\right)^2=\textstyle l(\bs\theta|{\x})+n_e\sum_j\theta_j^2\mbox{V}(e_{ij})\label{eqn:Ene}.
\end{align}

\section{\large Proof of Proposition \ref{prop:glmorthogonalregularization}} \label{app:glmorthogonalregularization}\vspace{-6pt}
Define Optimization Problem 1 
$\hat{\bs\theta}=\textstyle\arg\min_{\bs\theta}f(\bs\theta)=\arg\min_{\bs\theta}  l(\bs\theta|\x)+C_1\sum_{i=1}^{n_e} \left(\bs\e_i^T\bs\theta\right)^2.
$
Due to its convexity in $\bs\theta$, $\hat{\bs\theta}$ can be solved directly from $\triangledown_{\bs\theta}f(\hat{\bs\theta})=0$.  

Define a constrained optimization Problem 2
\begin{align}
\hat{\bs\theta}=\textstyle\arg\min_{\bs\theta} l(\bs\theta|\x),
\notag\\
\mbox{ s.t. } \textstyle  \sum_{i=1}^{n_e}(\e_i^T\bs\theta)^2\leq d,\label{eqn:constraint}
\end{align}
the corresponding Lagrangian for which is 
$L(\bs\theta)= l(\bs\theta|\x)+\lambda_L \left(\sum_{i=1}^{n_e} \left(\e_i^T\bs\theta \right)^2)-d\right)$,
and the KKT conditions are
\begin{align}
&\textstyle \triangledown L(\bs\theta^*)=0\notag\\
&\lambda_L \ge 0\notag\\
&\textstyle\lambda_L\left(\sum_{i=1}^{n_e} \left(\e_i^T\bs\theta^* \right)^2-d\right)=0.
\end{align} 
For Problem 2 to have the same solution as Problem 1, that is, $\bs\theta^*=\hat{\bs\theta}$, we set $\lambda_L=C_1 $ and $d=\sum_{i=1}^{n_e} \left(\e_i^T
\hat{\bs\theta} \right)^2$. The constraint in Eqn (\ref{eqn:constraint}) now becomes
\begin{align}\label{eqn:overallconstraint}
\textstyle  \sum_{i=1}^{n_e}(\e_i^T\bs\theta)^2\leq \sum_{i=1}^{n_e} \left(\e_i^T\hat{\bs\theta} \right)^2
\end{align}
Given that  $n_e$ noise data points are independent per the PANDA procedure, Eqn (\ref{eqn:overallconstraint}) can also be regarded as $n_e$ linear constraints on $\bs\theta$
\begin{align}
\textstyle\exists \ 0<d_i<\left(\sum_{i=1}^{n_e}(\e_i^T\hat{\bs\theta})^2 \right)^{1/2}: |\e_i^T\bs\theta|\leq d_i, i=1,\ldots,n_e\label{eqn:glmorthogonalregularization}
\end{align} 
\section{\large Proof of Theorem \ref{thm:as}}\label{app:pelf2nmelf}\vspace{-3pt}
We prove Theorem \ref{thm:as} for linear regression ($Y_i$ is Gaussian), Poisson regression, exponential regression, negative binomial regression, and logistic regression (when $Y_i$ is Bernoulli), respectively. WLOG, we use the bridge-type noise $e_{ij}\sim N(0,\lambda|\bs\theta|^{-\gamma})$ to demonstrate the proofs, which can be easily extended to other types of noises. Prior to the proof of Theorem \ref{thm:as}, we state a theoretical result in Claim \ref{cla:consistineq}, on which the subsequent proofs rely on.
\begin{cla}\label{cla:consistineq} If $l_p(\bs\theta|\tilde{\x},\tilde{\y})$ and $l_p(\bs\theta|\x)$ are convex functions w.r.t. $\bs\theta$ and share the same parameter space $\bs\theta$, then
$$\left|\inf\limits_{\bs\theta}l_p(\bs\theta|\tilde{\x},\tilde{\y})-\inf\limits_{\bs\theta}l_p(\bs\theta|\x)\right|\leq \sup\limits_{\bs\theta}\left|l_p(\bs\theta|\tilde{\x},\tilde{\y})-l_p(\bs\theta|\x)\right|$$
\end{cla}
\textbf{Proof}:  Since both $\inf\limits_{\bs\theta}l_p(\bs\theta|\tilde{\x},\tilde{\y})$ and $\inf\limits_{\bs\theta}l_p(\bs\theta|\x)$ are convex optimization problems, each has a global optimum, denoted by $\hat{\bs\theta}$ and $\tilde{\bs\theta}$, respectively. Therefore, $\left|\inf\limits_{\bs\theta}l_p(\bs\theta|\tilde{\x},\tilde{\y})\!-\!\inf\limits_{\bs\theta}l_p(\bs\theta|\x)\right|$ $= \left|l_p(\hat{\bs\theta}|\tilde{\x},\tilde{\y})-l_p(\tilde{\bs\theta}|\x)\right|$. Consider the following two scenarios,
	
i). if $l_p(\hat{\bs\theta}|\tilde{\x},\tilde{\y}) \geq l_p(\tilde{\bs\theta}|\x)$, then $l_p(\tilde{\bs\theta}|\tilde{\x},\tilde{\y}) \geq l_p(\hat{\bs\theta}|\tilde{\x},\tilde{\y}) \geq l_p(\tilde{\bs\theta}|\x)$ and
$$\left|l_p(\hat{\bs\theta}|\tilde{\x},\tilde{\y})-l_p(\tilde{\bs\theta}|\x)\right| = l_p(\hat{\bs\theta}|\tilde{\x},\tilde{\y})-l_p(\tilde{\bs\theta}|\x) \leq l_p(\tilde{\bs\theta}|\tilde{\x},\tilde{\y})-l_p(\tilde{\bs\theta}|\x)=\left| l_p(\tilde{\bs\theta}|\tilde{\x},\tilde{\y})-l_p(\tilde{\bs\theta}|\x)\right|$$
ii). if $l_p(\hat{\bs\theta}|\tilde{\x},\tilde{\y}) < l_p(\tilde{\bs\theta}|\x)$, then $l_p(\hat{\bs\theta}|\tilde{\x},\tilde{\y}) < l_p(\tilde{\bs\theta}|\x) < l_p(\hat{\bs\theta}|\x)$ and
$$\left|l_p(\hat{\bs\theta}|\tilde{\x},\tilde{\y})-l_p(\tilde{\bs\theta}|\x)\right| = l_p(\tilde{\bs\theta}|\x) -l_p(\hat{\bs\theta}|\tilde{\x},\tilde{\y}) \leq l_p(\hat{\bs\theta}|\x) -l_p(\hat{\bs\theta}|\tilde{\x},\tilde{\y}) =\left| l_p(\hat{\bs\theta}|\tilde{\x},\tilde{\y})-l_p(\hat{\bs\theta}|\x)\right|.$$
All taken together, 
$\left|l_p(\hat{\bs\theta}|\tilde{\x},\tilde{\y})\!-\!l_p(\tilde{\bs\theta}|\x)\right|\!\leq\!\max\!\left(\left| l_p(\tilde{\bs\theta}|\tilde{\x},\tilde{\y})\1-\!l_p(\tilde{\bs\theta}|\x)\right|, \left| l_p(\hat{\bs\theta}|\tilde{\x},\tilde{\y})-l_p(\hat{\bs\theta}|\x)\right|\right)$ \\ $\leq \sup\limits_{\bs\theta}\left|l_p(\bs\theta|\tilde{\x},\tilde{\y})-l_p(\bs\theta|\x)\right|$.

\subsection{Linear regression} 
In this case the regularization effects with $n_e\rightarrow\infty$ and $m\rightarrow\infty$ are the same.  
The loss function  upon convergence is 
$$\textstyle \bar{l}_p(\bs\theta|\tilde{\x},\tilde{\y})\!=\!\sum_{i=1}^{n}\!\!\left(\!y_i\!-\!\sum_jx_{ij}\theta_j\!\right)^2\!+m^{-1}\!\sum_{t=1}^{m}\!\sum_{i=1}^{n_e}\!\left(\!e_i\!-\!\sum_je^{(t)}_{ij}\theta_j\!\right)^2.$$
Since $e^{(t)}_{ij}=\sqrt{\lambda|\bs\theta|^{-\gamma}}z^{(t)}_{ij}$, where $z^{(t)}_{ij}\sim N(0,1)$. Therefore,
\begin{align}
\bar{l}_p(\bs\theta|\tilde{\x},\tilde{\y})=&\textstyle l(\bs\theta|\x)\!+m^{-1}\!\sum_{t=1}^{m}\sum_{i=1}^{n_e}\left(\sum_j \frac{\lambda\theta_j^2}{|{\theta_j}|^\gamma}z^{(t)2}_{ij}+2\sum_{j<k}\frac{\lambda\theta_j\theta_k}{|\theta_j\theta_k|^{\frac{\gamma}{2}}}z^{(t)}_{ik}z^{(t)}_{ij}\right)\notag\\
=&\textstyle l(\bs\theta|\x)\!+\!m^{-1}\!\!\sum_{t=1}^{m}\!\sum_j\!\!\left(\!\frac{\lambda\theta_j^2}{|{\theta_j}|^\gamma}\!\sum_{i=1}^{n_e}\!z_{ij}^{(t)2}\!\right)+2m^{-1}\sum_{t=1}^{m}\!\!\sum_{j<k}\!\left(\!\frac{\lambda\theta_j\theta_k}{|\theta_j\theta_k |^{\frac{\gamma}{2}}}\!\sum_{i=1}^{n_e}\! z^{(t)}_{ij}z^{(t)}_{ik}\!\!\right).\notag
\end{align}
Since $\sum_{i=1}^{n_e}\!z_{ij}^{(t)2}\sim \Gamma\left(\frac{n_e}{2},2\right)$ and 
$\Gamma\left(\frac{n_e}{2},2\right)
\approx N(n_e,2n_e)=n_e+ (2n_e)^{1/2}N(0,1)=n_e+(2n_e)^{1/2}z_1$ as  $n_e\rightarrow\infty$; 
$\sum_{i=1}^{n_e}\! z^{(t)}_{il}z^{(t)}_{iv}\sim  \Gamma\left(\frac{n_e}{2},2\right)-\Gamma\left(\frac{n_e}{2},2\right)\approx N(0,4n_e)= 2n_e^{1/2}N(0,1)=2n_e^{1/2}z_2$ as  $n_e\rightarrow\infty$, where $z_1\sim N(0,1)$ and $z_2\sim N(0,1)$.  Therefore, the distribution of  $\bar{l}_p(\bs\theta|\tilde{\x},\tilde{\y})$ can be approximated by
\begin{align}
&\textstyle l(\bs\theta|\x)\!+\sum_jn_e\lambda|\theta_j|^{2-\gamma}\label{app:asymptgauss}\\
&+\sum_j\!\left(\!n_e\lambda |\theta_j|^{2-\gamma} 2^{1/2}n_e^{-1/2}\!\left(\!\frac{1}{m}\!\sum_{t=1}^{m}z_1^{(t)}\!\right)\!\right)\!+\!
\!\sum_{j<k}\!\!\left(\!\frac{\lambda n_e\theta_j\theta_k}{|\theta_j\theta_k|^{\frac{\gamma}{2}}}\! \left(2n_e^{-1/2}\!\right)\!\left(\! \frac{1}{m}\!\sum_{t=1}^{m} z_2^{(t)}\!\right)\!\right)\notag\\
&= l_p(\bs\theta|\x)\!+(mn_e)^{-1/2}C_1N(0,1)\mbox{ where } C_1\!=\!n_e\lambda \bigg(2 \bigg|\bigg|\bigg(\frac{\bs\theta}{|{\bs\theta}|^{\frac{\gamma}{2}}}\bigg)\bigg(\frac{\bs\theta}{|{\bs\theta}|^{\frac{\gamma}{2}}}\bigg)^T\bigg|\bigg|_2^2\bigg)^{1/2},\label{app:asygauss1}
\end{align}
where $l_p(\bs\theta|\x)=l(\bs\theta|\x)\!+\sum_jn_e\lambda|\theta_j|^{2-\gamma}=\E_{\e}(l_p(\bs\theta|\tilde{\x},\tilde{\y}))$. Exactly the same Eqn (\ref{app:asygauss1}) can be obtained by letting $m\rightarrow\infty$ rather than $n_e\rightarrow\infty$. 

Per the strong law of large numbers (LLN), Eqn (\ref{app:asygauss1}) suggests $\bar{l}_p(\bs\theta|\tilde{\x},\tilde{\y})$ converges almost surely to its mean for all $\bs\theta\in\bs\theta$ as  $m\rightarrow\infty$ or $n_e\rightarrow\infty$ (with $n_e\lambda=O(1)$),  assuming $|\theta_j|$ belongs to a compact parameter space and is bounded by $B$.  Consequently, $\sup\limits_{\bs\theta}\left|\bar{l}_p(\bs\theta|\tilde{\x},\tilde{\y})-l_p(\bs\theta|\x)\right|\overset{\mbox{a.s.}}{\longrightarrow}0$
as $m\rightarrow\infty \mbox{ or } n_e\rightarrow\infty$. Per Claim \ref{cla:consistineq}, 
$\inf\limits_{\bs\theta}\bar{l}_p(\bs\theta|\tilde{\x},\tilde{\y})\overset{\mbox{a.s.} }{\longrightarrow}\inf\limits_{\bs\theta}l_p(\bs\theta|\x)$, and 
$\arg\inf\limits_{\bs\theta}\bar{l}_p(\bs\theta|\tilde{\x},\tilde{\y})\overset{\mbox{a.s.}}{\longrightarrow}\arg\inf\limits_{\bs\theta}l_p(\bs\theta|\x)$
due to the convexity of the loss function. 
  	
\subsection{Poisson Regression}
The averaged noise-augmented loss function over $m$ iterations upon convergence is 
\begin{align}
\!\!&\!\!\bar{l}_p(\bs\theta|\tilde{\x},\tilde{\y})\!=\!l(\bs\theta|\x)\!-\!\frac{1}{m}\!\sum_{t=1}^{m}\!\sum_{i=1}^{n_e}\!\!\!\left(\!e_i\!\!\left(\!\theta_0\!+\!\!\sum_j\!e_{ij}^{(t)}\theta_j\!\!\right)\!\!-\! \log(e_i!)\!-\!\exp\!\!\left(\!\theta_0\!+\!\!\sum_j\!e_{ij}^{(t)}\theta_j \!\right)\!\!\right)\!\!\!\label{app:poiexp0}\\
=&l(\bs\theta|\x)\!-\frac{1}{m}\!\sum_{t=1}^{m}e_i\sum_j \sum_{i=1}^{n_e} \theta_je_{ij}^{(t)}+\frac{1}{m}\!\sum_{t=1}^{m}\sum_{i=1}^{n_e}\exp\left(\!\theta_0\!+\!\sum_j\theta_je_{ij}^{(t)} \right)+C\notag\\	
=&l(\bs\theta|\x)\boxed{\!-\frac{1}{m}\!\sum_{t=1}^{m}\!e_i\!\sum_j\left(\!\frac{\sqrt{\lambda}\theta_j}{|\theta_j|^{\frac{\gamma}{2}}}\!\sum_{i=1}^{n_e}z^{(t)}_{ij}\!\right)\!\!+\!\frac{1}{m}\!\sum_{t=1}^{m}\!\!\sum_{i=1}^{n_e}\!\exp\!\left(\!\theta_0\!+\!\sum_j\!\frac{\sqrt{\lambda} \theta_j}{|\theta_j|^{\frac{\gamma}{2}}}z^{(t)}_{ij}\!\!\right)}\!\!+\!C,\label{app:poiexp1}\\
=&l(\bs\theta|\x)+P(\bs\theta)+C,\notag
\end{align}
where $P(\bs\theta)$ refers to the boxed expression in Eqn (\ref{app:poiexp1}), $z^{(t)}_{ij}\sim N(0,1),e_i\equiv n^{-1}\sum_{i=1}^ny_i$ that is a constant, and $C$ is a constant not related to $\bs\theta$. The regularizer $P(\bs\theta)$ is different for $n_e\rightarrow\infty$ vs $m\rightarrow\infty$. We thus consider each case separately.   

\emph{Case 1: $n_e\rightarrow\infty$, $n_e\lambda=O(1)$ and  fixed $m$}\\
Assume $m=1$ WLOG, then $z^{(t)}_{ij}$ can be abbreviated as $z_{ij}$. $n_e\rightarrow\infty$ and $\lambda n_e=O(1)$ implies that $\lambda\rightarrow0$, therefore, $\sum_j\!\frac{\sqrt{\lambda} \theta_j}{|\theta_j|^{\frac{\gamma}{2}}}z_{ij}\rightarrow0$ in Eqn (\ref{app:poiexp1}). Apply the second order Taylor expansion around $\sum_j\theta_jz_{ij}=0$ to Eqn (\ref{app:poiexp1}), as $n_e\rightarrow\infty$, 
\begin{align}
\bar{l}_p(\bs\theta&|\tilde{\x},\tilde{\y})\!\rightarrow\textstyle
l(\bs\theta|\x)\!-\!e_i\sum_j\!\left( \frac{\sqrt{\lambda}\theta_j}{|\theta_j|^{\frac{\gamma}{2}}}\sum_{i=1}^{n_e}z_{ij}\!\right)\!+\!\exp(\theta_0)\sum_{i=1}^{n_e}\sum_j\frac{\sqrt{\lambda} \theta_j}{|\theta_j|^{\frac{\gamma}{2}}}z_{ij}\notag\\
&\ \ \ \ \ \ \ \ \ \textstyle+\!\frac{1}{2}\exp(\theta_0)\sum_{i=1}^{n_e}\left(\sum_j\frac{\sqrt{\lambda} \theta_j}{|\theta_j|^{\frac{\gamma}{2}}}z_{ij} \right)^2+O\left(n_e^{-1}\right)C_1(\bs\theta)N(1,1)+C\label{app:poiexp2}\\
\approx&\textstyle l(\bs\theta|\x)\!+\!\frac{1}{2}\!\exp(\theta_0)\!\sum_j\!\left( \frac{{\lambda}\theta_j^2}{|\theta_j|^{\gamma}} \sum_{i=1}^{n_e}z_{ij}^2\!\right)\!+\!\exp(\theta_0)\!\sum_{j<k}\!\left(\! \frac{\!\lambda\theta_j\theta_k}{|\theta_j\theta_k|^{\frac{\gamma}{2}}}\!\sum_{i=1}^{n_e}\!z_{ij}z_{ik}\!\right)\notag\\
&\ \ \ \ \ \ \ \ \ +O\left(n_e^{-1}\right)C_1(\bs\theta)N(1,1)+C\label{app:poiexp3}\\
\rightarrow &\textstyle l(\bs\theta|\x)+\frac{\lambda n_e}{2}\!\exp(\theta_0)\!\sum_{ k}|{\theta_j}|^{2-\gamma}+\!O\left(n_e^{-0.5}\right)\!C_2(\bs\theta)\!N(0,1)\!+\!O\left(n_e^{-1}\right)\!C_1(\bs\theta)N(1,1)\!+\!C.\label{eqn:2O}
\end{align}
For Poisson regression, $e_i\equiv  n^{-1}\sum_{i=1}^ny_i$, the average of the observations in the outcome node (the log of which estimates $\theta_0)$ with the canonical log link function. In other words, when $n_e\rightarrow\infty$ $e_i=\exp(\theta_0)$; therefore, the second and third terms in Eqn (\ref{app:poiexp2}) cancel out. $C_1(\bs\theta)$ and $C_2(\bs\theta)$ are functions of $\bs\theta$ and the standard deviations associated with the  two asymptotic normality terms in Eqn (\ref{eqn:2O}) which result from the summation over $n_e$ noise terms per the CLT, and the $C_2(\bs\theta)$ term is the rate-limiting term and
\begin{equation}\label{app:asypoi1}
C_2(\bs\theta)=\frac{\lambda n_e}{2} \bigg(2 \!\exp(2\theta_0)\! \bigg|\bigg|\bigg(|{\bs\theta}|^{1-\frac{\gamma}{2}}\bigg)\bigg(|{\bs\theta}|^{1-\frac{\gamma}{2}}\bigg)^T\bigg|\bigg|_2^2\bigg)^{1/2} \mbox{ where } \lambda n_e=O(1).
\end{equation}
Note that $l(\bs\theta|\x)+\frac{\lambda n_e}{2}\!\exp(\theta_0)\!\sum_{ k}|{\theta_j}|^{2-\gamma}$ in Eqn (\ref{eqn:2O}) is $l_p(\bs\theta|\x)=\E_{\e}(l_p(\bs\theta|\tilde{\x},\tilde{\y})$ per Proposition \ref{prop:glmregularization} and Appendix \ref{app:glmregularization}.
As $n_e\rightarrow\infty$ and $\lambda n_e=O(1)$, per the strong LLN and Eqn (\ref{eqn:2O}),  $\bar{l}_p(\bs\theta|\tilde{\x},\tilde{\y})$  converges almost surely to $l_p(\bs\theta|\x)$.
Given the convexity of the loss function and per Claim \ref{cla:consistineq}, 
$\arg\inf\limits_{\bs\theta}\bar{l}_p(\bs\theta|\tilde{\x},\tilde{\y})\overset{{a.s.}}{\longrightarrow}\arg\inf\limits_{\bs\theta}l_p(\bs\theta|\x)$.

\emph{Case 2: $m\rightarrow\infty$ and  fixed $n_e$}\\
The 2nd term in Eqn (\ref{app:poiexp1}) is the summation of Gaussian variables, and the 3rd term follows a log-normal distribution. Therefore, we can rewrite Eqn (\ref{app:poiexp1})  as
\begin{align}
&\bar{l}_p(\bs\theta|\tilde{\x},\tilde{\y})=\textstyle l(\bs\theta|\x)\!-\!e_i\!\sum_j\frac{\sqrt{\lambda }n_e\theta_j}{\sqrt{m}|\theta_j|^{\frac{\gamma}{2}}}N(0,1)+m^{-1}\!\sum_{t=1}^{m}\sum_{i=1}^{n_e}\mbox{LogN}\left(\theta_0, \sum_j \frac{\lambda\theta_j^2}{|\theta_j|^\gamma}\right)+C.\label{eqn:PGMm}
\end{align}
Applying the CLT to Eqn (\ref{eqn:PGMm}) as $m\rightarrow\infty$, 
\begin{align}
&\textstyle\bar{l}_p(\bs\theta|\tilde{\x},\tilde{\y})\rightarrow l(\bs\theta|\x)\!-e_i\sum_j\frac{\sqrt{\lambda} n_e\theta_j}{\sqrt{m}|\theta_j|^{\frac{\gamma}{2}}}N(0,1) \label{app:asymptpoi}\\
&+ \left\{ \frac{n_e}{m}\left(\exp \left(\sum_j \frac{\sqrt{\lambda}\theta_j}{|\theta_j|^{\frac{\gamma}{2}}}\right)^2-1 \right)\exp\!\left(2\theta_0+\left(\sum_j \frac{\sqrt{\lambda}\theta_j}{|\theta_j|^{\frac{\gamma}{2}}}\right)^2  \right)\right\}^{1/2}\!N(0,1)+C,\notag
\end{align}
suggesting that $\bar{l}_p(\bs\theta|\tilde{\x},\tilde{\y})$ follows a Gaussian distribution asymptotically.  Per the strong LLN as $m\rightarrow\infty$, Eqn (\ref{app:asymptpoi}) converges almost surely to 
\begin{align}
\textstyle\E_{\e}(l_p(\bs\theta|\tilde{\x},\tilde{\y}))
&=\!l_p(\bs\theta|\x)\!=\!l(\bs\theta|\x)\!+\!P(\bs\theta)\!+\!C\notag\\
&=\!l(\bs\theta|\x)\!+\!n_e\exp\left(\theta_0\right)\exp\!\left(\!\frac{1}{2}\lambda\!\left(\sum_k {|\theta_j|^{1-\frac{\gamma}{2}}}\!\right)^2 \right)
\!\!+\!C\label{app:poiexp4}
\end{align}
for all $\bs\theta\in\bs\Theta$ assuming $\bs\Theta$ to be compact. 
Per claim \ref{cla:consistineq}, 
$\sup\limits_{\bs\theta}\left|\bar{l}_p(\bs\theta|\tilde{\x},\tilde{\y})-l_p(\bs\theta|\x)\right|\overset{\mbox{a.s.}}{\longrightarrow}0$ as $m\rightarrow\infty$, which leads to
$\inf\limits_{\bs\theta}\bar{l}_p(\bs\theta|\tilde{\x},\tilde{\y})\overset{\mbox{a.s.}}{\longrightarrow}\inf\limits_{\bs\theta}l_p(\bs\theta|\x)
\Rightarrow\arg\inf\limits_{\bs\theta}\bar{l}_p(\bs\theta|\tilde{\x},\tilde{\y})\overset{\mbox{a.s.}}{\longrightarrow}\arg\inf\limits_{\bs\theta}l_p(\bs\theta|\x)$ given the convexity of the loss function.

\subsection{Exponential Regression}
The averaged noise-augmented loss function over $m$ iterations upon convergence is 
$$ \textstyle \bar{l}_p({\bs\theta}|\tilde{\x},\tilde{\y})=l(\bs\theta|\x)-\textstyle m^{-1}\!\sum_{t=1}^{m}\!\sum_{i=1}^{n_e}\!\!\left(\theta_0+\sum_je^{(t)}_{ij}\theta_j\!-e_i\exp\left(\theta_0+\sum_je^{(t)}_{ij}\theta_j \right) \right), $$
where $e_i=n^{-1}\sum_{i=1}^ny_i$. The above loss function is equivalent to the loss function in Eqn (\ref{app:poiexp0}) in the PGM case except for the constant term that does not involve $\bs\theta$. Therefore, the proof for PGM  also applies in the case of EGM.

\subsection{Negative Binomial Regression}
The averaged noise-augmented loss function over $m$ iterations upon convergence is 
\begin{align}
&\bar{l}_p({\bs\theta}|\tilde{\x},\tilde{\y})=\textstyle l(\bs\theta|\x)-\!\frac{1}{m}\!\sum_{t=1}^{m}\!\sum_{i=1}^{n_e}\!\!\bigg(\!\log\!\left(\!\frac{\Gamma(e_i+r)r^{r}}{\Gamma(e_i\!+\!1)\Gamma(r)}\!\right)\!+\!e_i\!\!\sum_j\!e_{ij}^{(t)}\theta_j\notag\\
&\textstyle\qquad\qquad\quad -\!(r\!+\!e_i)\log\!\left(\!r\!+\!\exp\!\left(\!\theta_0\!+\!\sum_j\!e_{ij}^{(t)}\theta_j\!\right)\right)\!\bigg)\label{app:nb0}\\
&=l(\bs\theta|\x)+\!C\!-\!\frac{1}{m}\!\sum_{t=1}^{m}\!\sum_{i=1}^{n_e}\!\!e_i\!\!\sum_j\!e_{ij}^{(t)}\theta_j\!+\!\frac{1}{m}\!\sum_{t=1}^{m}\!\sum_{i=1}^{n_e}\!(r\!\!+\!e_i)\log\!\left(\!\!r\!+\!\exp\!\left(\!\theta_0\!+\!\!\sum_je_{ij}^{(t)}\theta_j\!\!\right)\!\!\right)\label{app:nb01}\\
&=l(\bs\theta|\x)+\!C\;\boxed{\!-\frac{1}{m}\!\sum_{t=1}^{m}\!\!e_i\!\sum_j\!\!\left(\! \frac{\sqrt{\lambda}\theta_j}{|\theta_j|^{\frac{\gamma}{2}}}\!\sum_{i=1}^{n_e}\!z_{ij}^{(t)}\!\!\right) \!+\!\frac{1}{m}\!\sum_{t=1}^{m}\!\!\sum_{i=1}^{n_e}(r+1)\!\log\!\!\left(\!\!r\!\exp\!\left(\!\theta_0\!+\!\sum_j\!\frac{\sqrt{\lambda} \theta_j}{|\theta_j|^{\frac{\gamma}{2}}}z_{ij}^{(t)}\!\right)\!\!\right)\!\!}\!\!\label{app:nb1}\\
&= l(\bs\theta|\x)+P(\bs\theta)+C=l_p(\bs\theta|\x)+C,\notag
\end{align}
where $P(\bs\theta$) refers to the boxed expression in Eqn (\ref{app:nb1}), $z^{(t)}_{ij}\sim N(0,1),e_i\equiv n^{-1}\sum_{i=1}^ny_i$ is a constant, and $C$ is a constant not related to $\bs\theta$. The regularizer $P(\bs\theta)$ is different for $n_e\rightarrow\infty$ vs $m\rightarrow\infty$. We thus consider each case separately.  

\emph{Case 1: $n_e\rightarrow\infty$ and $n_e\lambda=O(1)$ and  fixed $m$}\\
Let $m=1$ WLOG, thus $z_{ij}^{(t)}$ can be abbreviated as $z_{ij}$. Since $n_e\rightarrow\infty$ and $\lambda n_e=O(1)$, implying $\lambda\rightarrow 0$ and thus $\exp\left(\sum_j\frac{\sqrt{\lambda} \theta_j}{|\theta_j|^{\frac{\gamma}{2}}}z_{ij} \right)\rightarrow 1$. Applying the second order Taylor expansion  around $\sum_j\theta_jz_{ij}=0$ to Eqn (\ref{app:poiexp1}), we have
\begin{align}
&\bar{l}_p(\bs\theta|\tilde{\x},\tilde{\y})\!=l(\bs\theta|\x)\!-\!e_i\sum_j\left( \frac{\sqrt{\lambda}\theta_j}{|\theta_j|^{\frac{\gamma}{2}}}\sum_{i=1}^{n_e}z_{ij}\right)+\!\frac{(r+e_i)\exp(\theta_0)}{r+\exp(\theta_0)}\sum_{i=1}^{n_e}\sum_j\frac{\sqrt{\lambda} \theta_j}{|\theta_j|^{\frac{\gamma}{2}}}z_{ij}\notag\\
&\ \ \ \ +\!\frac{1}{2}\!\sum_{i=1}^{n_e}\frac{(r+e_i)r\exp(\theta_0)}{(r+\exp(\theta_0))^2}\left(\sum_j\frac{\sqrt{\lambda} \theta_j}{|\theta_j|^{\frac{\gamma}{2}}}z_{ij} \right)^2+O\left(n_e^{-1}\right)N(1,C_1(\bs\theta)) +C\label{app:nb2}\\
\rightarrow &l(\bs\theta|\x)\!+\!\frac{1}{2}\!\!\sum_j\!\frac{r\exp(\theta_0)}{r\!+\!\exp(\theta_0)}\!\!\left(\!\frac{{\lambda}\theta_j^2}{|\theta_j|^{\gamma}}\!\sum_{i=1}^{n_e}z_{ij}^2\!\!\right)\!\!+\!\!\!\sum_{k<l}\!\frac{r\exp(\theta_0)}{r\!+\!\exp(\theta_0)}\!\!\left(\!\frac{{\lambda}\theta_j\theta_k}{|\theta_j\theta_k|^{\frac{\gamma}{2}}}\!\!\sum_{i=1}^{n_e}\!z_{ij}e_{0il}\!\! \right)\notag \\
&+O\left(n_e^{-1}\right)N(1,C_1(\bs\theta))+\!C\label{app:nb3}\\
\rightarrow &\textstyle l(\bs\theta|\x)\!+\!\frac{1}{2}\!\sum_j\!\frac{r\exp(\theta_0)}{r+\exp(\theta_0)}\!\!\left(\!\frac{{\lambda}\theta_j^2}{|\theta_j|^{\gamma}}\!\sum_{i=1}^{n_e}z_{ij}^2\right)\!+\!O\!\left(\!n_e^{-1}\right)N(1,C_1(\bs\theta))\!+\!O\!\left(n_e^{-0.5}\right)C_2(\bs\theta)N(0,1)\!+\!C\!\!\label{eqn:2ONB}
\end{align}
In NB regression, the logarithm of the average of the observations in the outcome node estimates $\theta_0$ with the canonical log link function. In other words, when $n_e\rightarrow\infty$ $e_i=\exp(\theta_0)$, and $r+\exp(\theta_0)=r+e_i$; therefore, the second and third terms in Eqn (\ref{app:nb2}) cancel out and the forth term can be simplied as shown above. $C_1(\bs\theta)$ and $C_2(\bs\theta)$ are functions of $\bs\theta$ and the standard deviations associated with the  two asymptotic normality terms in Eqn (\ref{eqn:2ONB}) that result from the summation over $n_e$ noise terms per the CLT, and the $C_2(\bs\theta)$ term is the rate-limiting term and 
\begin{equation}\label{app:asynb1}
C_2(\bs\theta)=\frac{\lambda n_e}{2} \bigg(2 \bigg(\frac{r\exp(\theta_0)}{r\!+\!\exp(\theta_0)}\bigg)^2 \bigg|\bigg|\bigg(|{\bs\theta}|^{1-\frac{\gamma}{2}}\bigg)\bigg(|\bs\theta|^{1-\frac{\gamma}{2}}\bigg)^T\bigg|\bigg|_2^2\bigg)^{1/2}.
\end{equation}
Note that $l(\bs\theta|\x)+\frac{1}{2}\!\sum_j\!\frac{r\exp(\theta_0)}{r+\exp(\theta_0)}\!\!\left(\!\frac{{\lambda}\theta_j^2}{|\theta_j|^{\gamma}}\!\sum_{i=1}^{n_e}z_{ij}^2\right)$ in Eqn (\ref{eqn:2ONB}) is $l_p(\bs\theta|\x)\!=\!\E_{\e}(l_p(\bs\theta|\tilde{\x},\tilde{\y})$ per Proposition \ref{prop:glmregularization} and Appendix \ref{app:glmregularization}.
As $n_e\rightarrow\infty$ and $\lambda n_e=O(1)$, per the strong LLN and Eqn (\ref{eqn:2ONB}),  $\bar{l}_p(\bs\theta|\tilde{\x},\tilde{\y})$  converges almost surely to $l_p(\bs\theta|\x)$. Given the convexity of the loss function and  Claim \ref{cla:consistineq}, 
$$\arg\inf\limits_{\bs\theta}{l}_p(\bs\theta|\tilde{\x},\tilde{\y})\overset{{a.s.}}{\longrightarrow}\arg\inf\limits_{\bs\theta}l_p(\bs\theta|\x).$$

\emph{Case 2: $m\rightarrow\infty$ and  fixed $n_e$}\\
The second term in Eqn (\ref{app:nb01}) is the summation over Gaussian variables, therefore, the equation can be written as 
\begin{align}
\bar{l}_p(\bs\theta|\tilde{\x},\tilde{\y})&=\textstyle l(\bs\theta|\x)-\!e_i\sum_j\frac{\sqrt{\lambda} n_e\theta_j}{\sqrt{m}|\theta_j^{(t-1)}|^{\frac{\gamma}{2}}}N(0,1)\!+\!\frac{1}{m}\!\sum_{t=1}^{m}\!\sum_{i=1}^{n_e}U^{(t)}_i+C,\notag\\
&=\textstyle l(\bs\theta|\x)-\!e_i\sum_j\frac{\sqrt{\lambda} n_e\theta_j}{\sqrt{m}|\theta_j^{(t-1)}|^{\frac{\gamma}{2}}}N(0,1)\!+\!\frac{n_e}{m}\!\sum_{t=1}^{m}\!U^{(t)}+C,\label{eqn:NBm}
\end{align}
where $U^{(t)}_i\!=\!\!(r\!+e_i)\log\!\left(\!r\!+\!\exp\left(\!\sum_j\!e_{ij}^{(t)}\theta_j\!\right)\!\right)$. The second equation holds because $U^{(t)}_i$ is the same for all $i=1,\ldots,n_e$. Applying the CLT to the $U$-term in Eqn (\ref{eqn:NBm}) as $m\rightarrow\infty$, 
\begin{align}
&\textstyle\bar{l}_p(\bs\theta|\tilde{\x},\tilde{\y})\rightarrow l(\bs\theta|\x)\!-e_i\sum_j\frac{\sqrt{\lambda} n_e\theta_j}{\sqrt{m}|\theta_j|^{\frac{\gamma}{2}}}N(0,1)+n_e\E\left(U^{(t)}\right)+\frac{n_e}{\sqrt{m}}N\left(0,\sigma_U\right)\notag\\
&\textstyle= l(\bs\theta|\x)+n_e E(U^{(t)})\!-e_i\sum_j\frac{\sqrt{\lambda} n_e\theta_j}{\sqrt{m}|\theta_j|^{\frac{\gamma}{2}}}N(0,1)+\frac{n_e}{\sigma_U\sqrt{m}}N(0,1),\label{app:asymptnb}
\end{align}
where $\sigma_U$ is the standard deviation of $U^{(t)}$. Since $\log(r+\exp(*))\!\rightarrow\!\max\{\log(r),*\}$, as $*\rightarrow\pm\infty$, $\sigma_U$ is a finite.  Eqn (\ref{app:asymptnb}) suggests that $\bar{l}_p(\bs\theta|\tilde{\x},\tilde{\y})$ follows a Gaussian distribution as $m\rightarrow\infty$. 

Additionally, applying the strong LLN to Eqn (\ref{app:nb01}), $\bar{l}_p(\bs\theta|\tilde{\x},\tilde{\y})$  converges almost surely to its mean $l_p(\bs\theta|\x)=\E(l_p(\bs\theta|\tilde{\x},\tilde{\y}))$ for all $\bs\theta\in\bs\theta$ as $m\rightarrow\infty$, assuming $\bs\theta$ to be compact; that is, 
\begin{align}
\bar{l}_p(\bs\theta|\tilde{\x},\tilde{\y})&\rightarrow l_p(\bs\theta|\x)+C= l(\bs\theta|\x)\!+n_e \E(U^{(t)}_i)+C.\label{app:nbbin1}
\end{align}
It follows that $\sup\limits_{\bs\theta}\left|\bar{l}_p(\bs\theta|\tilde{\x},\tilde{\y})-l_p(\bs\theta|\x)\right|\overset{\mbox{a.s.}}{\longrightarrow}0\mbox{ as $m\rightarrow\infty$}
\Rightarrow\inf\limits_{\bs\theta}l_p(\bs\theta|\tilde{\x},\tilde{\y})\overset{\mbox{a.s.}}{\longrightarrow}\inf\limits_{\bs\theta}l_p(\bs\theta|\x)
\Rightarrow\arg\inf\limits_{\bs\theta}l_p(\bs\theta|\tilde{\x},\tilde{\y})\overset{\mbox{a.s.}}{\longrightarrow}\arg\inf\limits_{\bs\theta}l_p(\bs\theta|\x)$ given the convexity of the loss function.

\subsection{Binomial Regression}
The averaged noise-augmented loss function over $m$ iterations upon convergence is 
$$\bar{l}_p({\bs\theta}|\tilde{\x},\tilde{\y})=l(\bs\theta|\x)-\frac{1}{m}\sum_{t=1}^{m}\sum_{i=1}^{n_e}\left(e_i\sum_je^{(t)}_{ij}\theta_j-\log\left(1+\exp\left(\theta_0+\sum_je^{(t)}_{ij}\theta_j\right)\right)\right),$$
which is a special case of Eqn (\ref{app:nb0}) when  $r=1$, and the proof for NBGM also applies to BGM.  


\section{\large Proof of Proposition \ref{prop:consistmcl}}\label{app:consistmcl}
In the case of multicollinearity, PANDA with  sparsity regularization might experience difficulty in learning minimizer $\hat{\bs\theta}_p^{(n_e)}$ (or $\hat{\bs\theta}_p^{(m)}$) when $n_e($ or $m)\rightarrow\infty$. In such a case, we prove that there exists $\epsilon>0$ and a sub-sequence $[n_e]_i$ (or $[m]_i$), such that letting $\bs\theta_p^i\overset{\Delta}{=}\hat{\bs\theta}_p^{[n_e]_i}$ (or $\hat{\bs\theta}_p^{[m]_i}$), then $d\!\left(\!\bs\theta_p^i,\bs\Theta^0 \!\right)\!>\!\epsilon$, where $\bs\Theta^0$ is the optimum parameter set.  Denote $l_p^i\!=\!l_p(\bs\theta_p^i|\tilde{\x},\tilde{\y})$, then by Eqn (\ref{eqn:sup}), there exists a sub-sequence $[i]$, such that,
\begin{align} \label{eqn:supboundmulti1}
\Pr\left( \sup\limits_{\bs\theta}\left|\bar{l}_p(\bs\theta|\tilde{\x},\tilde{\y})-\bar{l}_p(\bs\theta|\x)\right|>\delta \right)<k^{-1}, k\in N.
\end{align}
Since $\bs\theta$ is compact, the sub-sequence $[i]$ converges to a point $\hat{\bs\theta}^*\in\bs\Theta$, $  d\left(\hat{\bs\theta}^*,\bs{\Theta}^0 \right)\geq\epsilon$,  $\hat{\bs\theta}^*\notin\bs{\Theta}^0$. On the other hand, for any $\bs\theta \in \bs\Theta$, we have
\begin{align*}
\bar{l}_p(\hat{\bs\theta}^*|\tilde{\x},\tilde{\y} )-\bar{l}_p(\bs\theta|\x)=& (\bar{l}_p(\hat{\bs\theta}^*|\tilde{\x},\tilde{\y} )-\bar{l}_p(\bs\theta_p^{[i]}|\tilde{\x},\tilde{\y} ) ) + (\bar{l}_p(\bs\theta_p^{[i]}|\tilde{\x},\tilde{\y} ) -l_p^{[i]}(\bs\theta_p^{[i]})) \\
&+(l_p^{[i]}(\bs\theta_p^{[i]}) -l_p^{[i]}(\bs\theta) ) +( l_p^{[i]}(\bs\theta)- \bar{l}_p(\bs\theta|\x)).
\end{align*}
By the continuity of the loss function and $\lim\limits_{i\rightarrow\infty}\bs\theta_p^{[i]}=\hat{ \bs\theta }^*$, the first term in the above equation is arbitrarily small with $i \rightarrow\infty$; by equation (\ref{eqn:supboundmulti1}), the second and forth terms are arbitrarily small with $i \rightarrow\infty $, and the third term is non-positive. Since $\bs\theta\in\bs\Theta$ is arbitrary , we must have $\hat{\bs\theta }^*\in\bs{\Theta}^0$, which is a contradiction. The Proposition is proved.

\section{\large Proof of Proposition \ref{prop:fisher}}\label{app:fisher}
\vspace{-3pt}
WLOG, we derive the Fisher information with the bridge-type noise. The proofs for other types of noise are similar. The Fisher information matrix $I_{\tilde{\x},\tilde{\y}}(\bs\theta)$ on the augmented data is obtained by taking the expectation of the negative second derivative of the noise-augmented loss function in Eqn (\ref{eqn:GLMloss}) over the distribution of data $\x$ and augmented noise $\e$.
\begin{align*}
I_{\tilde{\x},\tilde{\y}}(\bs\theta)
=\E_\x\left(\x^T\bs{B}''(\x)\x\right)+\E_\e\left(\e_{\x}^T\bs{B}''(\e_{\x})\e_\x\right)
=\textstyle{I_{\x,\y}}(\bs\theta)+\E_\e\left(\sum_{i=1}^{n_e}\e_{\x,i}^T{B}''(\e_i\bs\theta)\e_{\x,i}\right),
\end{align*}
where $\bs{B}''(\x)=\diag\{B''(\x_1\bs\theta),\ldots,B''(\x_n\bs\theta\!)\}$ and $\bs{B}(\e_{\x})=\diag\{B''(\e_{\x,1}\bs\theta),\ldots,$ $B''(\e_{\x,n_e}\bs\theta\!)\}$.  Let $\lambda n_e=O(1)$ and $\mbox{V}(\e_{\x,i})$ denote the covariance matrix of $\e_{\x,i}$; take the second-order Taylor expansion around $\e_{\x,i}\bs\theta=0$, we have
\begin{align*}
I_{\tilde{\x},\tilde{\y}}(\bs\theta)=&{I_{\x,\y}}(\bs\theta)+ n_e{B}''(0)\mbox{V}(\e_{\x,i})+O(\lambda n_e^{1/2})J_p\notag\\
=&\textstyle {I_{\x,\y}}(\bs\theta)+(\lambda n_e){B}''(0) \mbox{diag}\{|\bs\theta_{j1}|^{-\gamma},\ldots,|\bs\theta_{jp}|^{-\gamma}\} +O(\lambda n_e^{1/2})J_p,
\end{align*}  
where $J_p$ is a $p\times p$ matrix with all elements equal to 1.

\section{\large Proof of Proposition \ref{prop:asymp.dist.UGM}}\label{app:CI.UGM}\vspace{-3pt}
Given $n^{-1/2}l'(\bs\theta|\x)\overset{d}{\rightarrow} N(0, I^{-1}_1(\bs{\bs\theta}))$, where $l'(\bs\theta|\x)$ is the first derivative of the negative log-likelihood function given the observed data $\x$ and  $I_1(\bs{\bs\theta})$ is the information matrix over one observation. It follows that
\begin{equation}\label{eqn:H1}
n^{-1/2}(l'(\bs\theta|\x)+l'(\bs\theta|\e))=n^{-1/2}l'(\bs\theta|\tilde{\x},\tilde{\y})\overset{d}{\rightarrow} N(n^{-1/2} l'(\bs\theta|\e), I_1(\bs{\bs\theta}))
\end{equation}
where $\e$ is the augmented noise and $l'(\bs\theta|\e)\!=\!\sum_{i=1}^{n_e}l'(\bs\theta|{\e_i})$. Let $\bs\phi(\e)\!=\! n^{-1/2} l'(\bs\theta|\e)$ and it expectation over the distribution of $\e$ can be worked out for different types of noise. For example, with the bridge-type noise, $\bs\phi(\e)\!=\! n^{-1/2} l'(\bs\theta|\e)$  and $\E_{\e}(\bs\phi)=\frac{\lambda n_e}{\sqrt{n}}\sigma^2\mbox{sgn}(\theta_0)$ for Gaussian outcome nodes,  $\frac{\lambda n_e}{8\sqrt{n}} \mbox{sgn}(\theta_0)+\frac{\lambda^2 n_e}{\sqrt{n}}O(|\theta_0|)$ for Bernoulli outcome nodes,  $\frac{\lambda n_e}{2\sqrt{n}}\mbox{sgn}(\theta_0)\!+\!\frac{\lambda^2 n_e}{{n}}O(|\theta_0|)$ for exponential and Poisson outcome nodes and $\frac{\lambda n_e r}{2(r+1){n}}\mbox{sgn}(\theta_0)\!+\!\frac{\lambda^2 n_e}{\sqrt{n}}O(|\theta_0|)\!$ for NB outcome nodes. If  $\lambda n_e\! =\!o(\sqrt{n})$, then $\E_{\e}(\bs\phi)\!\rightarrow\!0$ as $n\!\rightarrow\!\infty$.

Upon the convergence of the PANDA algorithm, the MLE of $\bs\theta$ based on $(\tilde{\x},\tilde{\y})$ is the minimizer $\hat{\bs\theta}_{j,\e}$  from solving $l'(\hat{\bs\theta}_{j,\e})=0$, its  first-order Taylor expansion around $\bs\theta$ is
$l'(\hat{\bs\theta}_{\e})\approx l'(\bs\theta|\tilde{\x},\tilde{\y})+l''(\bs\theta|\tilde{\x},\tilde{\y})(\hat{\bs\theta}_{\e}-\bs\theta)=0$. Therefore, $\hat{\bs\theta}_{\e}-\bs\theta=-(l''(\bs\theta|\tilde{\x},\tilde{\y}))^{-1}l'(\bs\theta|\tilde{\x},\tilde{\y})$ and  $\sqrt{n}\left(\hat{\bs\theta}_{\e}-\bs\theta\right)=-(n^{-1}l''(\bs\theta|\tilde{\x},\tilde{\y}))^{-1}\left(n^{-1/2}l'(\bs\theta|\tilde{\x},\tilde{\y})\right)$, where $l''(\bs\theta|\tilde{\x},\tilde{\y})$ is the Hessian matrix and $l''(\bs\theta|\tilde{\x},\tilde{\y})\rightarrow I_p(\bs\theta)$ as $n\rightarrow \infty$. Taken together with Eqn (\ref{eqn:H1}),  assume $\lambda n_e\! =\!o(\sqrt{n})$,  by the  Slutsky's theorem, as $n\rightarrow\infty$
\begin{align*}
\sqrt{n}\left(\hat{\bs{\theta}}_{\e}-\bs{\theta})\right)&\!=\! (n^{-1}l''(\bs\theta|\tilde{\x},\tilde{\y}))^{-1}\left(n^{-1/2}l'(\bs\theta|\tilde{\x},\tilde{\y})\right)\overset{d}{\rightarrow}\! N\left(\0 , I_p(\bs\theta)^{-1}I(\bs\theta)I_p(\bs\theta)^{-1} \right)\!\overset{\Delta}{=}N(\0,\Sigma_{\e}).
\end{align*}
When the mean of $m>1$ estimates over  consecutive iteration are taken as the final estimate for $\bs{\theta}$, that is $\bar{\bs\theta}=m^{-1}\sum_{t=1}^m\hat{\bs\theta}_{\e}^{(t)}$, the variability among the $m$ consecutive estimates will need to be accounted for and be reflected in the variance of the final estimate. It is easy to establish this in the Bayesian framework. Specifically, 
\begin{align*}
\E(\bs\theta|\x)&=\textstyle\E_\e(\E(\bs\theta|\tilde{\x},\tilde{\y}))=\E_\e(\hat{\bs{\bs\theta}}_{\e})=
m^{-1}\sum_{t=1}^m\hat{\bs\theta}^{(t)}_{\e}\triangleq \bar{\bs\theta}\mbox{ as }m\rightarrow\infty\\
V(\bs\theta|\x)&=
\E_\e(V(\bs\theta|\tilde{\x},\tilde{\y}))+V_\e(\E(\bs\theta|\tilde{\x},\tilde{\y}))=\E_\e(\Sigma_{\e})+V_\e(\hat{\bs{\bs\theta}}_{\e})\triangleq \bar{\Sigma}+ \Lambda\\
&=\textstyle m^{-1}\sum_{t=1}^m \Sigma_{\e}^{(t)}+(m-1)^{-1}\sum_{t=1}^m
\left(\hat{\bs{\bs\theta}}^{(t)}-\bar{\bs\theta}\right)
\left(\hat{\bs{\bs\theta}}_{\e}^{(t)}-\bar{\bs\theta}\right)' \mbox{ as }m\rightarrow\infty.
\end{align*}
Per the large-sample Bayesian theory, the posterior mean and variance of $\bs{\bs\theta}$ given $\x$ are asymptotically equivalent ($n\rightarrow\infty$) to the MLE for $\bs{\bs\theta}$ and the inverse information matrix of $\bs\theta$ contained in $\x$. In other words,
$$\textstyle \sqrt{n}(\bar{\bs{\bs\theta}}-\bs{\bs\theta})\rightarrow N\left( \0,\bar{\Sigma}+ \Lambda \right).$$
In the case of a finite $m$ (as in practical application), $\V(\bs\theta|\x)$ is estimated
by $\bar{\Sigma}+(1+m^{-1})\Lambda$ with the correction for the finite $m$. 
Applying Proposition \ref{prop:asymp.dist.UGM} with lasso-type noise, we have 
$$\sqrt{n}(\hat{\bs{\bs\theta}}-\bs{\bs\theta})
\rightarrow N\left(n^{-1/2}\lambda n_e \mbox{sgn}(\bs\theta) M^{-1},\sigma^2M^{-1}(\y'\x)M^{-1} \right),$$
where $M=(\y'\x+\diag(\lambda n_e|\bs\theta|^{-1}))$ and $\sigma^2$ is the variance of the error term in the linear regression, and is estimated by 
\begin{align*}
\hat{\sigma}^2=&\mbox{SSE}(n-\nu)^{-1}
=(n-\nu)^{-1}(\x\bs\theta+\epsilon)'(I-H)(\x\bs\theta+\epsilon)\\
=&(n-\nu)^{-1}\epsilon'(I-H)\epsilon+(n-\nu)^{-1}\left(\bs\theta'\x'(I-H_\x\bs\theta+2\bs\theta'\x'(I-H)\epsilon\right)
\end{align*}
where $H=\x(\y'\x+\diag(\lambda n_e|\bs\theta|^{-1}))^{-1}\y' \mbox{ and  }\nu=\mbox{trace}(H)$.

\section{\large A Formal Test on the Convergence of the PANDA Algorithm}\label{sec:convergence}
When presenting the PANDA algorithm  in Sec \ref{sec:algorithm}, we state that a formal statistical test can be used to test convergence. This test is based on the assumption of $n_e\!\rightarrow\!\infty$ or $m\!\rightarrow\!\infty$ and should work well when either $n_e$ or $m$ is large in practice. WLOG, we establish the test below for $n_e\!\rightarrow\!\infty$; the procedure is similar for $m\!\rightarrow\!\infty$ by replacing $n_e$ with $m$.

Theorem \ref{thm:as} shows that as  $n_e\rightarrow\infty$, the distribution of the loss function in iteration $t$ converges to a Gaussian distribution (Eqn (\ref{eqn:d1})). The asymptotic Gaussian distribution involves $C_1(\bs\theta)$, which is unknown and can be estimated by plugging the $\hat{\bs\theta}^{(t)}$ from the current iteration $t$. Specifically, 
$$\textstyle C_1^{(t)}= \frac{\lambda n_e}{2}\!\left(\kappa  \bigg|\bigg|\!\left(\hat{\bs\theta}^{(t)}\big|{\hat{\bs\theta}^{(t)}}\big|^{-\gamma/2}\right)\left(\hat{\bs\theta}^{(t)}\big|{\hat{\bs\theta}^{(t)}}\big|^{-\gamma/2}\right)^T\bigg|\bigg|_2^2\right)^{1/2},$$
where $\kappa$ is a constant that depends on the type of $Y$ ($\kappa=8$ for Gaussian,  $2\exp(2\theta_0)/(1+\exp(2\theta_0))^4$ for Bernoulli, $2\exp(2\theta_0)$ for Poisson, $2$ for Exponential, and $2r^2\exp(2\theta_0)/(r+\exp(\theta_0))^2$ for NB; see Eqns (\ref{app:asygauss1}), (\ref{app:asypoi1}) and (\ref{app:asynb1})). Let $d^{(t)}=\bar{l}_p(\tilde{\x}^{(t+1)},\tilde{\y})-\bar{l}_p(\tilde{\x}^{(t)},\tilde{\y})$ denote the difference in the loss function from two consecutive iterations of the PANDA algorithm, which is $n_e^{-1/2}(C_1^{(t+1)}z^{(t+1)}-C_1^{(t)}z^{(t)})$ per Eqn (\ref{eqn:d1}). If the PANDA algorithm converges, the estimates $\hat{\bs\theta}^{(t)}$ stabilizes, so does $C_1^{(t)}$; in other words, $C_1^{(t+1)}\approx C_1^{(t)}$ and a nonzero $d^{(t)}$ is mostly due to the randomness of the injected noise with an expected mean of 0; that is,
\begin{equation}\label{eqn:d}
z^{(t)}= d^{(t)}/\sqrt{n^{-1}_e \left[C_1^{(t)2}+C_1^{(t+1)2}\right]}.
\end{equation}
Since $z^{(t)}$ is independent from $z^{(t+1)}$ (augmented noises  are drawn independently across iteration).  If $|z^{(t)}|>z_{1-\alpha/2}$, then we may claim the PANDA algorithm has not converged at iteration $t$ at the significance level of $\alpha$. 

The denominator in Eqn (\ref{eqn:d}) assumes $C_1^{(t)}$ and $C_1^{(t+1)}$ are independent whey are likely to positively correlated as both use the original data $(\x,\y)$. With the under-estimated variance, $z^{(t)}$ would be over-estimated, and convergence is likely to rejected more often than necessary.

\section{\large Minimizer of Averaged noise-augmented Loss Function vs Averaged minimizer of Noise-augmented Loss Functions}\vspace{-3pt}
Per Proposition \ref{prop:glmregularization}, one would take the average over $m$ noise-augmented loss function $l(\Theta|\x,\e)$ to yield a single minimizer $\hat{\bs{\theta}}$, which is the Monte Carlo version of $\E_{\e}(l_p(\bs{\theta}|\x,\e)$ as $m\rightarrow\infty$. However, PANDA would lose its computational edge. To maintain the computational advantage for PANDA, we  instead calculate $\bar{\bs{\theta}}$, the  average of $m$ minimizers of $l(\Theta|\x,\e)$ from the latest $m$ iterations, which is the approach that the PANDA algorithm uses.  
We establish in Corollary \ref{cor:average} that $\bar{\bs{\theta}}$ and $\hat{\bs{\theta}}$  are equivalent under some regularity conditions.   We also present some numerical examples below to  illustrate the similarity between  $\bar{\bs{\theta}}$ and $\hat{\bs{\theta}}$.
\begin{cor}[\textbf{First-order equivalence between minimizer of averaged noise-augmented loss functions vs averaged minimizers of single noise-augmented loss functions}]\label{cor:average}
The average $\bar{\bs\theta}$ of $m$ minimizers of the $m$ perturbed loss functions upon convergence is first-order equivalent to the minimizer $\hat{\bs\theta}$ of the averaged $m$  noise-augmented loss functions as $m\rightarrow\infty$ or  as $n_e\rightarrow\infty$ while $V(\theta_jn_e)=O(1)$. In addition, The  higher-order difference between  $\bar{\bs\theta}$ and $\hat{\bs\theta}$ also approaches 0 as $n_e\rightarrow\infty$ while $V(\theta_jn_e)=O(1)$. \vspace{-3pt}
\end{cor}
Proof: WLOG, we work with the bridge-type noise. in this proof. The average of the minimizers of the $m$ loss functions is 
\begin{align}\label{eqn:thetabar}
\bar{\bs\theta}=\textstyle  m^{-1}\!\sum_{t=1}^{m}\left(\x'\x+\sum_{i=1}^{n_e}\e_{i,\x}^{(t)'}\e^{(t)}_{i,\x}\right)^{-1}\!\!\x'\y,
\end{align}
where $e_{ij}\sim N(0,\lambda|\theta_j|^{-1})$. Let $\sum_{i=1}^{n_e}\!\e_{i,\x}^{(t)'}\e^{(t)}_{i,
\x}\!=\!\E\left(\!\sum_{i=1}^{n_e}\!\e_{i,\x}^{(t)'}\e^{(t)}_{i,\x}\!\right) + A^{(t)}\!= \!\diag(\lambda n_e|\bs\theta|^{-\gamma})+\bar{A}^{(t)}$. $A^{(t)}$ can be regarded as the sample deviation of $\sum_{i=1}^{n_e}\e_{i,\x}^{(t)'}\e^{(t)}_{i,\x}$ from its mean. Let $\bar{A}=m^{-1}\sum_{t=1}^{m}\bar{A}^{(t)}$, the elements of which are
\begin{align}\label{eqn:Abar}
\begin{cases}
\bar{A}[j,j]
\!=\!m^{-1}\sum_{t=1}^{m}\!\sum_{i=1}^{n_e}e_{ij}^{(t)2}-\lambda n_e|\theta_j|^{-1}&\sim \lambda|m\theta_j|^{-1}(\chi^2_{n_em}\!-\!n_em)\\
\bar{A}[j,k]
\!=\! m^{-1}\sum_{t=1}^{m}\sum_{i=1}^{n_e}e^{(t)}_{ij}e^{(t)}_{ij}
&\sim \lambda|\theta_j\theta_j|^{-\frac{1}{2}}m^{-1}\!\sum_{t=1}^{m}\!\sum_{i=1}^{n_e}\!z_{ti}z'_{ti}
\end{cases},
\end{align}
where $z_{ti}\sim N(0,1)$ and $z_{ti}'\sim N(0,1)$ independently.  Let $S=(\x'\x\!+\diag(\lambda n_e|\bs\theta|^{-1}))^{-1}$. The Taylor expansion of the inverse of the sum of two matrices, assuming $A^{(t)}$ to be  a small increment, is $(S^{-1}+A^{(t)})^{-1}\!=\!S-SA^{(t)}S+SA^{(t)}SA^{(t)}S+\ldots$ Therefore, Eqn (\ref{eqn:thetabar}) becomes
\begin{align}\label{eqn:bar}
\bar{\bs\theta}=\textstyle S\x'\y- S\left(\bar{A}+O(\lambda^2 n_e)\right)S\x'\y. \end{align}
On the other hand, the minimizer of the average of $m$ loss functions  is
\begin{align}
\hat{\bs\theta}=&\textstyle\!\left(\x'\x+\sum_{i=1}^{n_em}\hat\e_{ij}'\hat\e_{ij}\right)^{-1}\!\!\x'\y\!=\! \left(\x'\x\!+\!\diag(\lambda n_e|\bs\theta|^{-1})\!+\!\hat{A}\right)^{-1}\!\!\x'\y,\notag\\
=&S\x'\y- S\left(\hat{A} +O(\lambda^2 n_e)\right)S\x'\y,
\label{eqn:hat}
\end{align}
where $e_{ij}\sim N(0,\lambda|m\theta_j|^{-1})$ for the sake of yielding the same regularization effect as imposed on $\bar{\bs\theta}$; and $\hat{A}$ is defined in a similar manner as $\bar{A}$, the elements of which are
\begin{align}\label{eqn:Ahat}
\begin{cases}
\hat{A}[j,j]=\textstyle\sum_{i=1}^{n_em}e_{ij}^2-\lambda n_e|\theta_j|^{-1}&\sim \lambda|m\theta_j|^{-1}(\chi^2_{n_em}-n_em)\\
\hat{A}[j,k]=\textstyle\sum_{i=1}^{n_em}e_{ij}e_{ij} &\sim\frac{\lambda}{m}|\theta_j\theta_k|^{-\frac{1}{2}}\sum_{i=1}^{n_em}z_iz_i'\\
\end{cases},
\end{align}
where $z_i\sim N(0,1)$ and $z_i'\sim N(0,1)$ independently. $\bar{A}$ and $\hat{A}$ in Eqn (\ref{eqn:Abar}) and (\ref{eqn:Ahat}) follow the same distribution. The expected values of $\bar{A}[j,j], \bar{A}[j,k], \hat{A}[j,j]$, and $\hat{A}[j,k]$ are all equal to zero; the variance of $\bar{A}[j,j]$ and $\hat{A}[j,j]$ is $\lambda^2|m\theta_{jk}|^{-2}2n_em=2\lambda (\lambda n_e)|\theta_{jk}|^{-2}2/m$, and that of $\bar{A}[j,k]$ and $\hat{A}[j,k]$ is $\lambda^2m^{-2}|\theta_{jk}\theta_{jl}|^{-1}n_em=\lambda(\lambda n_e)|\theta_{jk}|^{-2}2/m$. As $m$ increases, both variance terms shrink to 0. As $n_e$ increases while $O(n_e\lambda)=1$, then both variance terms shrinks to 0 as well. In other words, we expect $\bar{A}$ and $\hat{A}$ to be very similar. As such, $\bar{\bs\theta}$ in Eqn (\ref{eqn:bar}) and $\hat{\bs\theta}$ in Eqn (\ref{eqn:hat}) are also very similar. In addition, as $n_e$ increases and $\lambda n_e=O(1)$, the higher-order terms also goes to 0.

\begin{figure}[!htb]
\centering
\begin{minipage}{0.495\textwidth}
linear regression\\
\includegraphics[width=1\linewidth]{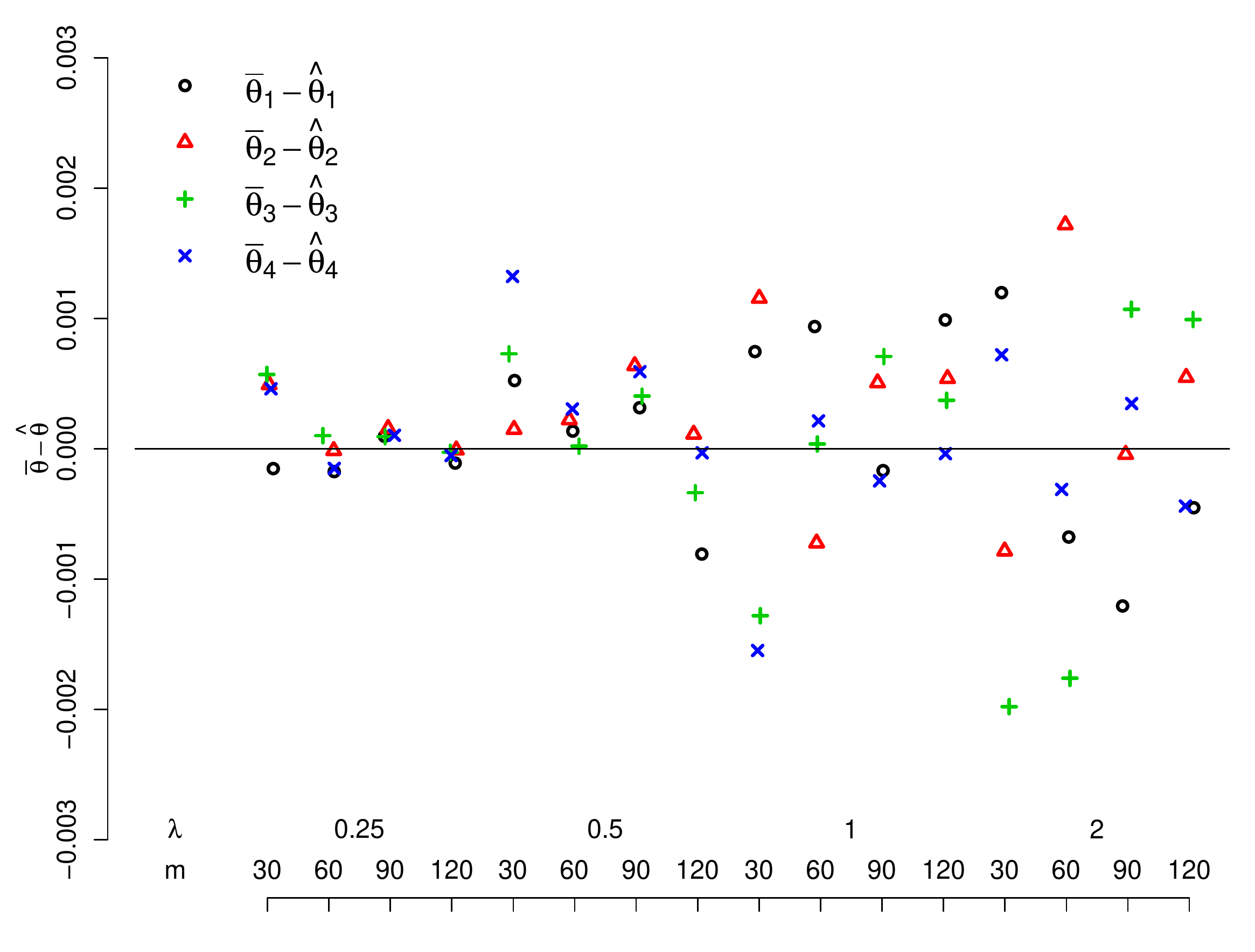}\\
\end{minipage}
\begin{minipage}{0.495\textwidth}
Poisson regression \\
\includegraphics[width=1\textwidth]{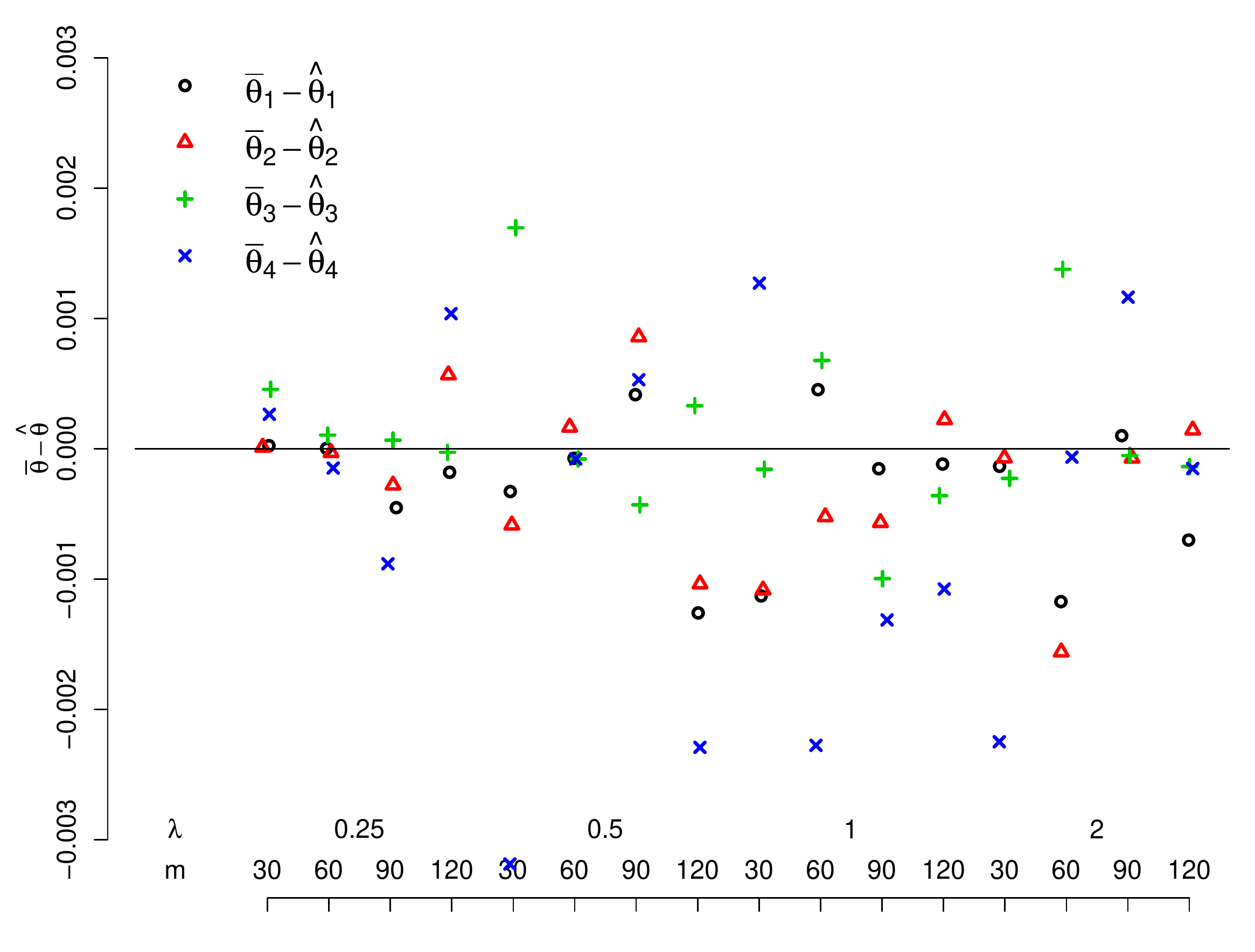}
\end{minipage}
\caption{Differences between $\bar{\bs{\theta}}$ and $\hat{\bs{\theta}}$ in linear regression (top) and in Poisson regression (bottom)}\label{fig:average1}
\end{figure}

To first illustrate the similarity between $\bar{\bs\theta}$ and $\hat{\bs\theta}$, we simulated data ($n=30$) from linear regression and a Poisson regression models, where the linear predictor is $\X^T\bs\theta=X_1+0.75X_2+0.5X_3+0X_4$. $\X$ and the error in the linear regression was simulated from N$(0, 1)$ independently. The PANDA augmented noises $\e$ in both cases were drawn from N$(0,\lambda^2)$ with $n_e=200$. We examined $m=30,60,90,120$ and $\lambda^2=0.25,0.5,1,2$, calculated $\hat{\bs{\theta}}$ and $\bar{\bs{\theta}}$, and plotted their differences in Figure \ref{fig:average1}. The results show minimal difference between $\hat{\bs{\theta}}$ and $\bar{\bs{\theta}}$.

\section{\large PANDA in Each Iteration Realizes Weighted Ridge in Linear Regression} \label{sec:wrr}
\begin{cor}[\textbf{PANDA and weighted ridge regression}]\label{cor:wrr}
The OLS estimator in each iteration of PANDA on the noise augmented data is equivalent to the weighted ridge estimator $\hat{\bs{\theta}}
=\left(\x^T\x\!+\!\e^T\e\right)^{-1}\x^T\y$.
\vspace{-9pt}
\end{cor}
The proof is straightforward. Let $\tilde{\x}=(\x,\e_x)^T$. In each iteration of PANDA,  the OLS estimator $\hat{\bs{\theta}}=(\tilde{\x}^T\tilde{\x})^{-1}\tilde{\x}^T(\y,\mathbf{0})=(\x^T\x+\e_x^T\e_x)^{-1}{\x}\y$, leading to Corollary \ref{cor:wrr}. If $n_e\rightarrow \infty$, then $\e_x^T\e_x\rightarrow n_e\mbox{V}(\e_x)$. For example, if the NGD is N(0,$\lambda|\theta|^{-\gamma}_j$), then  $n_e\mbox{V}(\e_x)= \mbox{diag}((n_e\lambda)|\theta|^{-\gamma}_j)$; and  $(\lambda n_e)$ can be tuned as one single tuning parameter.

\end{document}